%% file: RobotFab_Paper.tex
\documentclass[final,5p,times,twocolumn]{elsarticle} 

\biboptions{sort&compress}

\graphicspath{{./figures/}}
\newcommand*{\BiblioPath}{./biblio}%

\usepackage{amsmath}
\usepackage{amssymb}
\usepackage{soul} 
\usepackage{pdflscape} 
\usepackage{enumitem} 

\usepackage{ragged2e}
\usepackage[hyphens]{url} 
\usepackage[colorlinks = True, breaklinks=true]{hyperref} 
\hypersetup{allcolors=blue, pdfauthor = author}
\usepackage[nameinlink]{cleveref}

\usepackage{caption}
\usepackage{subcaption}
\usepackage{float} 
\usepackage[export]{adjustbox} 
\usepackage{stfloats} 

\usepackage[table]{xcolor} 
\usepackage{makecell} 
\usepackage{array} 
\usepackage{booktabs} 
\usepackage[flushleft]{threeparttable} 
\usepackage{longtable} 
\usepackage{multirow} 
\usepackage{tabularx} 
\usepackage{siunitx} 

\journal{Engineering Structures}
\begin{document}

\begin{frontmatter}

\title{Three Cooperative Robotic Fabrication Methods for the Scaffold-Free Construction of a Masonry Arch}

\author[1,2]{Edvard P.G. Bruun\corref{cor1}}
\ead{ebruun@princeton.edu}

\author[2]{Rafael Pastrana}
\author[3]{Vittorio Paris}
\author[4]{Alessandro Beghini}
\author[3]{Attilio Pizzigoni}
\author[2]{Stefana Parascho}
\author[1]{Sigrid Adriaenssens}

\cortext[cor1]{Corresponding author}

\address[1]{Form Finding Lab, Civil and Environmental Engineering Department, Princeton University}
\address[2]{CREATE Lab, School of Architecture, Princeton University}
\address[3]{Department of Engineering, University of Bergamo}
\address[4]{Associate Director, Skidmore, Owings \& Merrill (SOM)}


\begin{abstract}
    Geometrically complex masonry structures (e.g., arches, domes, vaults) are traditionally built with expensive scaffolding or falsework to provide stability during construction. The process of building such structures can potentially be improved through the use of multiple robots working together in a cooperative assembly framework. Here a robot is envisioned as both a placement and external support agent during fabrication -- the unfinished structure is supported in such a way that scaffolding is not required. The goal of this paper is to present and validate the efficacy of three cooperative fabrication approaches using two or three robots, for the scaffold-free construction of a stable masonry arch from which a medium-span vault is built. A simplified numerical method to represent a masonry structure is first presented and validated to analyse systems composed of discrete volumetric elements.  This method is then used to evaluate the effect of the three cooperative robotic fabrication strategies on the stability performance of the central arch. The \textit{sequential method} and \textit{cantilever method}, which utilize two robotic arms, are shown to be viable methods, but have challenges related to scalability and robustness. By adding a third robotic agent, it becomes possible to determine a structurally optimal fabrication sequence through a multi-objective optimization process. The \textit{optimized three robot method} is shown to significantly improve the structural behavior over all fabrication steps. The modeling approaches presented in this paper are broadly formulated and widely applicable for the analysis of cooperative robotic fabrication sequences for the construction of discrete element structures across scales and materials.
    
\end{abstract}


\end{frontmatter}

\raggedbottom
\hyphenpenalty=1000


\input{\SectionPath/1intro}

\input{\SectionPath/2lit}

\input{\SectionPath/3modeling}

\input{\SectionPath/4twoRob}

\input{\SectionPath/5threeRob}
\input{\SectionPath/6conclusions}

\input{\SectionPath/7acknowledgements}

\bibliographystyle{\BiblioPath/elsarticle-num} 
\bibliography{\BiblioPath/RobotFab_Paper}

\input{\SectionPath/99appendix}


\end{document}

%% file: sections/1intro.tex

\section{Introduction}\label{intro}
    A self-supporting construction process is one where no external support (e.g., temporary scaffolding or formwork) is required for the structure to remain stable as it is being built. Note that the terms self-supporting \citep{deuss_assembling_2014} or self-balancing \citep{paris_statics_2020} are used interchangeably in the literature -- \textit{self-supporting} is used herein to describe this structural behavior. Such methods have been used throughout history to build large-scale complex masonry structures, relying on both the brick tessellation pattern and a careful design of the overall form to guarantee stability at all phases \citep{choisy_art_1883, el-naggar_les_1999, besenval_technologie_1984, heyman_stone_1997, pizzigoni_brunelleschis_2015, pizzigoni_herringbone_2018, ochsendorf_guastavino_2010}. Although the contemporary building industry has generally moved away from many of these traditional methods, recently there has been a resurgence of interest on the investigation of self-supporting construction methods \citep{vouga_design_2012, deuss_assembling_2014, goessens_feasibility_2018, som_bricks_2019, paris_statics_2020, loing_free-form_2020, drew_lock-block_2021}. The emergence of automation, pre-fabrication, computational design, and robotic construction has created an opportunity to re-imagine how self-supporting construction techniques can find new relevance today \citep{wu_robotic_2018}.
    
    
    In striving to develop self-supporting construction methods for masonry construction, a robot becomes more than just a fabrication tool, but is central to shaping what type of structure is feasible to build --  informing the fabrication process through a Robot Oriented Design (ROD) framework \citep{bock_robot-oriented_2015}. Due to their task-versatility \citep{bravo-palacios_one_2020} and spatial precision \citep{eversmann_robotic_2017}, standardized industrial robotic arms have seen continuing widespread growth in industry adoption over the last decade \citep{ifr_world_2018}. Their application is improved by virtue of being able to perform a wide range of functions when paired with customized end-effectors designed for versatility \citep{han_concept_2020}. During fabrication, in a cooperative assembly context (i.e., multiple robots working together), robots with a gripping functionality can take turns performing either the function of picking up and aggregating structural components, or holding and providing temporary support over indefinite periods of time to a partially completed structure \citep{parascho_robotic_2020, parascho_lightvault_2021, parascho_cooperative_2017}. Therefore, when properly sequenced, a cooperative fabrication method has the potential to achieve complex structural goals -- such as building a discrete element structure without temporary scaffolding while maintaining appropriate structural behavior.
    
    Developing such fabrication-informed construction sequences presents a departure from a traditional structural engineering workflow, which places emphasis on the design of the finished structure \citep{sharif_bim_2015}. While this final state is conceivably optimized for a certain structural behaviour or load combination, the construction sequencing and intermediate form required to reach this finished state is seldom optimized. But designing and optimizing a fabrication strategy around the goal of self-support is computationally intensive as all intermediate steps in the construction process need to be evaluated for structural performance -- for the construction of a masonry system this can mean potentially evaluating the placement of 100's of bricks. Thus the goal of this paper is twofold: (1) to develop, validate, and apply a computationally efficient framework for a high-level structural analysis of discrete element assemblies, (2) to apply the framework to validate the viability of three different cooperative robotic approaches used for to the scaffold-free construction of the central arch, which is part of a complex curved masonry vault detailed in \Cref{sec:2_prototype}.
    
    \subsection{Summary of Contents}
        The paper starts with a brief review of literature on the topic of automation in masonry construction, cooperative robotic fabrication for discrete element structures, and modeling approaches for such structures. The following section provides a description of the masonry vault prototype referred to in this study, and the development and calibration of a simplified modeling approach used to represent it. Next, three cooperative robotic methods are described and analyzed in the context of the scaffold-free construction of a masonry arch: \textit{sequential method}, \textit{cantilever method}, and the \textit{optimized three robot method}. The paper concludes with a discussion of the results from applying these methods and a summary of the main contributions from this study.

%% file: sections/2lit.tex

\section{Review of Robotic Fabrication and Masonry Structures}
    In the context of the Architetecture, Engineering and Construction (AEC) field, robotic manufacturing was first applied at a significant scale to the prefabrication of modular homes in Japan in the 1970s \citep{bock_construction_2007, bock_future_2015} and proliferated in numerous specialized applications in that country over the next decade \citep{albus_trip_1986, cousineau_construction_1998, huang_factor_1990, ueno_construction_1986}. The industry has experienced increasing adoption through technological improvements to automation and as a robot's ability to work in unstructured site environments has improved \citep{bock_site_2016, skibniewski_robotics_1989}. Continuing future growth in adoption is motivated by numerous advantages to the construction industry: substantially improving productivity for complex structures  \citep{davila_delgado_robotics_2019}, improving worker safety \citep{castro-lacouture_construction_2009}, and reducing material and labour costs \citep{garcia_de_soto_productivity_2018, kumar_robotics_2016}.
    
    \subsection{On-site Automated Bricklaying in Industry}\label{sec:2_automated_brick}
        While construction robotics has seen a growth in application for off-site prefabrication \citep{bock_robotic_2015}, adoption has lagged for certain complex structural forms or material systems that are typically assembled on-site (e.g., stone/brick masonry \citep{block_structural_2018, rippmann_computational_2016, davis_innovative_2012, borne_droneport_2016} or cast-in-place concrete shells \citep{chilton_heinz_2010, chilton_rooted_2017}). Although Single Task Construction Robots (STCRs) \citep{bock_construction_2016} can be developed for specialized on-site applications, these are generally less common and harder to implement as they are motivated by project-specific economic factors \citep{bock_site_2016}.
        
        Discrete element assemblies, such as masonry structures, are favourable candidates for robotic automation as they are physically taxing to construct manually \citep{vink_physical_2002, hess_ergonomic_2010}. The construction of masonry structures is also highly repetitive and individual units are relatively lightweight, which makes such structures fit for development around an automated framework \citep{dakhli_robotic_2017}. While robotic prefabrication of masonry in the factory setting is possible, the output is geometrically limited to highly standardized vertical block walls due to fixed space and equipment design constraints \citep{bock_robotic_2015}. Thus, automation in masonry construction is generally better suited to in-situ fabrication approaches as a means to better address site variability or construct more complex geometry. The technical exploration of automated in-situ masonry construction dates back to the beginning of the 20th century, with a patent for an automated brick-laying machine submitted in 1904 \citep{thomson_brick-laying_1904}. A functioning automated linear bricklaying machine prototype was documented on-site in the 1960s \citep{british_pathe_mechanical_1967}. But it was first in the late 1980s that developments in the automation of in-situ masonry construction started to flourish \citep{malinovsky_robotic_1990, chamberlain_progress_1991, altobelli_prototype_1993}, from balancer and handling assistance machines \citep{bock_robotic_2015}, to Selective Compliance Articulated Robot Arm (SCARA) systems such as: the Solid Material Assembly System (SMAS) \citep{bock_early_2009, kodama_robotized_1988}, the Robotic Construction System for Computer Aided Construction (ROCCO) \citep{bock_robot-oriented_2015, andres_first_1994}, and the Blockbot \citep{slocum_blockbot_1988, lehtinen_outlines_1989}. Recent years have seen further commercial developments in the form of the Semi-Automated Mason (SAM100) \citep{podkaminer_sam100_2021} and Hadrian X robot \citep{pivac_hadrian_2021}, which can be thought of as technical successors to the previously developed articulated robotic arm systems \citep{pritschow_mobile_1994, pritschow_technological_1996}. SAM100 consists of a robotic arm mounted on a track, while Hadrian X uses a custom gripper mounted on a mobile truck crane for improved reachability \citep{dakhli_robotic_2017, sklar_robots_2015}.

    \subsection{Digital Fabrication and Geometric Complexity}\label{sec:2_dfab} 
        The automated brick-laying systems summarized in \Cref{sec:2_automated_brick} were focused on industry applications for the construction of straight walls and generally improving construction productivity. But only in the last two decades has the field of digital fabrication (dfab) been seriously applied in an academic context to construction at the large-scale with the specific intention of expanding the geometric design space through the use of robots \citep{gramazio_digital_2008, gramazio_made_2014}. The first such project saw stationary industrial robots used to stack bricks to build prefabricated load-bearing but non-standardised undulating walls \citep{bonwetsch_informed_2006, bonwetsch_digitally_2007, kohler_gantenbein_2014}. Later the DFAB house \citep{empa_dfab_2021} became the showcase for how different non-standardized components in a building can be shaped through digital design and robotic fabrication process \citep{willmann_robotic_2016, hack_mesh_2017, hack_structural_2020}.  In the context of masonry construction, the development of specialized digital design software \citep{bonwetsch_brickdesign_2012, mele_compas_2017} and augmented reality \citep{jahn_holographic_2018, jahn_holographic_2019, mitterberger_augmented_2020} have allowed for ever more geometrically complex geometries to be materialized on site. Industrial robotic arms have also been been applied in complex setups, for example on gantry cranes \citep{piskorec_brick_2018} and as mobile platforms \citep{dorfler_mobile_2016, giftthaler_mobile_2017}.

    \subsection{Cooperative Robotics and Discrete Element Structures} \label{sec:lit_discrete}
        Despite the technical advances associated with the dfab movement, robotic fabrication is still rarely utilized beyond the construction of vertical layer-based structures \citep{bartschi_wiggled_2010, kohler_programmed_2014}. Structural stability and the detrimental effect of tensile forces in cantilevered or spanning forms are reasons why masonry robotic brick fabrication projects have been limited to vertical assemblies. Similar challenges are faced in additive concrete manufacturing, which draws inspiration from self-supporting masonry approaches as a means to develop more robust methods for 3D printing overhanging geometry \citep{carneau_exploration_2019, carneau_additive_2020, motamedi_supportless_2019}.
        
        Recent work has suggested that cooperative fabrication can be a means to address issues surrounding stability during construction and to therefore realize complex structural forms that would not be possible otherwise. An example of this is the assembly of metal rods into complex (i.e. non-uniform) space frame structures \citep{parascho_cooperative_2017, parascho_cooperative_2019} where complex joints between rods had to be carefully sequenced using two robots. Or in the fabrication of custom non-planar timber modules where cooperating robots were used to minimize the need for scaffolding \citep{thoma_robotic_2018}. Another example is the construction of a small-scale branching structure, where two robots would take turns supporting the structure as it was being assembled \citep{bruun_humanrobot_2020}. While operating at different scales, these projects all explore the potential of creating a construction sequence where two robots work together cooperatively to assemble a geometrically complex discrete element structure. 
        
    \subsection{Analysis and Modeling of Masonry Structures} 
        A number of modeling strategies exist to analyze the structural behavior of three-dimensional masonry structures. As recently surveyed in \citep{sarhosis_review_2016,daltri_modeling_2020, fang_assessing_2019}, finite element (FE), discrete element (DE), and geometric approaches are all feasible. In an FE setting, a continuum mesh of shell elements are calibrated to approximate the macro-behaviour of a masonry structure \citep{kaminski_tests_2010, sarhosis_review_2016}. The interaction between masonry elements and joint interfaces is homogenized to this end. Meanwhile, DE strategies are able to accurately capture global- and local-scale behaviour by representing a masonry structure as a system of three-dimensional rigid or deformable bodies \citep{thavalingam_computational_2001,simon_discrete_2016, cundall_formulation_1988, hart_formulation_1988}. As shown in literature \citep{roberti_distinct_1998, kassotakis_discrete_2017, meriggi_distinct_2019}, DE approaches are particularly suitable to highlight the detailed structural phenomena between masonry elements and their joint interfaces, such as detachment, collision and sliding. While commercial software packages exist to facilitate the structural analysis of a masonry structure using a FE (e.g. Abaqus \citep{abaqus_inc_abaqus_2020}) or a DE focus (e.g. 3DEC \citep{itasca_consulting_group_inc_3dec_2016}), creating and running these models can be a time-consuming and computationally expensive unsuitable for explorative optimization processes. Inspired by Heyman's safe theorem \citep{heyman_stone_1966, heyman_stone_1997}, geometric strategies assess the static equilibrium of a masonry structure by modeling it as a network of compression forces. In the literature, geometrical approaches tailored to two-dimensional \citep{block_as_2006, lau_equilibrium_2006, michiels_form-finding_2018} and three-dimensional masonry structures \citep{odwyer_funicular_1999, block_thrust_2007, vouga_design_2012} have been proposed. Unlike their FE or DE counterparts, geometric strategies do not solicit material information about a masonry structure and are agnostic to the arrangement of masonry elements and joint types. While this can simplify and speed up the modeling process, geometric approaches overlook potentially relevant structural actions occurring at the joint interfaces which may be critical to assessing the inter-construction states in a masonry structure.
        
    \subsection{Next Steps in Robotic Fabrication}
        Research on robotic construction in industry has focused on technical advances in the automation of traditional building practices. Meanwhile, recent academic research has focused on expanding design possibilities, using robots to achieve local geometric differentiation. We see opportunities in utilizing robotic construction to combine both of these approaches -- keeping material efficiency and geometric complexity central to the process with the overall goal of making construction less labor intensive and more productive. Recent work \citep{parascho_robotic_2020, parascho_lightvault_2021, han_concept_2020} has demonstrated the feasibility of moving away from the vertical construction paradigm common to robotic masonry fabrication, by sequencing two robotic arms to build a complex vaulted structure without temporary scaffolding. This paper builds on this work by providing a numerical basis for how to develop and evaluate such robotic fabrication sequences. It also extends the fabrication framework developed in \citep{parascho_robotic_2020, parascho_lightvault_2021, han_concept_2020} by introducing the concept of multi-objective optimization to determine how robotic support positions can be determined while satisfying structural performance criteria.

%% file: sections/3modeling.tex

\section{Methodology}\label{mod_approach}

    \subsection{Masonry Vault Prototype}\label{sec:2_prototype}
\begin{figure}[H]
            \centering
    		\includegraphics [trim={0cm 0cm 0cm 0cm}, clip, width=0.95\linewidth]{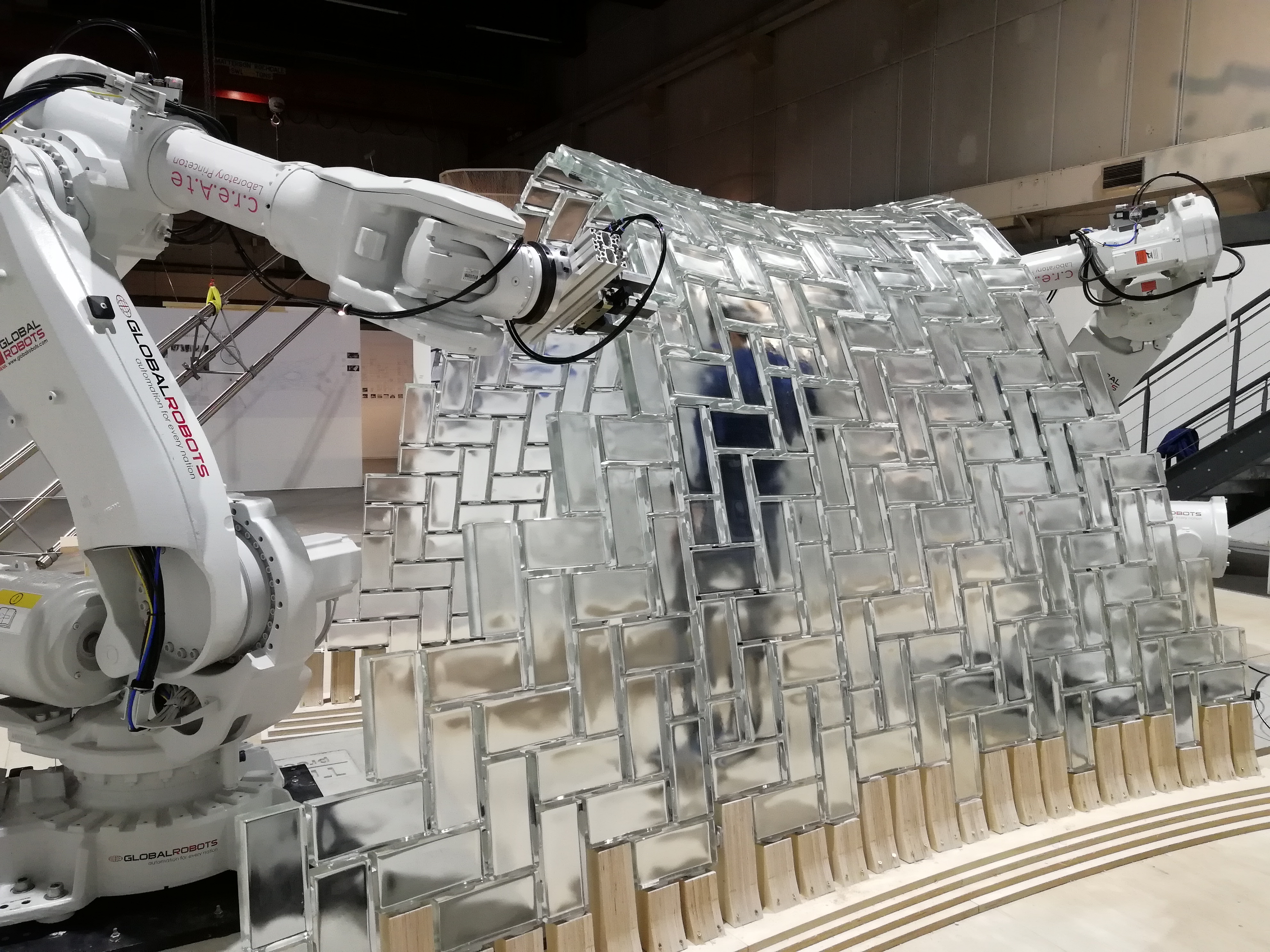}
    		\caption{The vault prototype assembled with two industrial robotic arms.}
            \label{fig:full_vault} 
        \end{figure}

\begin{figure*}[ht]
            \centering
        	\begin{subfigure}[t]{0.57\linewidth}
        		\includegraphics [trim={45cm 135cm 45cm 35cm}, clip, width=0.99\textwidth]{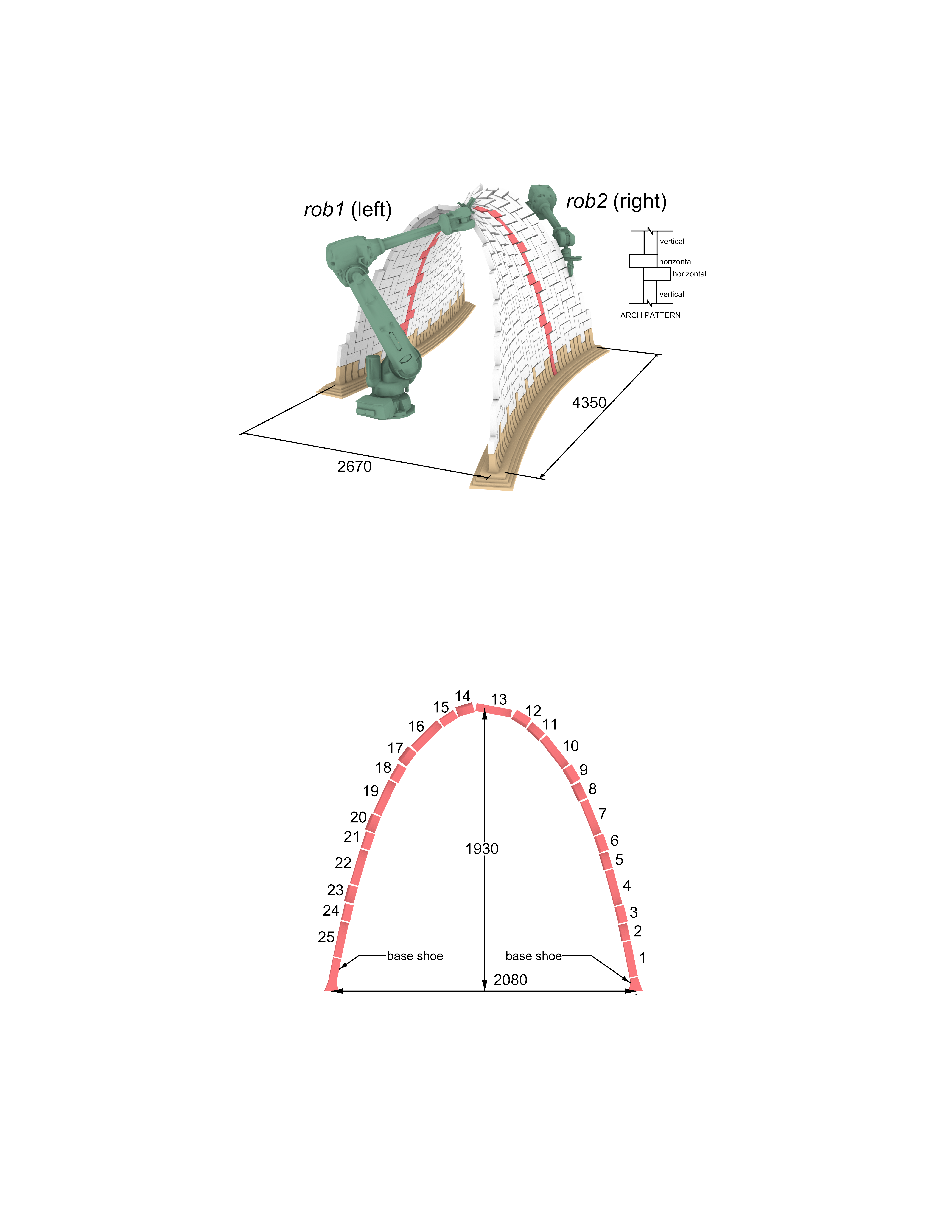}
        	\end{subfigure}
        	\begin{subfigure}[t]{0.42\linewidth}
        		\includegraphics [trim={55cm 40cm 55cm 130cm}, clip,width=0.99\textwidth]{midarch}
        	\end{subfigure}
        	\caption{Geometry of the vault prototype and central arch (dimensions in mm).}
            \label{fig:arch_vault} 
        \end{figure*}

        In this paper we explore cooperative robotic building sequences in the context of a recently constructed masonry vault prototype \citep{parascho_lightvault_2021}. \Cref{fig:arch_vault,,fig:full_vault} show the 338-brick building-scale vault, which is tiled using a herringbone tesselation pattern. This structure is assembled using two ABB 6400 industrial robotic arms (referred to as rob1 and rob2 throughout the paper), with a maximum reach of 2.55 m and payload of 40 kg. These robots have a position repetition accuracy of 0.4 mm, with 98\% of movements within 1 mm \citep{abb_product_2020}. Details of the full project and the development of the robotic fabrication are described in recent work \cite{han_concept_2020, parascho_robotic_2020, parascho_lightvault_2021}. 
        
\begin{figure}[ht]
            \centering
        	\includegraphics [trim={110cm 173cm 26cm 25cm}, clip, width=0.90\linewidth]{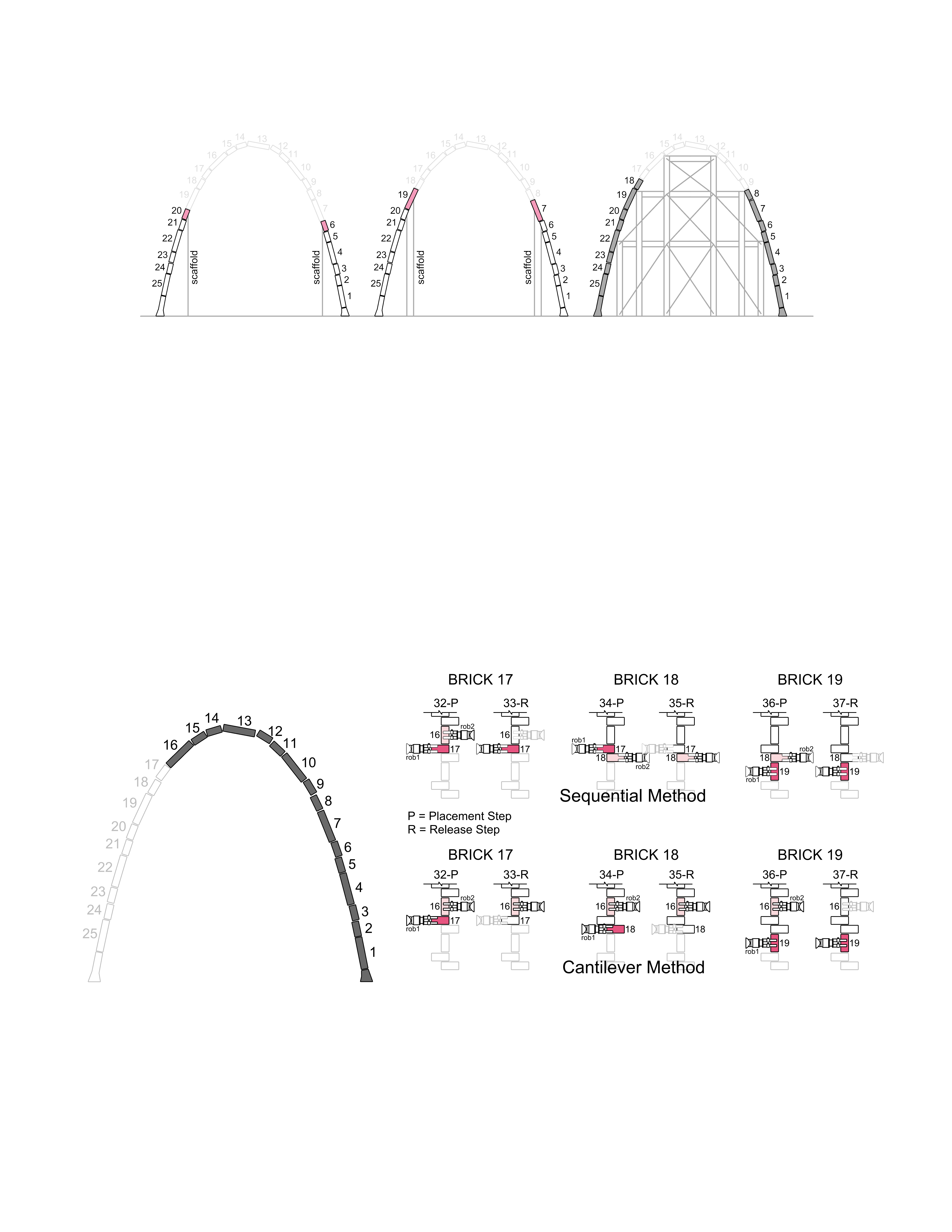}
            \caption{Scaffolding in a traditional two-sided masonry arch construction sequence.}
            \label{fig:2robot_traditional}
        \end{figure}

        The construction of this prototype can be split up into two main phases: (1) the central arch, (2) the full vault. The central arch consists of 25 rectangular bricks, arranged with a span/height of 2.08/1.93 m respectively as shown in \Cref{fig:arch_vault}. Each brick has standardized dimensions of $246 x 116 x 53 mm$, and is made from cast glass with a density of $2420 \: kg/m^3$ (i.e., self-weight of $3.66 kg$ per brick). This structure represents a significant step forward in the field of robotic masonry construction, breaking with the vertical layer-based construction that is demonstrated in preceding robotic construction projects discussed in \Cref{sec:2_automated_brick,,sec:2_dfab}. The first construction phase is critical when developing a scaffold-free fabrication method since the arch is not self-stable in its unfinished state. Thus the robotic fabrication methods must break from the traditional two-sided construction approach where external scaffolding is required as shown in \Cref{fig:2robot_traditional}. Instead, the arch is built up from only one end, while utilizing the robots as mobile temporary supports to the unfinished arch during the full construction sequence as outlined in \citep{parascho_robotic_2020}.
         
         Once the central arch is completed (see \Cref{fig:arch_vault}), the second phase consists of building up the rest of the vault around it over a skewed plan area of $2670 x 4350 mm$. A step-wise sequence is used, described further in \cite{parascho_robotic_2020}, where the central arch acts as a support from which to extend the structure out in overlapping layers. An interlocking herringbone tessellation pattern is used throughout the full vault to provide local support to the surrounding bricks. Therefore, the central arch is built with a horizontal-horizontal-vertical 3-brick pattern, which allows it to adequately interlock with the surrounding vault as outlined in \citep{parascho_robotic_2020}.
        
    \subsection{Computational Modeling Approach}\label{sec:comp_model_approach}
        The purpose of the computational modeling is to obtain values of displacements, forces and stability feedback throughout all the stages in the construction process for the masonry arch. The three robotic fabrication sequences proposed in \Cref{sec: fab_strategies_2rob,sec: fab_strategies_3rob} are evaluated on the basis of their structural performance. The complexity or precision of a Finite Element (FE) mesh-based or Discrete Element (DE) model is deemed unnecessary since only the linear elastic behaviour of the masonry vault is required to assess performance in the context of the construction sequence. Therefore, only the static force distribution and instantaneous displacements are used as the basis for evaluating performance (i.e., nonlinear or long-term responses are not being evaluated). Furthermore, the large number of independent models that need to be evaluated to fully characterise the solution space of a building sequence optimization problem of this scale also preclude the use of FE mesh-based or DE analysis methods. For example, the placement of each new brick constitutes a new structural model, which requires updating geometry, loading, and support conditions. Finally, a traditional modeling environment (i.e., non-parametric) is not suited to the optimization process used to calculate support points described in \Cref{sec: fab_strategies_3rob}.
        
        Based on these analysis requirements, a method based on a highly abstracted structural model using only linear elements is proposed here. First, the masonry structure is modelled as a geometric system, represented as a network graph (i.e., a collection of nodes connected by edges) as a way to simplify the discrete element assembly while still capturing its geometry and topology \citep{kaveh_role_2005, kaveh_efficient_2010}. Next, this geometric representation of the structure is turned into a simplified FE model, using linear elements and joints to represent the edges and nodes of the network. Specifically, edges become the structural elements (i.e., bricks) and nodes become either the centroids of the bricks or flexible connection joints between bricks (i.e., mortar bond) as shown in \Cref{fig:doublecross}. Generating such a model from a given set of discrete elements can be scripted in a parametric environment \citep{rutten_grasshopper_2007}, allowing for the rapid generation of numerous models representing the different construction steps, and simultaneous linear elastic analysis using existing FE software \citep{preisinger_karambatoolkit_2014}. Linear and rotational stiffness values can be assigned to the elements and joints respectively. Rigid elements and flexible joints are used to represent the deformations that occur in the masonry assembly analysed in the linear range.
        
        With such a model, based on a network representation, one compromises in accuracy but gains far more in "explorativity" by virtue of its simplicity and application in a parametric environment. \Cref{sec:double_cross} will describe this approach in more detail as it is specifically applied to the analysis of the masonry vault prototype, highlighting the calibration performed to determine reasonable joint spring stiffness values.
    
    \subsubsection{The Double-Cross Model}\label{sec:double_cross}
\begin{figure}[H]
            \centering
        	\begin{subfigure}[b]{1\linewidth}
        	    \centering
        		\includegraphics [trim={25cm 135cm 40cm 40cm}, clip, width=0.70\textwidth]{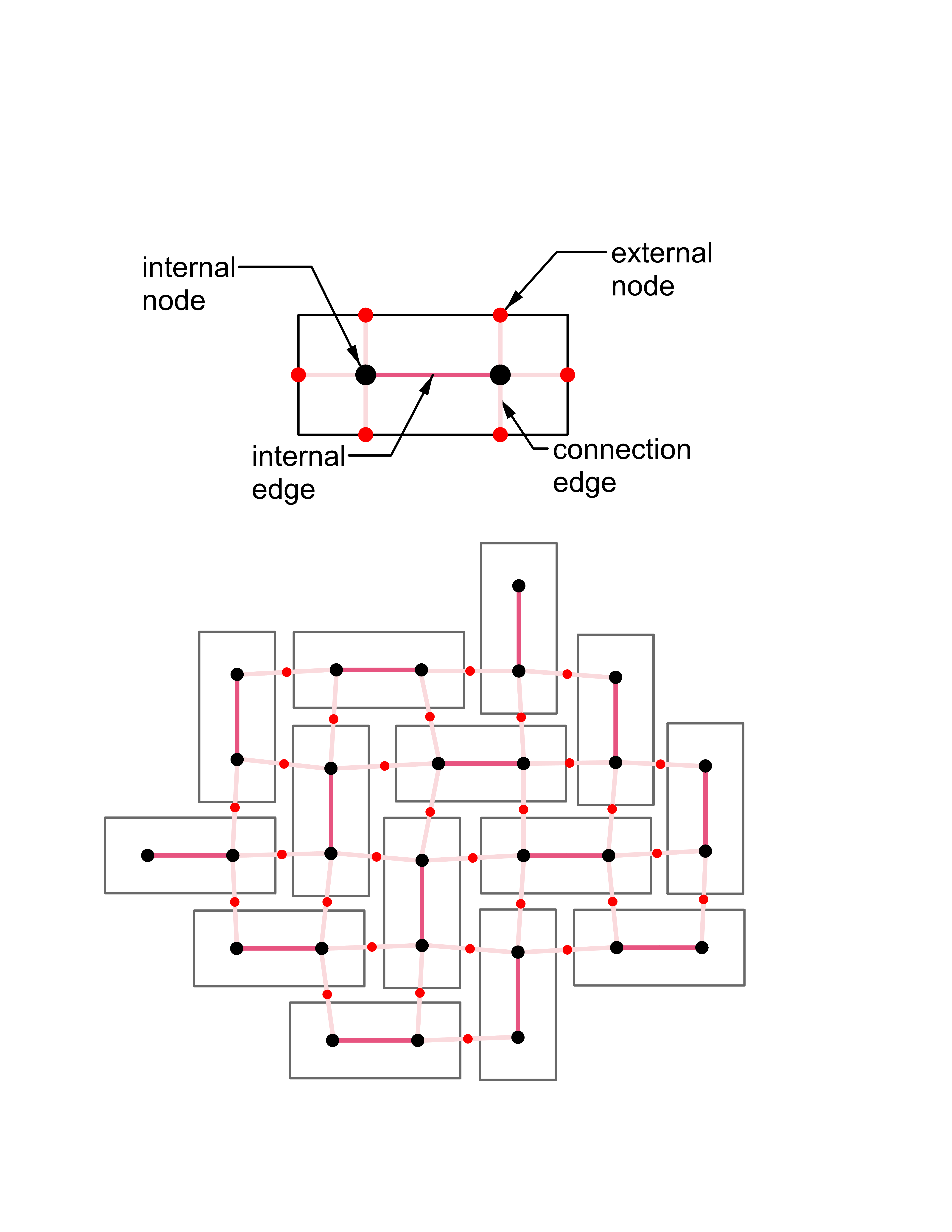}
        		\caption{components of a double-cross network model for a single brick}
        	\end{subfigure}
            
        	\begin{subfigure}[b]{0.49\linewidth}
        	    \centering
        		\includegraphics [trim={0cm 20cm 53cm 120cm}, clip,width=0.99\textwidth]{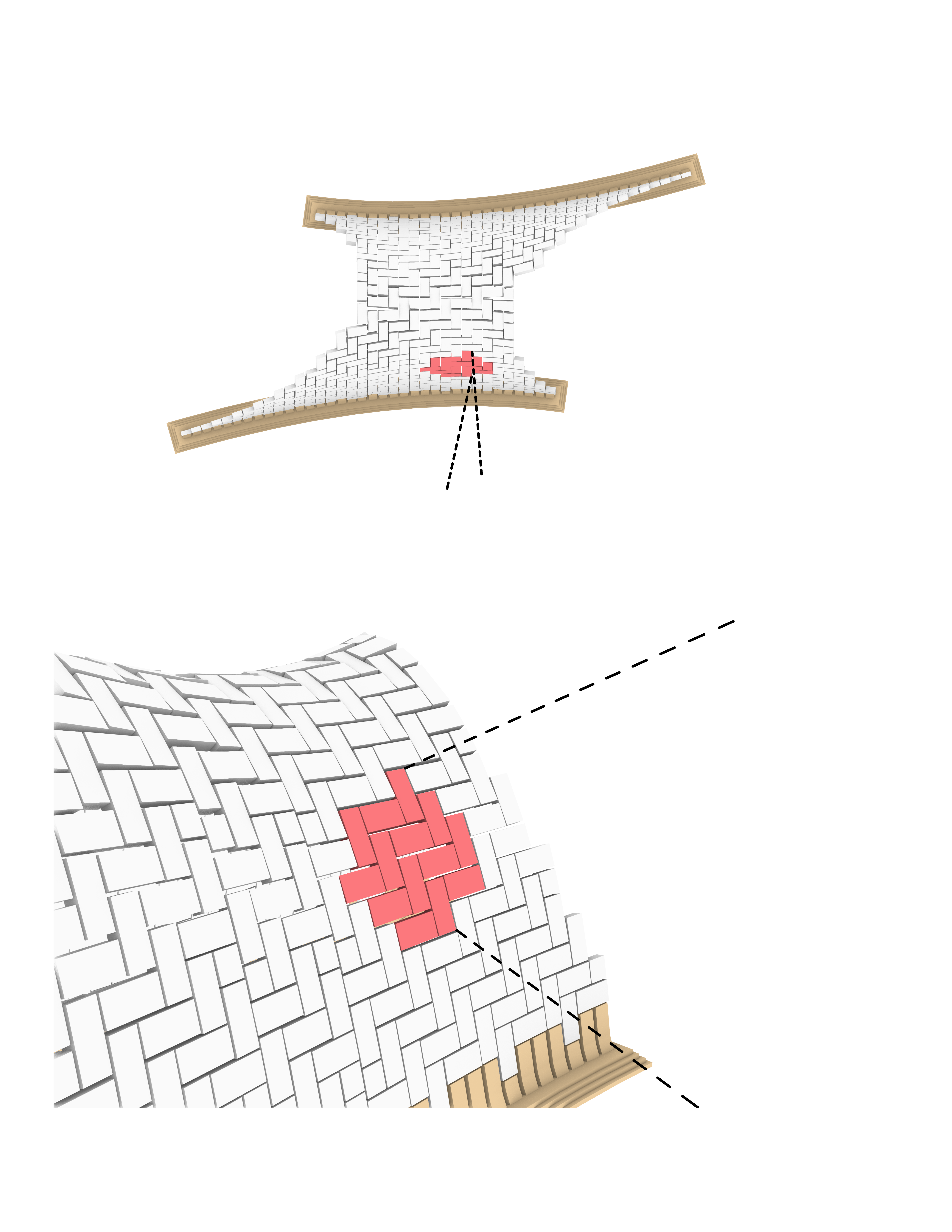}
        		\caption{subset of bricks from the vault}
        	\end{subfigure}
        	\begin{subfigure}[b]{0.50\linewidth}
        		\includegraphics [trim={18cm 25cm 25cm 97cm}, clip, width=0.99\textwidth]{double_cross}
        		\caption{network representation}
        	\end{subfigure}
        	\caption{Representing a masonry structure with the double-cross model.}
            \label{fig:doublecross} 
        \end{figure}

        To model masonry assemblies we simplify the geometry of each masonry unit to a ``double-cross" network; the name is derived from the schematic appearance of this representation shown in \Cref{fig:doublecross}. This configuration is chosen since it represents the physical geometry and connection topology in a typical masonry structure. Each brick is assigned a total of six external nodes (two per long side and one on either short side) and two inner nodes (at the quarter-points, or centroids of each half-brick). In a typical assembly, individual bricks can be joined together by connecting their neighboring centroids with a connection edge, and splitting this element to create an external node at the location where the gap between the bricks occurs (i.e., the external nodes of two connected bricks coincide). Both internal (between two internal nodes) and external (between an internal and external node) edges are modeled as rigid elements. They connect to at least one rigid internal joint, which by formulation does not allow for any relative rotation in the edges connected to it. The result is capturing the overall rigidity of a masonry unit using only simple joint and linear elements. Meanwhile, the external nodes represent the connection points between rigid bricks, and are modeled as joints with linear and rotational spring stiffness that represent the flexibility of the connection mortar. \Cref{fig:doublecross} shows such a representation for a small subset of bricks arranged in a herringbone pattern taken from the vault prototype.
        
    \subsubsection{Joint and Element Calibration }\label{sec:2_joint_calibration}
        
    	\begin{table*}[ht]
    	
        	\renewcommand{\arraystretch}{0.9}
        	\small
        	\centering
        	\caption{Summary of structural parameters for full-scale masonry arch tests with published load-deflection data}
        	\vspace{-2.5mm}
    	
        \begin{threeparttable}
    		\begin{tabular}{ll c rrrrc c rlll}
    			\specialrule{.10em}{0.2em}{.2em}
    			\centering
    			&\multirow{3}{*}{\normalsize{Reference}}	
    			&\phantom{a}%
    			&\multicolumn{1}{c}{\normalsize{$L$}}
    			&\multicolumn{1}{c}{\normalsize{$r$}} 
    			&\multicolumn{1}{c}{\normalsize{$b$}} 
    			&\multicolumn{1}{c}{\normalsize{$t$}} 
    			&\multicolumn{1}{c}{\normalsize{$\#_{bricks}$}}
    			&\phantom{\makecell{\vspace{0.5em}}}
    			&\multicolumn{1}{c}{\normalsize{$h$}}
    			&\multicolumn{1}{c}{\normalsize{$\gamma_{f}$}} 
    			&\multicolumn{1}{c}{\normalsize{$\gamma_{m}$}} 
    			&\multicolumn{1}{c}{\normalsize{$x$}} 
    			\\	
    			[0.8ex]
    			\cmidrule{4-8} \cmidrule{10-13}
    			&\phantom{a}	
    			&\phantom{a}
    			&\multicolumn{1}{c}{$\left[m\right]$}
    			&\multicolumn{1}{c}{$\left[m\right]$} 
    			&\multicolumn{1}{c}{$\left[m\right]$} 
    			&\multicolumn{1}{c}{$\left[mm\right]$} 
    			&\phantom{a}%
    			&\phantom{a}%
    			&\multicolumn{1}{c}{$\left[mm\right]$}
    			&\multicolumn{1}{c}{$\left[\frac{kg}{m^3}\right]$}
    			&\multicolumn{1}{c}{$\left[\frac{kg}{m^3}\right]$}
    			&\multicolumn{1}{c}{$\left[m\right]$}
    			\\
    			\specialrule{0.06em}{0.2em}{.2em}
    $1$  & Royles and Hendry (1991) \cite{royles_model_1991} & &2.180 &1.090 &1.690 &103 &46\tnote{1}  & &240 &1430 &2000\tnote{1}  &0.750\\
    $2$  & Melbourne and Walker (1988) \cite{melbourne_load_1988} & &1.600 &0.800 &1.000 &100 &31 & &150 &1560 &2100 &0.550\\
    $3$  & Melbourne et al. (1997) \cite{melbourne_collapse_1997} & &3.220 &0.860 &2.880 &215 &48\tnote{1}  & &170 &2260 &2400 &0.860\\
    $4$  & Melbourne and Gilbert (1995) \cite{melbourne_behaviour_1995} & &3.220 &0.860 &2.880 &215 &48\tnote{1}  & &300 &2260 &2400 &0.860\\
    $5$  & Melbourne and Gilbert (1995) \cite{melbourne_behaviour_1995} & &5.450 &1.470 &3.010 &445 &81\tnote{1}  & &350 &2260 &2400 &1.470\\
    $6$  & Gilbert et al. (2007) \cite{gilbert_small_2007} & &3.220 &0.860 &1.010 &215 &48\tnote{1}  & &305 &1950 &2360 &0.860\\
    $7$  & Towler and Sawko (1982) \cite{towler_limit_1982} & &4.220 &1.110 &1.100 &215 &70\tnote{2} & &250 &2000\tnote{2} &2000\tnote{2} &2.110\\
    $8$  & Swift et al. (2013) \cite{swift_physical_2013} & &3.220 &0.860 &1.010 &215 &48 & &300 &2040 &2000\tnote{2} &0.860\\
    			\specialrule{0.10em}{0.2em}{.2em}
    		\end{tabular}
            \begin{tablenotes}
              \scriptsize
              \item[1] data not found in original paper, taken from summary by \cite{kaminski_tests_2010}
              \item[2] data not found in original paper, estimated
            \end{tablenotes}
          \end{threeparttable}
          
    	\label{table:exp_data}
    	\end{table*}
    
        Despite being a simplification, the double-cross representation must still be able to capture displacements and force redistribution in a realistic manner. We use structural stiffness as a proxy for this -- if an analyzed structure exhibits accurate stiffness (i.e., load/deflection), then the assumption is that deflections and force redistribution are also accurate in the structure \citep{turner_stiffness_1956}.
    
\begin{figure}[H]
            \centering
        	\includegraphics [trim={10cm 40cm 20cm 140cm}, clip, width=0.99\linewidth]{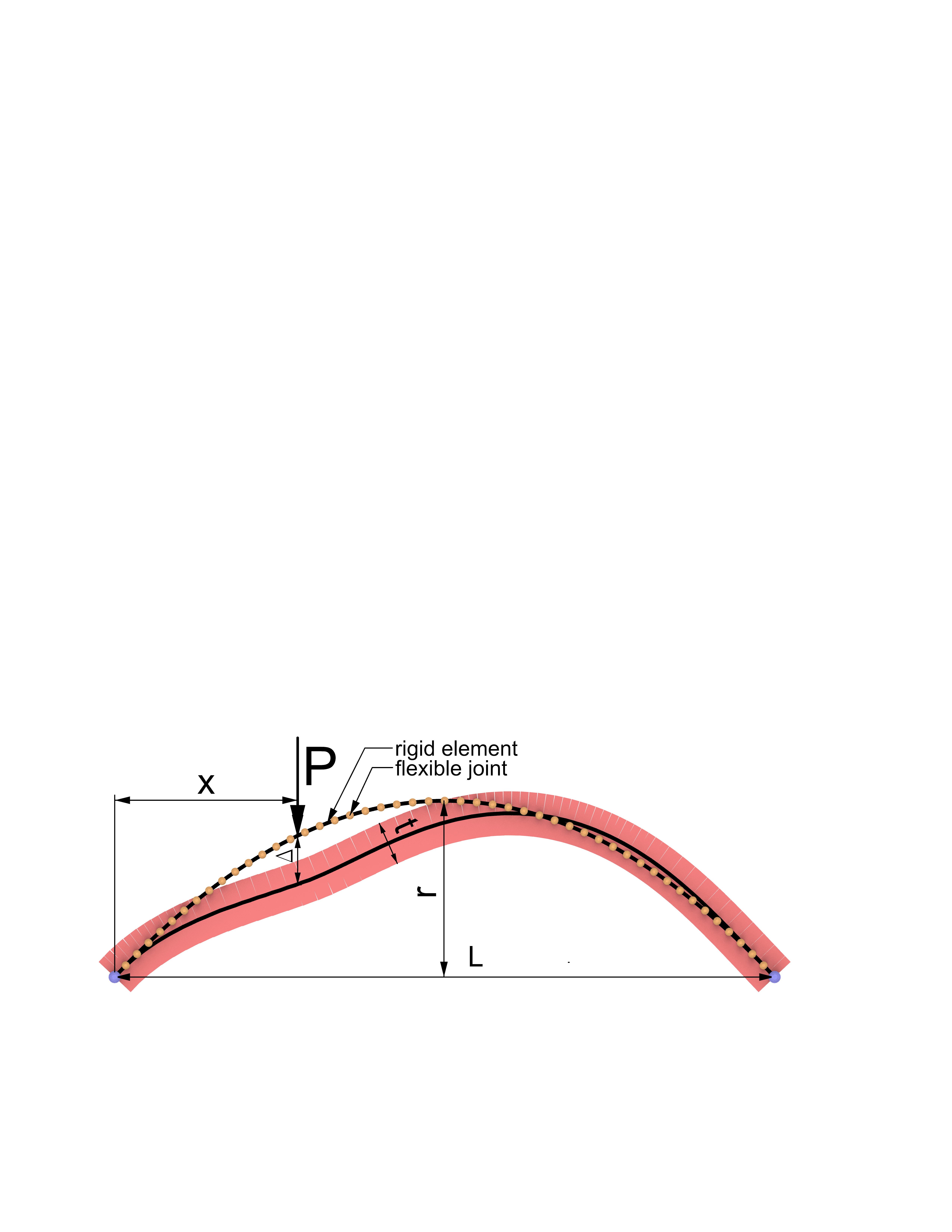}
            \caption{Deformed shape superimposed on the analytical representation of a generic masonry barrel vault load test.}
            \label{fig:arch_test}
        \end{figure}

        To improve the results from the double-cross model, the linear and rotational springs in the connection joints are calibrated on the basis of existing experimental results from full-scale masonry barrel vault load tests documented in the literature \citep{royles_model_1991,melbourne_load_1988,melbourne_collapse_1997,melbourne_behaviour_1995,gilbert_small_2007,towler_limit_1982,swift_physical_2013}. The reader is referred to \Cref{table:exp_data} for a list of references and values for all parameters defining the different structures: L = span length, r = radius, b = width (into the page), t = arch thickness, $\#_{bricks}$ = number of bricks in the arch, h = infill height above the crown, $\gamma_f$ = infill density, $\gamma_m$ = masonry density, x = loading distance. A generic experimental setup is schematically shown in \Cref{fig:arch_test}. These experiments take a monotonically increasing load (P) and apply it across the width (b) of the barrel vault, measuring the displacement under the load ($\Delta$). For calibration purposes, we only use results in the literature where experimental load-deflection data is provided, and where spandrels are detached from the arch structure. 
    
\begin{figure}[H]
            \centering
        	\includegraphics [trim={0cm 0cm 0cm 0cm}, clip, width=0.99\linewidth]{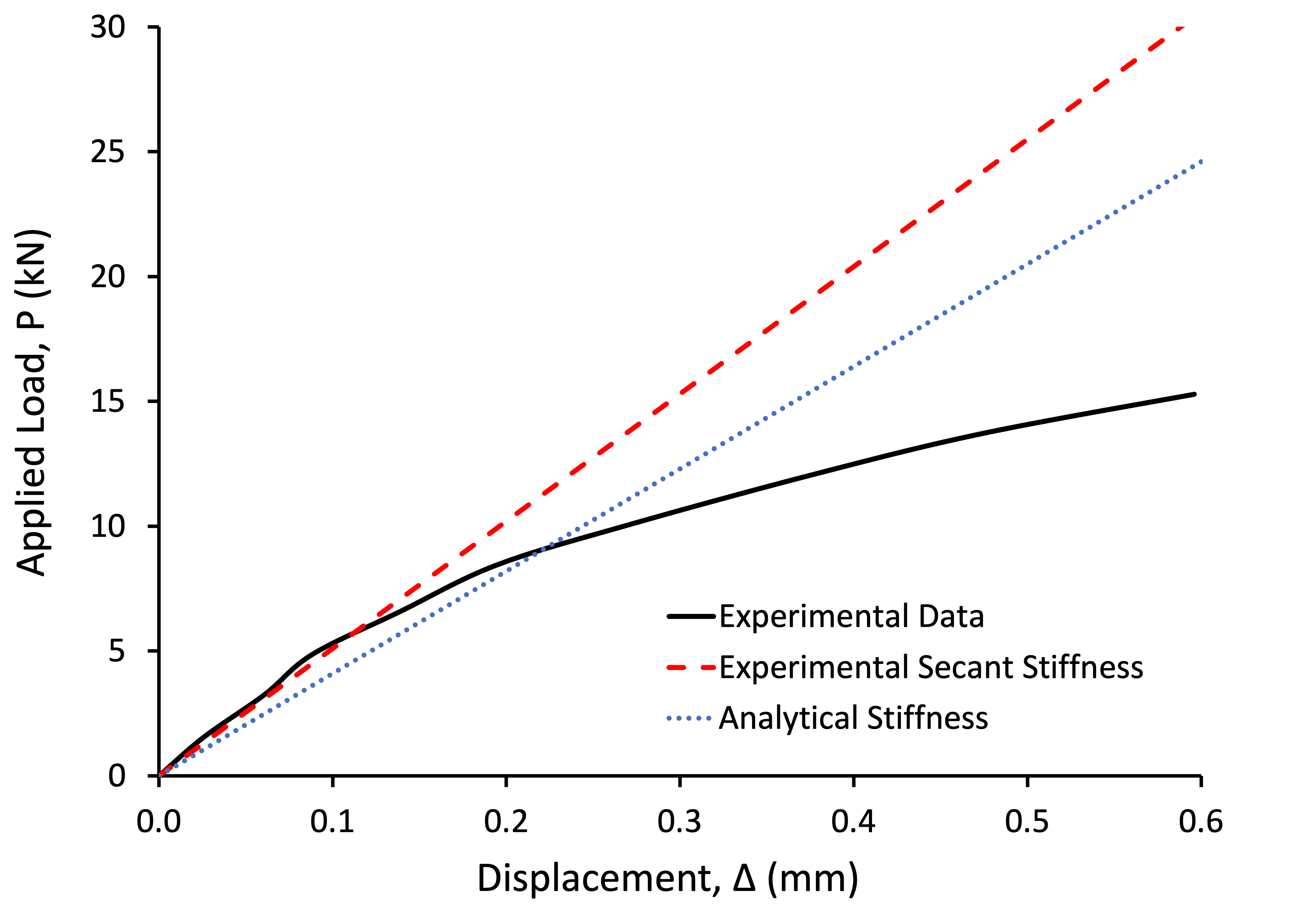}
            \caption{Comparison of the initial experimental and analytical secant stiffness.}
            \label{fig:secant_stiff}
        \end{figure}

        Linear elastic FE analysis models are created for each of the masonry vaults summarized in  \Cref{table:exp_data}. These models are based on the same approach described in the \Cref{sec:comp_model_approach}: each brick in the arch is modeled as a rigid element connected to its neighbor at a flexible joint as shown in \Cref{fig:arch_test}. The full width (into the page) of the arch is represented by a single element since only in-plane behavior is examined in these tests. From this linear elastic analysis, the stiffness ($k_m$) is obtained as a ratio of the applied load to the vertical displacement at the loaded location. This analytical stiffness is then compared to the experimental secant stiffness ($k_e$) in the initial linear elastic region, which is based on the published load-displacement test data. \Cref{fig:secant_stiff} shows an example of this process using the data from Test \#1 \citep{royles_model_1991}. The rotational and linear spring stiffness of the joints are then calibrated to ensure the best agreement between experimental and analytical stiffness for all the available tests.
    
        The results of the calibration are summarized in \Cref{table:model_results}, which also shows the relative error between the experimental ($k_e$) and analytical ($k_m$) stiffness when using calibrated linear and rotational joint springs in the analytical model. When compared to the experimental data, a calibrated linear spring stiffness of $100 \times 10^6 \: kN/m^3$ and rotational spring stiffness of $0.9 \times 10^6 \: kN/rad \cdot m^2$ are determined to minimize the error across the full data set. The linear spring stiffness falls within the $100 \times 10^6 - 100 \times 10^7 \: kN/m^3$ range as per recommendations \citep{simon_discrete_2016}. The rigid element and flexible joint approach used in the double-cross model is thus capable of capturing the in-plane load-deflection behaviour of a masonry structure in the elastic range. 
    
    	\begin{table}[ht]
        	\renewcommand{\arraystretch}{1.0}
        	\small
        	\centering
        	\caption{Initial experimental secant arch stiffness compared to calibrated element-spring model stiffness}
            \vspace{-2.5mm}
        
    		\begin{tabular}{cl rrr}
    			\specialrule{.10em}{0.2em}{.2em}
    			\centering

    			&\multirow{3}{*}{\normalsize{Reference}}
    			&\multicolumn{1}{c}{\normalsize{$k_{e}$}}
    			&\multicolumn{1}{c}{\normalsize{$k_{m}$}}
    			&\multicolumn{1}{c}{\normalsize{$e_{rel}$}}
    			\\	
    			[0.8ex]
    			\cmidrule{3-5}
    			&\phantom{a}
    			&\multicolumn{1}{c}{$\left[\frac{kN}{mm}\right]$}
    			&\multicolumn{1}{c}{$\left[\frac{kN}{mm}\right]$}
    			&\multicolumn{1}{c}{$\left[\%\right]$}
    			\\
    			\specialrule{0.06em}{0.2em}{.2em}
                $1$ & Royles and Hendry (1991) \citep{royles_model_1991}  &51 &41 &-19.6\\
                $2$ & Melbourne and Walker (1988) \citep{melbourne_load_1988} &138 &143 &3.6\\
                $3$ & Melbourne et al. (1997) \citep{melbourne_collapse_1997} &378 &434 & 14.8\\
                $4$ & Melbourne and Gilbert (1995) \citep{melbourne_behaviour_1995} &535 &417 & -22.1 \\
                $5$ & Melbourne and Gilbert (1995) \citep{melbourne_behaviour_1995} &803 &264 & -67.1\\
                $6$ & Gilbert et al. (2007) \citep{gilbert_small_2007} &132 &140 &6.1\\
                $7$ & Towler and Sawko (1982) \cite{towler_limit_1982}  &70 &71 &1.4\\
                $8$ & Swift et al. (2013) \cite{swift_physical_2013} &120 &141 &17.5\\
    			\specialrule{0.10em}{0.2em}{.2em}
    		\end{tabular}
    			
    	\label{table:model_results}
    	\end{table}

	\subsubsection{3D Validation of Model}
	    Further validation of the proposed model is performed to verify the ability of the double-cross model to capture out-of-plane displacements when using the joint stiffness results determined in \Cref{sec:2_joint_calibration}. This validation is performed against static DE models implemented using the 3DEC software \citep{itasca_consulting_group_inc_3dec_2016}, representing snapshots of the prototype arch during various stages of construction. These models are numerical approximations of the real behavior, but by formulation are more realistic and accurate and computationally more expensive. Thus, the goal is to verify that the results of the DE models show good agreement with those of the simplified double-cross representation for situations that cause out-of-plane effects in the arch.
	    
	    In the DE model, the partial arch is modelled as an assembly of rigid bodies, whose shape corresponds to the masonry bricks (i.e., rectangular prisms with 8 vertices). The joints between the bricks are modelled by interfaces ruled by a Mohr-Coulomb model \citep{simon_discrete_2016}, with the parameters based on the characteristics of the epoxy mortar used (i.e., Oatey\textsuperscript{\textregistered} Fix-It\texttrademark\  Stick \citep{oatey_oatey_2020}). The cohesion used is 2 MPa/m and the tensile stress cut-off is 3 MPa. Joint stiffness and shear stiffness are set to $jK_n = 100 \times 10^6 \: kN/m^3$ and $jK_s = 10 \times 10^6 \: kN/m^3$ with a friction angle of 25$^{\circ}$ \citep{forgacs_minimum_2017}.
	    
\begin{figure}[H]
            \centering
    	    \includegraphics [trim={0cm 0.2cm 0cm 0.2cm}, clip, width=0.85\linewidth]{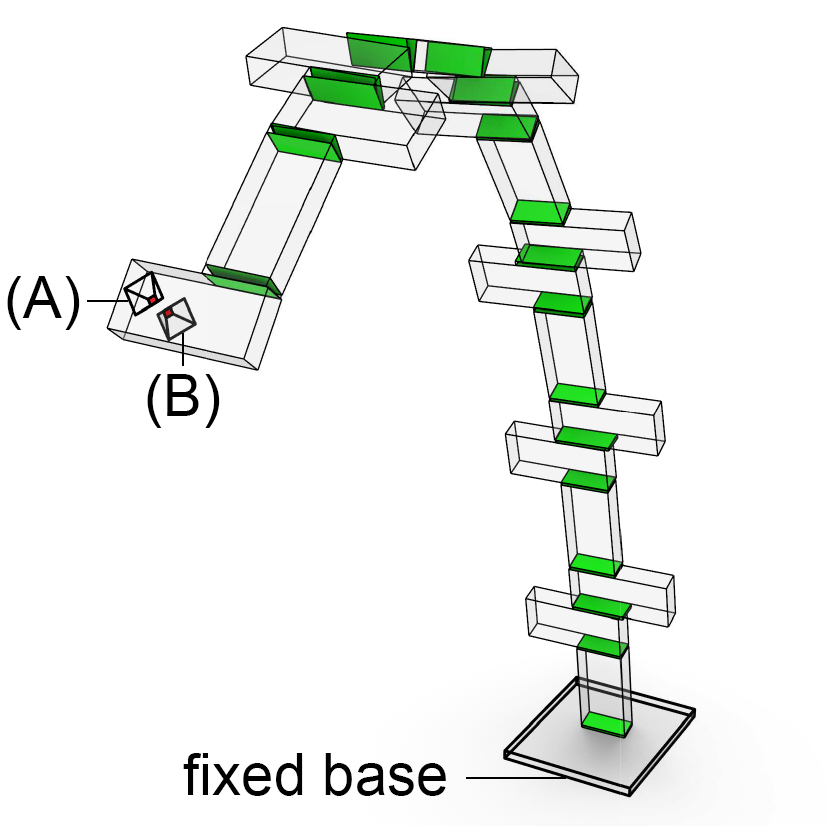}
    	    \caption{3DEC model where robotic support is modelled as a two-block interface restraint (A and B).}
            \label{fig:3dec_image}
        \end{figure}

	    For the construction stages investigated, the robotic arm acting as a support to the structure is simulated with two fixed rigid bodies. These are labelled at (A) and (B) in \Cref{fig:3dec_image}, which otherwise shows a visualization of the discrete element model with the brick interfaces highlighted. These two rigid bodies effectively support the active block in the arch at two small interfaces to simulate how a robot grips the brick. Unlike the typical brick-brick mortar joints, at the pinch location the normal and shear stiffness values are an order of magnitude greater to mimic the effect of a robotic gripper support, thus preventing any sliding deformation. Therefore, the arch is vertically supported while still able to rotate about the axis formed by the rigid support blocks. Global out-of-plane displacements of the arch are determined as a function of this support condition.
	    
\begin{figure}[H]
            \centering
            \begin{subfigure}[t]{0.42\linewidth}
    	        \includegraphics [trim={0.1cm 0.2cm 4.8cm 1.3cm}, clip, width=0.99\linewidth]{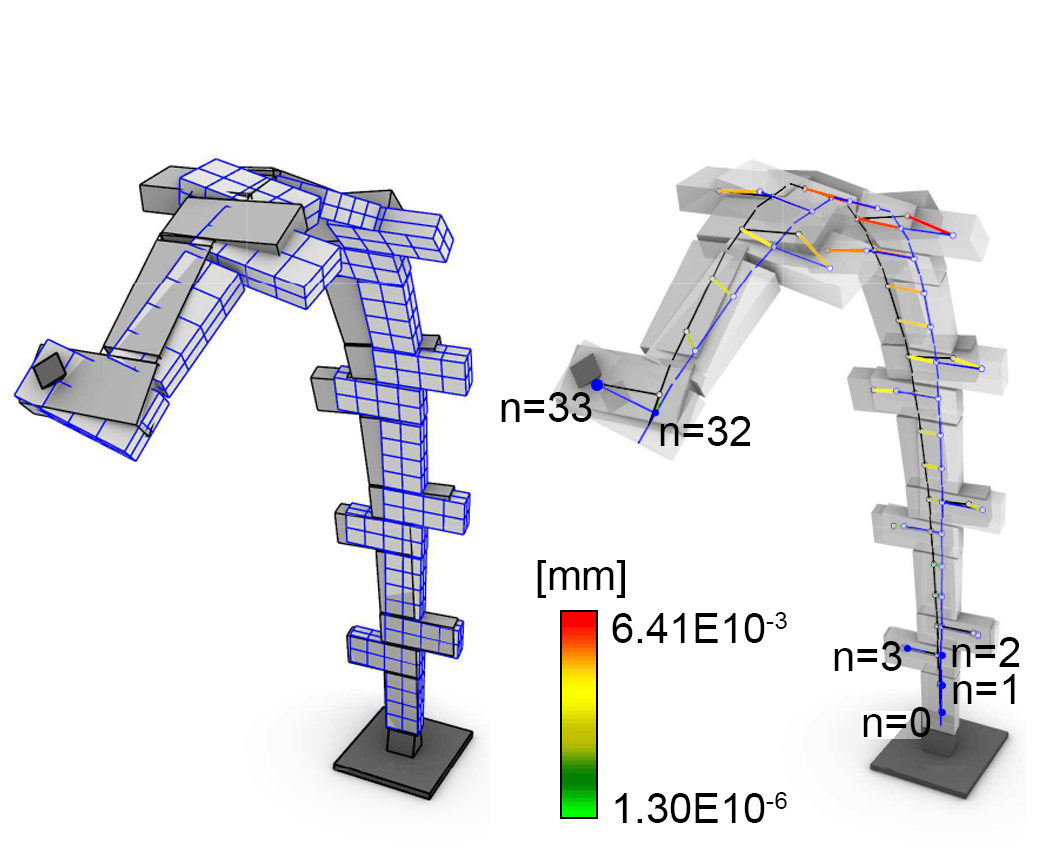}
                \caption{displacements of the rigid blocks}
            \end{subfigure}   
            \begin{subfigure}[t]{0.49\linewidth}
    	        \includegraphics [trim={4.1cm 0.2cm 0.1cm 1.3cm}, clip, width=0.99\linewidth]{3DEC_disp_nodes}
    	        \caption{equivalent nodal displacements}
            \end{subfigure}   
            
            \caption{DE analysis of the partial arch.}
            \label{fig:3dec_image2}
        \end{figure}

	    The validation is performed across four separate models corresponding to steps in the arch construction sequence where the robot places and supports bricks 17, 18, 23 and 24. These construction steps are expected to produce out-of-plane twisting as seen in the \textit{sequential method} discussed further in \Cref{sec: fab_strategies_2rob}. All DE analyses are performed with self-weight loading and supports at both the location where the robotic arm holds a brick and the fixed base. The DE computations are carried out over a sufficient time scale for the structure to reach equilibrium (i.e., until the unbalanced force is 0 kN), and in that state the displacements at the vertices of the discrete blocks are output. These rigid block vertex displacements are then converted to equivalent displacements at nodal locations coinciding with the inner nodes in the double-cross representation (see \Cref{fig:3dec_image2}). This conversion is done based on weighted averages of the corner displacements as the blocks are rigid and therefore experience no relative deformation between vertices.
	    
        The results of this comparison are shown in \Cref{table:validation_results}, where the average relative error in the displacements between nodes across the discrete and finite element models is used as the comparison metric for each of the four partial arches modeled. This average error is based on results ranging from 33 to 47 data points for model 1 to 4 respectively. The magnitude of the errors is at a similar order of magnitude to those obtained in the in-plane calibration (see \Cref{sec:2_joint_calibration}), which means that the results for the out-of-plane behaviour are satisfactory using the simplified formulation. A comparison between the experimental calibration and DE validation tests show that when calibrated the calibrated double-cross model is capable of representing both the in-plane and out-of-plane behaviour (i.e., within a ~20\% margin) of a masonry structure.
        
    	\begin{table}[ht]
        	\renewcommand{\arraystretch}{1.0}
        	\centering
        	\caption{Average relative error in displacements across models}
        	\vspace{-2.5mm}
        	
    		\begin{tabular}{cccr}
    			\specialrule{.10em}{0.2em}{.2em}
    			\multicolumn{1}{c}{\normalsize Model}
    			&\multicolumn{1}{c}{\normalsize Bricks}
    			&\multicolumn{1}{c}{\normalsize $\#_{nodes}$}
    			&\multicolumn{1}{c}{\normalsize $\%_{error}$}
    			\\
    			\specialrule{0.06em}{0.2em}{.2em}
                $1$ &17 &33 &20.9\\
                $2$ &18 &35 &24.7\\
                $3$ &23 &45 &-14.9\\
                $4$ &24 &47 &-15.3\\
    			\specialrule{0.10em}{0.2em}{.2em}
    		\end{tabular}
    		
    	\label{table:validation_results}
    	\end{table}

%% file: sections/4twoRob.tex

\begin{figure*}[b!]
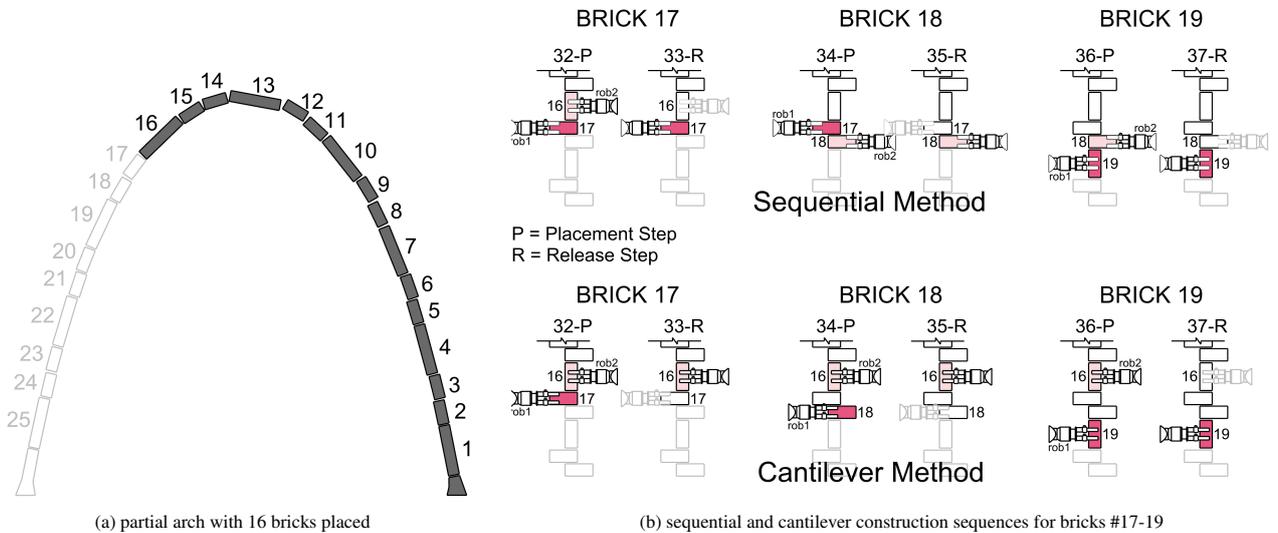

    \begin{subfigure}[t]{0.39\linewidth}
        \includegraphics [trim={10cm 47cm 105cm 125cm}, clip, width=1\textwidth]{sequence_2rob}
        \caption{partial arch with 16 bricks placed}
    \end{subfigure}
    \begin{subfigure}[t]{0.56\linewidth}
        \includegraphics [trim={77cm 47cm 10cm 125cm}, clip, width=1\textwidth]{sequence_2rob}
        \caption{sequential and cantilever construction sequences for bricks \#17-19}
        \label{fig:two_robot_methods}
    \end{subfigure}
    \caption{Unidirectional construction sequences where two robots alternate the support of the partial arch and the placement of new bricks.}
    \label{fig:2rob_methods}
\end{figure*}

\section{Central Arch Fabrication with Two Robots}\label{sec: fab_strategies_2rob}
    The first phase of the vault's construction (i.e., the central arch) represents a challenge from the perspective of a scaffold-free construction method, since before the central arch is complete the structure is not self-stable. To unlock the construction potential of using two robots, we build on an existing cooperative robotic fabrication strategy first introduced in \cite{parascho_robotic_2020}. A unidirectional construction approach is implemented, where the arch is built up from one end to the other. Contrary to the traditional method, discussed in \Cref{sec:2_prototype}, there is only one active build edge. This means the robots can be sequenced in a cooperative manner by taking turns either placing or supporting the partial arch at this active edge. In this way, the full arch can be constructed without any external scaffolding by relying on robotic support.
    
    Two different build strategies, referred to as the \textit{sequential method} or \textit{cantilever method} are defined in \cite{parascho_robotic_2020}. \Cref{fig:2rob_methods} schematically shows an example of these methods at work when when adding three bricks to the partial arch starting from brick \#17. In both sequences, each build step consists of one robot performing either of two actions: placement of a brick (P), or release of a brick (R) (i.e., adding a new brick takes two discrete steps). The strengths and weaknesses of each method are discussed in \Cref{sec:sequential,,sec:cantilever}, with detailed structural results for each build step when constructing the full 25-brick central arch using each sequence in Appendix A.

    \subsection{Sequential Method} \label{sec:sequential}
\begin{figure}[H]
            \centering
            \begin{subfigure}[b]{0.49\linewidth}
                \centering
        		\includegraphics [angle=0,trim={60cm 160cm 60cm 18cm}, clip,width=0.99\linewidth]{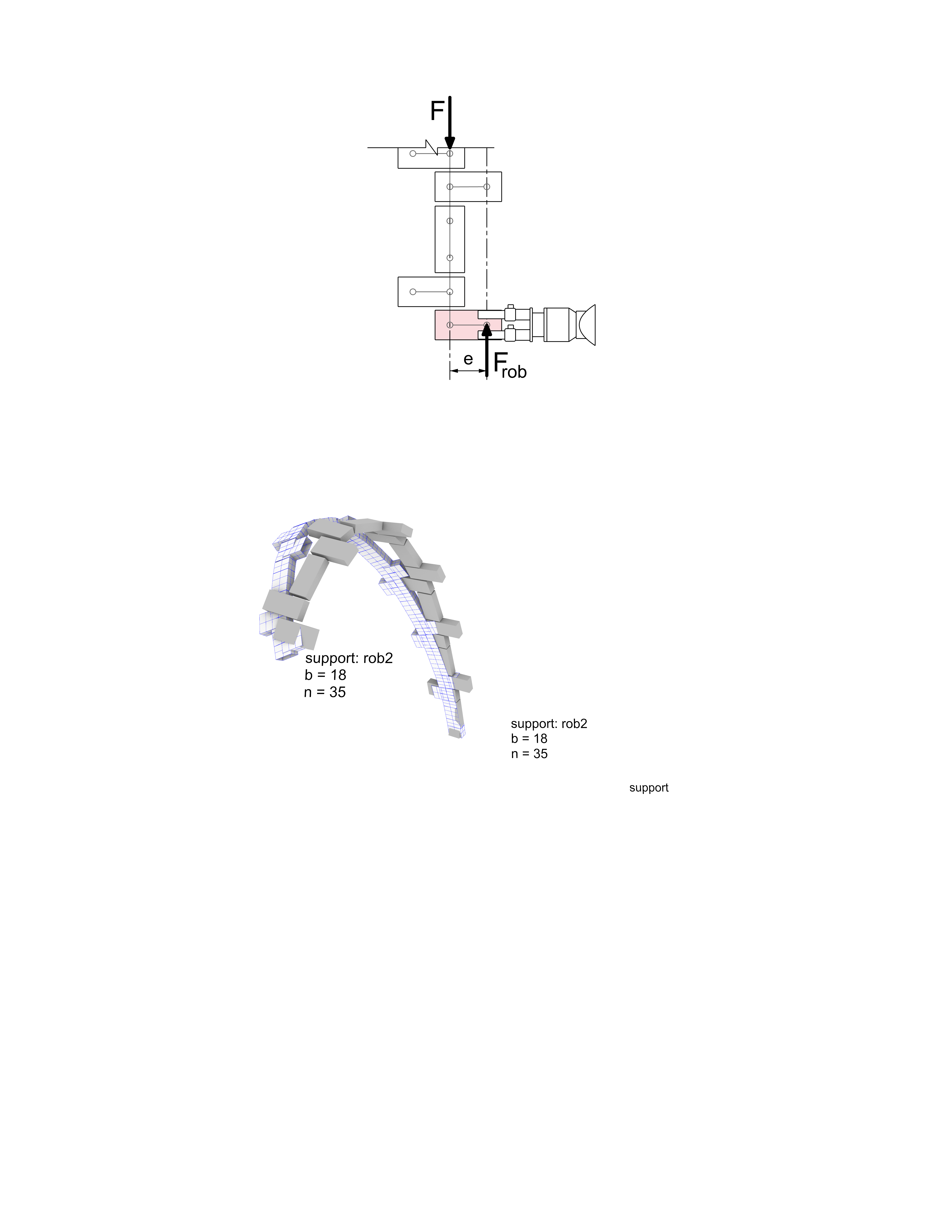}
        		\caption{Offset robotic support location}
        		\label{fig:sequential_twist_offcenter}
        	\end{subfigure}
        	\begin{subfigure}[b]{0.49\linewidth}
        	    \centering
        		\includegraphics [angle=0,trim={48cm 92cm 85cm 98cm}, clip,width=0.99\linewidth]{twist}
        		\caption{Displaced shape for Step \#35}
        		\label{fig:sequential_twist_35}
        	\end{subfigure}
            \caption{Twisting of the arch during construction with the sequential method.}
            \label{fig:sequential_twist}
        \end{figure}

        The \textit{sequential method} (shown in the top of \Cref{fig:two_robot_methods}), where the left and right robots alternate placing bricks, is the intuitive way to sequence two robots. This is a relatively simple sequence to implement robotically, but there are structural issues that arise from the support position location as function of the herringbone tessellation pattern. The bricks in the arch must be placed in a horizontal-horizontal-vertical pattern (as shown in \Cref{fig:arch_vault,fig:sequential_twist_offcenter}) to allow the arch to later interlock with the rest of the vault. When using two robots (i.e., rob1 and rob2) on either side of the arch, there are stages in the construction when there is only one robot supporting the structure (i.e., a release (R) step) as the other robot releases the structure to retrieve the next brick. At these stages there is only one robot supporting the arch while gripping a horizontal brick. \Cref{fig:sequential_twist} illustrates this condition and the force couple, $M = F \cdot e$, that results from such a situation, causing critical out-of-plane twisting displacement that the arch is not capable of resisting. This moment scales with the magnitude of the force (F), which is equal to the thrust exerted by the partial arch on the robot as a function of the geometry and self-weight distribution in the structure. The lever arm (e) is equal to the deviation between the thrust line and support point, which is $135 \pm 15 mm$.
        
        This off-center loading condition occurs at eight different steps in the full construction sequence (i.e., steps 9, 11, 21, 23, 33, 35, 45, and 47). \Cref{table:sequential_twist} summarizes the main structural results from these steps, showing the following: sequence step number; type of step (release or placement); total bricks in the arch; what brick (b) and node (n) each robot is gripping; $F_{rob}$ = load supported by the robot; $F_{sup}$ = load at the base support; $M_{sup}$ = moment at the base support; $\Delta_{max}$ = maximum nodal displacement. $C_1 = M_{sup}/M_{sup,avg}$ is the moment multiple based on an average support moment of $6.8 \: N \cdot m$ over the full sequence, and $C_2 = \Delta_{max}/\Delta_{max,avg}$ is the displacement multiple based on an average maximum nodal displacement of $7.3 \times 10^{-3} \: mm$ over the full sequence. The full data for all 49 steps required to build the 25-brick arch with the \textit{sequential method} can be found in Appendix A, \Cref{table:experimental_data1}.
        
       	\begin{table*}[ht]
    	\small
    	\renewcommand{\arraystretch}{1.0}
    	\centering
    	\caption{Analysis results for the off-center support steps in the sequential method}
    	\vspace{-2.5mm}
            
            \begin{threeparttable}
    		\begin{tabular}{c ccc r cccc r cccScr SS}
    			\specialrule{.10em}{0.2em}{.2em}
    			&\multirow{2}{*}{\normalsize  Step}
    			&\multirow{2}{*}{\normalsize  Type}
    			&\multirow{2}{*}{\normalsize \makecell{Total \\ Bricks}}
    			&\phantom{\makecell{\vspace{0.5em}}}
    			&\multicolumn{2}{c}{\normalsize $rob1$}
    			&\multicolumn{2}{c}{\normalsize $rob2$}
    			&\phantom{a}
    			&\multicolumn{1}{c}{\normalsize  $F_{rob1}$}
    			&\multicolumn{1}{c}{\normalsize  $F_{rob2}$}
    			&\multicolumn{1}{c}{\normalsize  $F_{sup}$}
    			&\multicolumn{1}{c}{\normalsize  $M_{sup}$}
    			&\multicolumn{1}{c}{\normalsize  $C_1$}
    			&\phantom{a}
    	        &\multicolumn{1}{c}{\normalsize  $\Delta_{max}$}
    	        &\multicolumn{1}{c}{\normalsize  $C_2$}
    			\\	
    			[0.8ex]
    			\cmidrule{6-9}
    			\cmidrule{11-15}
    			\cmidrule{17-18}
    			&\multicolumn{1}{c}{}
    			&\phantom{a}
    			&\multicolumn{1}{c}{}
    			&\phantom{a}
    			&\multicolumn{1}{c}{\small b}
    			&\multicolumn{1}{c}{\small n}
    			&\multicolumn{1}{c}{\small b}
    			&\multicolumn{1}{c}{\small n}
    			&\phantom{a}
    			&\multicolumn{1}{c}{\small $[N]$}
    			&\multicolumn{1}{c}{\small $[N]$}
    			&\multicolumn{1}{c}{\small $[N]$}
    			&\multicolumn{1}{c}{\small $[N \cdot m]$}
    			&\phantom{a}
    			&\phantom{a}
    			&\multicolumn{1}{c}{\scriptsize  $[10^{-3} \: mm]$}
    			&\phantom{a}
    			\\
    			\specialrule{0.06em}{0.2em}{.2em}
                &9 &release &5 &&5 &8 &-- &-- &&75.0 &-- &106.1 &6.8 &1.0 &&1.3 &0.2\\
                &11 &release &6 &&-- &-- &6 &11 &&-- &85.2 &132.2 &8.3 &1.2 &&2.2 &0.3\\
                &21 &release &11 &&11 &20 &-- &-- &&86.6 &-- &311.7 &21.5 &3.2 &&12.8 &1.8\\
                &23 &release &12 &&-- &-- &12 &23 &&-- &89.7 &347.9 &22.0 &3.2 &&16.9 &2.3\\
                \specialrule{0.06em}{0.2em}{.2em}
                &33 &release &17 &&17 &33 &-- &-- &&170.8 &-- &470.0 &21.3 &3.1 &&70.8 &9.7\\
                &35 &release &18 &&-- &-- &18 &35 &&-- &203.8 &468.9 &29.6 &4.4 &&97.5 &13.4\\
                &45 &release &23 &&23 &44 &-- &-- &&369.3 &-- &477.1 &56.4 &8.3 &&264.4 &36.2\\
                &47 &release &24 &&-- &-- &24 &47 &&-- &405.5 &474.7 &66.0 &9.7 &&306.5 &42.0\\
    			\specialrule{0.10em}{0.2em}{.2em}
    		\end{tabular}
    		\end{threeparttable}	
    		
        \label{table:sequential_twist}
    	\end{table*}   
	
        Four of the off-center steps occur while building up to the crown (i.e., steps 9, 11, 21, and 23) and four while building down from the crown (i.e.,steps 33, 35, 45, and 47). The four steps after the crown experience maximum nodal displacements at multiples of 9.7 to 42.1 times greater than the average maximum displacements for all other typical loading steps (i.e., $7.3 \times 10^{-3}$ mm), and moment multiples of 4.1 to 9.7 times the average moment across all the steps (i.e., $6.8 \: N \cdot m$). An example of such a twisting step is 35-R where $rob2$ is holding brick 18 at node 35, which is schematically shown in \Cref{fig:sequential_twist_35}. This step has a displacement and moment multiple of 13.4 and 4.4 respectively. In general, the large displacement multiples experienced during this fabrication process suggests that the sequential method while easy to implement is not robust and scalable. This method is not suitable for structures with larger axial thrusts. Ideally maximum out-of-plane displacements and moments should be kept below the value for Step 35 (i.e., $\approx100 \times 10^{-3} \: mm$ and $\approx30 \: N \cdot m$), which can be detrimental and lead to collapse of the arch. While this collapse mechanism is a function of the geometry, small and medium span arches can be fabricated without collapse using the \textit{sequential method}, but at the larger scale this twisting behavior must be avoided.
        
    \subsection{Cantilever Method} \label{sec:cantilever}
        The \textit{cantilever method} (shown in the bottom of \Cref{fig:two_robot_methods}) is a more complex 2-robot sequence that was developed to solve the twisting issue experienced in the \textit{sequential method} \citep{parascho_robotic_2020}. The method is based around an alternating three brick placement sequence for each robot. \Cref{fig:2rob_methods} illustrates one full construction loop: \textit{rob2} supports the arch at brick 16, while \textit{rob1} places 3 bricks in sequence (i.e., steps 32, 34 and 36), after which their roles swap. Despite this difference, the \textit{cantilever method} shares $1/3$ of its steps in common with the \textit{sequential method}. For example, as shown in \Cref{fig:2rob_methods}, steps 32 and 37 out of the six steps in one loop are the same across both methods. These common steps are also highlighted in the full 49 step building sequence for the 25-brick arch in Appendix A, \Cref{table:experimental_data2}. The main benefit of the \textit{cantilever method} is that the support point swap (brick 19 in step 37 in the example above) always occurs when a robot is supporting the arch on a vertical brick, which is centered on the thrust line of the arch.
        
       	\begin{table*}[ht]
    	\renewcommand{\arraystretch}{1.0}
    	\small
    	\centering
    	\caption{Comparison of cantilever to sequential method at critical steps}
    	\vspace{-2.5mm}
            
            \begin{threeparttable}
    		\begin{tabular}{c cr SSSr SSSr SSS}
    			\specialrule{.10em}{0.2em}{.2em}
    			&\multirow{4}{*}{\normalsize Step}
    			&\phantom{\makecell{\vspace{0.5em}}}
    			&\multicolumn{3}{c}{\normalsize $M_{sup}$}
    			&\phantom{\tiny a}
    			&\multicolumn{3}{c}{\normalsize $\Delta_{max}$}
    			&\phantom{\tiny a}
    			&\multicolumn{3}{c}{\normalsize $T_{max}$}
    			\\	
    			[0.8ex]
    			\cmidrule{4-6}
    			\cmidrule{8-10}
    			\cmidrule{12-14}
    			&\phantom{a}
    			&\phantom{a}
    			&\multicolumn{1}{m{2em}}{\small $C$}
    			&\multicolumn{1}{c}{\small $S$}
    			&\multicolumn{1}{c}{\small $C/S$}
    			&\phantom{a}
    			&\multicolumn{1}{c}{\small $C$}
    			&\multicolumn{1}{c}{\small $S$}
    			&\multicolumn{1}{c}{\small $C/S$}
    			&\phantom{a}
    			&\multicolumn{1}{c}{\small $C$}
    			&\multicolumn{1}{c}{\small $S$}
    			&\multicolumn{1}{c}{\small $C/S$}
    			\\
    			&\phantom{a}
    			&\phantom{a}
    			&\multicolumn{1}{c}{\scriptsize $[N \cdot m]$}
    			&\multicolumn{1}{c}{\scriptsize $[N \cdot m]$}
    			&\phantom{a}
    			&\phantom{a}
    			&\multicolumn{1}{c}{\scriptsize  $[10^{-3} \: mm]$}
    			&\multicolumn{1}{c}{\scriptsize  $[10^{-3} \: mm]$}
    			&\phantom{a}
    			&\phantom{a}
    			&\multicolumn{1}{c}{\scriptsize  $[N]$}
    			&\multicolumn{1}{c}{\scriptsize  $[N]$}
    			&\phantom{a}
    			\\
    			\specialrule{0.06em}{0.2em}{.2em}
    &	9	&&	2.9	&	6.8	&	0.43	&&	0.4	&	1.3	&	0.31	&&	68.6	&	36.9	&	1.9	\\
    &	11	&&	1.6	&	8.3	&	0.19	&&	0.6	&	2.2	&	0.27	&&	69.1	&	45.5	&	1.5	\\
    &	21	&&	20.7	&	21.5	&	0.96	&&	8.5	&	12.8	&	0.66	&&	78.4	&	23.7	&	3.3	\\
    &	23	&&	18.6	&	22.0	&	0.85	&&	7.3	&	16.9	&	0.43	&&	90.8	&	11.4	&	8.0	\\
    \specialrule{0.06em}{0.2em}{.2em}
    &	33	&&	4.6	&	21.3	&	0.22	&&	11.2	&	70.8	&	0.16	&&	29.6	&	0.8	&	37.0	\\
    &	35	&&	5.3	&	29.6	&	0.18	&&	10.4	&	97.5	&	0.11	&&	59.3	&	0.8	&	74.1	\\
    &	45	&&	3.7	&	56.4	&	0.07	&&	15.9	&	264.4	&	0.06	&&	34.8	&	0.8	&	43.5	\\
    &	47	&&	0.9	&	66.0	&	0.01	&&	11.1	&	306.5	&	0.04	&&	69.6	&	0.8	&	87.0	\\
    			\specialrule{0.10em}{0.2em}{.2em}
    		\end{tabular}
    		
            \begin{tablenotes}
                \scriptsize
                \item[] C = Cantilever Method, S = Sequential Method
            \end{tablenotes}
    		\end{threeparttable}	
    		
    	\label{table:cantilever_compare}
    	\end{table*}     
        
\begin{figure}[H]
            \centering
    	    \includegraphics [angle=0,trim={0cm 0cm 0cm 0cm},clip,width=0.99\linewidth]{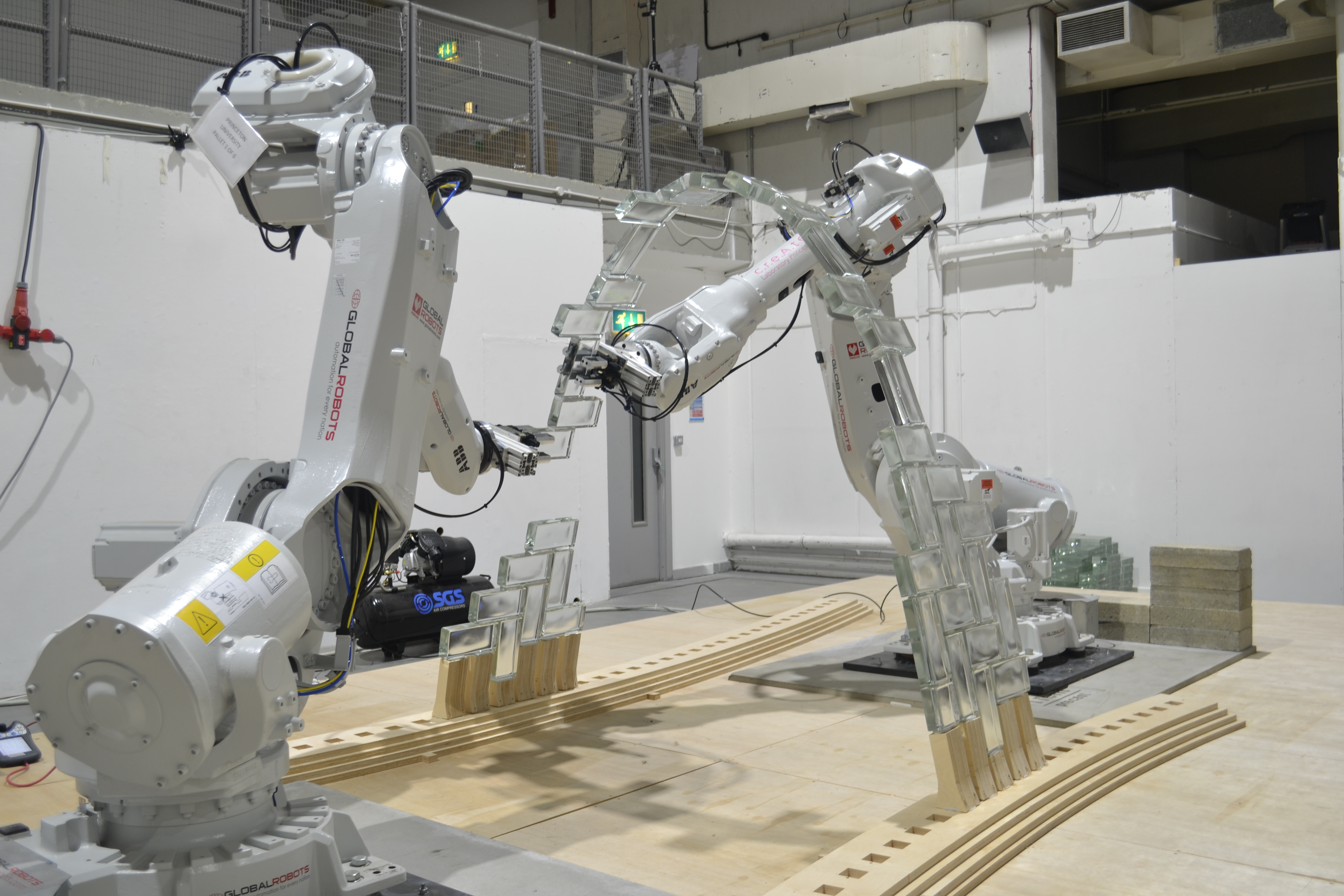}
            \caption{Two robots build the central arch using the cantilever method.}
            \label{fig:2robot_sequence1}
        \end{figure}

        Avoiding a support point swap while gripping a horizontal brick off-center leads to more consistent structural behavior across all the fabrication steps (i.e., no large jump in moments and displacements between steps). This consistency is shown in \Cref{table:cantilever_compare} when comparing the moment and maximum displacement between the two methods, for all the critical steps discussed in \Cref{sec:sequential}. Based on the moment and maximum displacement metrics, the \textit{cantilever method} consistently outperforms the \textit{sequential method}. This performance is exemplified in the four critical steps after the crown (i.e., steps 33, 35, 45 and 47), where moments and maximum displacements are reduced by 78-99\% and 84-96\% respectively. Eliminating the twist allows the central arch to maintain stability and to be constructed without temporary scaffolding, as shown in \Cref{fig:2robot_sequence1}.

        While the \textit{cantilever method} successfully mitigates the critical twisting action, it does introduce a new challenge: during certain stages in the sequence there are up to three bricks being cantilevered from the closest robotic support point. This partial assembly is therefore relying on the tensile capacity of a single brick-brick connection at the fixed support. This is not a reliable strategy since it is highly sensitive to the connection and brick material used. For the prototype, this was solved by using fast-setting epoxy putty \citep{oatey_oatey_2020} that could support a cantilevered load equivalent to five bricks \citep{parascho_robotic_2020}. But in general, this direct reliance on the mortar strength raises questions about scalability and efficiency of the \textit{cantilever method}. The connection material needs to be fully cured before proceeding to the next step, which slows down construction significantly. In our case, this was not a significant factor as the epoxy only required $\approx$ 15 mins to cure. But if traditional mortar is used, coupled with heavier construction loads (i.e., heavier or larger bricks), resisting the tension caused by the cantilevered support condition within a reasonable time-frame would not be feasible.
        
        In general, the \textit{cantilever method} performs worse with respect to tensile forces in the members than the \textit{sequential method}, which is typically undesirable in masonry construction. This comparison in made in the $T_{max}$ column in \Cref{table:cantilever_compare}, where the maximum tensile force in an element at a step is shown for both methods. \Cref{table:2robot_tension} shows the average maximum tension forces in the elements across all steps in the full construction sequence, steps before the crown, and steps after the crown. These values are consistently 40-80\% higher for the \textit{cantilever method}, although the absolute maxima are the same for both.
        
        \begin{table}[ht]
        	\renewcommand{\arraystretch}{1.0}
        	\small
        	\centering
        	\caption{Tensile forces in both 2-robot sequences (in N)}
        	\vspace{-2.5mm}
        	
    		\begin{tabular}{lSS}
    			\specialrule{.10em}{0.2em}{.2em}
    			\multicolumn{1}{c}{}
    			&\multicolumn{1}{c}{\normalsize Sequential}
    			&\multicolumn{1}{c}{\normalsize Cantilever}
    			\\
    			\specialrule{0.06em}{0.2em}{.2em}
                Average $T_{max}$ in full sequence &38.0 &53.7\\
                Average $T_{max}$ before crown &55.2 &80.0\\
                Average $T_{max}$ after crown &17.4 &30.6\\
                Maximum $T_{max}$ &114.4 &111.5\\
    			\specialrule{0.10em}{0.2em}{.2em}
    		\end{tabular}
    		
    	\label{table:2robot_tension}
    	\end{table}

    \subsection{General Challenges with 2-Robot Fabrication Methods}    
        Neither of the 2-robot construction methods discussed in \Cref{sec:sequential,,sec:cantilever} are a perfect solution to the scaffold-free construction goal. The \textit{sequential method} suffers from twisting displacements at certain steps, but minimizes overall tensile forces since the arch is always supported on the last brick placed. On the other hand, the \textit{cantilever method} minimizes the out-of-plane deformations during construction but suffers from a reliance on the tensile capacity of the mortar. These two methods do not constitute all the possible sequences, but all 2-robot fabrication sequences face issues related to scaling the size of the structure and reaching the maximum load capacity of the robots themselves. Regardless of the specific fabrication sequence, when using two robots there is always a situation where only one robot is gripping the structure. The load path is determinate, so the support force grows proportional to the size and weight of the arch. The maximum forces in the robots for the full construction sequences are shown in \Cref{table:2robot_forces}. Although the two methods have different overall behaviors, the maximum force each robot is required to support is similar. 
        
        \begin{table}[ht]
        	\renewcommand{\arraystretch}{1.0}
        	\small
        	\centering
        	\caption{Maximum forces supported by the robots}
        	\vspace{-2.5mm}
        	
    		\begin{tabular}{lc rr}
    			\specialrule{.10em}{0.2em}{.2em}
    			\centering
    			\phantom{a}
    			&\phantom{\makecell{\vspace{0.5em}}}
    			&\multicolumn{1}{c}{\normalsize{rob1}}
    			&\multicolumn{1}{c}{\normalsize{rob2}}
    			\\	
    			[0.8ex]
    			\cmidrule{3-4}
    			\phantom{a}
    			&\phantom{a}
    			&\multicolumn{1}{c}{$N$}
    			&\multicolumn{1}{c}{$N$}
    			\\
    			\specialrule{0.06em}{0.2em}{.2em}
                Sequential Method &&369 &406\\
                Cantilever Method &&362 &409\\
    			\specialrule{0.10em}{0.2em}{.2em}
    		\end{tabular}
    		
    	\label{table:2robot_forces}
    	\end{table}

%% file: sections/5threeRob.tex

\section{Central Arch Fabrication with Three Robots}\label{sec: fab_strategies_3rob}
    Adding a third robot to the construction sequence can mitigate many of the issues that arise from using either of the 2-robot methods discussed in \Cref{sec:sequential,,sec:cantilever}, and thereby improving the overall scalability of a scaffold-free fabrication method by reducing the maximum support force in each robot, and minimizing moments and displacements in the structure. A third robot (\textit{rob3}) is envisioned as a mobile agent, that has additional flexibility to place and support bricks from either side of the arch, unlike \textit{rob1} and \textit{rob2} which are fixed on either side of the structure. In this paper we will only investigate the structural influence of this third robot, rather than challenges associated with the developing the fabrication setup around reachability and collision concerns.
    
    Based on the conclusions drawn from developing the 2-robot methods come the following design criteria for a 3-robot sequence as desired:
    
    \begin{enumerate}
        \item no twisting: have at least two robots gripping the structure at all times
        \item no cantilevering: a robot must always be supporting the very last brick in the partial arch
    \end{enumerate}
    
    The inclusion of a third agent capable of placing and supporting means that the fabrication sequence is no longer prescriptive -- there is an additional robotic support available at each step beyond what is required for stability, which can be moved to support any free point on the structure. The selection of the location of the additional robotic support can be formulated as an optimization problem, where the new support position is calculated to maximize a user-specified set of structural criteria. In \Cref{sec:opt_algo,,sec:optimization,,sec:optresults} we discuss the following: the optimization approach that is used to select support points, how a 3-robot optimization-based fabrication could be implemented, and the structural results of using an optimized fabrication sequence.
    
    \subsection{Optimization Algorithm}\label{sec:opt_algo}
       The goal of the optimization is to minimize the displacement and forces experienced by the partial arch during each construction step, and thus determine the point on the structure that a robotic support should move to at the start of each optimization loop (outlined in \Cref{sec:5_3rob,,sec:5_3rob_2}). The optimization process is formulated in \Cref{eq:opt}, where an optimal node ($n^{\star}$) is defined as the support node on the structure that minimizes the sum of the average R-terms (i.e., performance ranks) at each of the fabrication steps for which the support will be present. These rank terms represent the relative performance in a particular structural criterion when placing a support at the node being examined. First, an exhaustive set of FE structural analyses based on the double-cross representation are performed at each fabrication step, where the support point is moved to every possible support location on the structure. The number of possible support locations increases with the size of the partial arch. The structural behavior from each of these models is recorded (i.e., maximum tension in the element interfaces, moment at the support, forces in the robotic supports, and maximum displacement), and once all locations have been evaluated they are ranked from best to worst.
       
       For example, if there are a total of 17 bricks already placed in the partial arch structure, there are a potential $17 \times 2 = 34 $ possible support nodes in the structure for a robot to move to in the following step (less any that are not reachable based on kinematics and if a robot is already supporting that location). Thus, 34 models would be run for this step, varying the support location in each, and then ranking these models based on how they performed in each of the structural categories of interest. The sets of criteria can be chosen by the user, in the case of the arch, the two criteria to improve are shown in \Cref{eq:opt2,eq:opt3}. \Cref{eq:opt2} minimizes the following: maximum tension in the bricks ($T_{max}$), support moment at the base ($M_{sup}$), forces in all supporting robots ($R_{rob1}, R_{rob2}, R_{rob3}$). \Cref{eq:opt3} minimizes the following: maximum tension in the bricks ($T_{max}$), support moment at the base ($M_{sup}$), forces in the active robot moving to an optimized location ($R_{rob}$), and maximum displacement in the arch ($\Delta_{max}$). 
        
        \begin{align} \label{eq:opt}
         n^{\star} &= \operatorname*{argmin}_{(support \: node)} \sum_{i=1}^{4} \overline{R_i} \\
         \nonumber\text{where}: \\
         \overline{R} &= {\rm avg}\left( R^{T_{max}} + R^{M_{sup}} + R^{F_{rob1}} + R^{F_{rob2}} + R^{F_{rob3}}  \right)  \label{eq:opt2}\\
         \overline{R} &= {\rm avg}\left( R^{T_{max}} + R^{M_{sup}} + R^{F_{rob}} + R^{{\Delta_{max}}}  \right)  \label{eq:opt3}
        \end{align}
        
        This optimization, based on multiple structural criteria, ensures that the selected support location leads to general improvements to all the important structural behavior criteria while minimizing negative side-effects. For example, we want to avoid reducing the deflection at the cost of higher forces in the supporting robots. \Cref{eq:opt2} is used for optimizations that occur before the crown of the arch has been reached, since deflections are not significant up to this point. \Cref{eq:opt3} is used for the optimizations after the crown, where the deflection and support force criteria are critical. Only the active robot -- the robot that is being moved in the optimization loop -- has its support force optimized.  
        
    \subsection{Optimization-Based Fabrication Sequences}\label{sec:optimization}
       \subsubsection{Modified Sequential Method}\label{sec:5_3rob_2}
           Given an existing 2-robot sequence, one can determine where the third support should be placed to mitigate any structural issues. For example, starting with the \textit{sequential method} and the challenges with twisting, the 3rd support could be placed on the structure to counteract this action. \Cref{table:optimization1} illustrates such a hypothetical scenario, starting with the partial arch during the \textit{sequential method} at step 8. Normally the following step would cause off-center twisting to occur as \textit{rob2} releases its grip (see Appendix A, \Cref{table:experimental_data1} for the full sequential sequence). But the third support is free to be placed at an optimal location on the structure before this twisting occurs, which is highlighted by the cells labelled ``O1'' in \Cref{table:optimization1}. This location is identified on the basis of an optimization; the selected brick and node to grip must lead to improved structural behavior in all three of the steps it will be present in. The objective function is formulated as choosing a support point that leads to the greatest reduction in displacements, forces, and moments experienced by the structure. The issue with twisting on step 9 has now been resolved through the inclusion of the \textit{rob3} support, and this process is continued for the full fabrication sequence as the third support is continuously moved to optimize the behavior over the following range of three steps: $O2, O3, O4,...,On$. But not all fabrication steps are optimized when adding a third support to an existing 2-robot method in this way. For example, the steps when \textit{rob3} releases the structure and moves (10 and 12 in \Cref{table:optimization1}) are the same as in the standard 2-robot \textit{sequential method}.
           
            \definecolor{babyblue}{rgb}{0.54, 0.81, 0.94}
        
        	\begin{table}[ht]
        	\scriptsize
        	\renewcommand{\arraystretch}{1.0}
        	\centering
        	\caption{Modifying the sequential method with an optimized third support}
        	\vspace{-2.5mm}
                
                \begin{threeparttable}
        		\begin{tabular}{cc r cccccc}
        			\specialrule{.10em}{0.2em}{.2em}
        			\multirow{2}{*}{\normalsize  Step\tnote{1}}
        			&\multirow{2}{*}{\normalsize \makecell{Total \\ Bricks}}
        			&\phantom{\makecell{\vspace{0.5em}}}
        			&\multicolumn{2}{c}{\normalsize $rob1$}
        			&\multicolumn{2}{c}{\normalsize $rob2$}
        			&\multicolumn{2}{c}{\normalsize $rob3$}
        			\\	
        			[0.8ex]
        			\cmidrule{4-9}
        			\multicolumn{1}{c}{}
        			&\phantom{a}
        			&\phantom{a}
        			&\multicolumn{1}{c}{\small b}
        			&\multicolumn{1}{c}{\small n}
        			&\multicolumn{1}{c}{\small b}
        			&\multicolumn{1}{c}{\small n}
        			&\multicolumn{1}{c}{\small b}
        			&\multicolumn{1}{c}{\small n}
        			\\
        			\specialrule{0.06em}{0.2em}{.2em}
        			\vdots &\vdots &&\vdots &\vdots &\vdots  &\vdots &\vdots &\vdots \\
                    8 &4 &&5 &8 &4 &7 &-- &-- \\
                     &4 &&5 &8 &4 &7 &\cellcolor{babyblue}O1 &\cellcolor{babyblue}O1 \\
                    9\tnote{2} &4 &&5 &8 &-- &-- &\cellcolor{babyblue}O1 &\cellcolor{babyblue}O1 \\
                     &5 &&5 &8 &6 &11 &\cellcolor{babyblue}O1 &\cellcolor{babyblue}O1 \\
                    10 &5 &&5 &8 &6 &11 &-- &-- \\
                    &5 &&5 &8 &6 &11 &\cellcolor{babyblue}O2 &\cellcolor{babyblue}O2 \\
                    11\tnote{2} &5 &&-- &-- &6 &11 &\cellcolor{babyblue}O2 &\cellcolor{babyblue}O2 \\
                    &6 &&7 &13 &6 &11 &\cellcolor{babyblue}O2 &\cellcolor{babyblue}O2 \\
                    12 &6 &&7 &13 &6 &11 &-- &-- \\
                    \vdots &\vdots &&\vdots &\vdots &\vdots  &\vdots &\vdots &\vdots \\
        			\specialrule{0.10em}{0.2em}{.2em}
        		\end{tabular}
        		
                \begin{tablenotes}
                    \scriptsize
                    \item[] \sethlcolor{babyblue} \hl{------} = optimized support location
                    \item[1] step numbering corresponds to the sequential method sequence
                    \item[2] steps experiencing off-center twisting in sequential method
                \end{tablenotes}
        		\end{threeparttable}	
        			
        	\label{table:optimization1}
        	\end{table}

        \subsubsection{Optimized Three Robot Method}\label{sec:5_3rob}
            In contrast to a partially optimized sequence, as described in \Cref{sec:5_3rob_2}, a fully optimized fabrication sequence is one where each step is covered by an optimization calculation for any one of the robots. Such a fabrication is superior to a modified 2-robot sequence, as the position of all three robots is actively included in the optimization process. Therefore, the whole fabrication sequence can be redefined with respect to a set of user-specified structural optimization criteria in addition to the no twisting and cantilevering design criteria, rather than just building on an existing and potentially limited 2-robot sequence. \Cref{table:optimization2} describes an \textit{optimized three robot method}, where it can be seen that no step is without an optimized point, which is highlighted in blue.
            
        	\begin{table}[ht]
        	\scriptsize
        	\renewcommand{\arraystretch}{1.0}
        	\centering
        	\caption{3-robot cascading optimization}
        	\vspace{-2.5mm}
                
                \begin{threeparttable}
        		\begin{tabular}{cc r cccccc}
        			\specialrule{.10em}{0.2em}{.2em}
        			\multirow{2}{*}{\normalsize  Step}
        			&\multirow{2}{*}{\normalsize \makecell{Total \\ Bricks}}
        			&\phantom{\makecell{\vspace{0.5em}}}
        			&\multicolumn{2}{c}{\normalsize $rob1$}
        			&\multicolumn{2}{c}{\normalsize $rob2$}
        			&\multicolumn{2}{c}{\normalsize $rob3$}
        			\\	
        			[0.8ex]
        			\cmidrule{4-9}
        			\multicolumn{1}{c}{}
        			&\phantom{a}
        			&\phantom{a}
        			&\multicolumn{1}{c}{\small b}
        			&\multicolumn{1}{c}{\small n}
        			&\multicolumn{1}{c}{\small b}
        			&\multicolumn{1}{c}{\small n}
        			&\multicolumn{1}{c}{\small b}
        			&\multicolumn{1}{c}{\small n}
        			\\
        			\specialrule{0.06em}{0.2em}{.2em}
1	&	1	&&	1	&	1	&	--	&	--	&	--	&	--	\\
2	&	2	&&	1	&	1	&	2	&	2	&	--	&	--	\\
3	&	3	&&	1	&	1	&	2	&	2	&	\cellcolor{babyblue}O1	&	\cellcolor{babyblue}O1	\\
4	&	3	&&	--	&	--	&	2	&	2	&	\cellcolor{babyblue}O1	&	\cellcolor{babyblue}O1	\\
5	&	4	&&	4	&	7	&	2	&	2	&	\cellcolor{babyblue}O1	&	\cellcolor{babyblue}O1	\\
6	&	4	&&	4	&	7	&	--	&	--	&	\cellcolor{babyblue}O1	&	\cellcolor{babyblue}O1	\\
7	&	4	&&	4	&	7	&	\cellcolor{babyblue}O2	&	\cellcolor{babyblue}O2	&	\cellcolor{babyblue}O1	&	\cellcolor{babyblue}O1	\\
8	&	4	&&	4	&	7	&	\cellcolor{babyblue}O2	&	\cellcolor{babyblue}O2	&	--	&	--	\\
9	&	5	&&	4	&	7	&	\cellcolor{babyblue}O2	&	\cellcolor{babyblue}O2	&	5	&	8	\\
10	&	5	&&	--	&	--	&	\cellcolor{babyblue}O2	&	\cellcolor{babyblue}O2	&	5	&	8	\\
11	&	5	&&	\cellcolor{babyblue}O3	&	\cellcolor{babyblue}O3	&	\cellcolor{babyblue}O2	&	\cellcolor{babyblue}O2	&	5	&	8	\\
12	&	5	&&	\cellcolor{babyblue}O3	&	\cellcolor{babyblue}O3	&	--	&	--	&	5	&	8	\\
13	&	6	&&	\cellcolor{babyblue}O3	&	\cellcolor{babyblue}O3	&	6	&	11	&	5	&	8	\\
14	&	6	&&	\cellcolor{babyblue}O3	&	\cellcolor{babyblue}O3	&	6	&	11	&	--	&	--	\\
15	&	6	&&	\cellcolor{babyblue}O3	&	\cellcolor{babyblue}O3	&	6	&	11	&	\cellcolor{babyblue}O4	&	\cellcolor{babyblue}O4	\\
16	&	6	&&	--	&	--	&	6	&	11	&	\cellcolor{babyblue}O4	&	\cellcolor{babyblue}O4	\\
17	&	7	&&	7	&	13	&	6	&	11	&	\cellcolor{babyblue}O4	&	\cellcolor{babyblue}O4	\\
18	&	7	&&	7	&	13	&	--	&	--	&	\cellcolor{babyblue}O4	&	\cellcolor{babyblue}O4	\\
19	&	7	&&	7	&	13	&	\cellcolor{babyblue}O5	&	\cellcolor{babyblue}O5	&	\cellcolor{babyblue}O4	&	\cellcolor{babyblue}O4	\\
20	&	7	&&	7	&	13	&	\cellcolor{babyblue}O5	&	\cellcolor{babyblue}O5	&	--	&	--	\\
21	&	8	&&	7	&	13	&	\cellcolor{babyblue}O5	&	\cellcolor{babyblue}O5	&	8	&	15	\\
	&	&&	&	&	&	&	\phantom{O12}	&	\phantom{O12}	\\
                    \vdots &\vdots &&\vdots &\vdots &\vdots  &\vdots &\vdots &\vdots \\
	&	&&	&	&	&	&	\phantom{O12}	&	\phantom{O12}	\\
                    
58	&	17	&&	--	&	--	&	\cellcolor{babyblue}O14	&	\cellcolor{babyblue}O14	&	17	&	33	\\
59	&	17	&&	\cellcolor{babyblue}O15	&	\cellcolor{babyblue}O15	&	\cellcolor{babyblue}O14	&	\cellcolor{babyblue}O14	&	17	&	33	\\
60	&	17	&&	\cellcolor{babyblue}O15	&	\cellcolor{babyblue}O15	&	--	&	--	&	17	&	33	\\
61	&	18	&&	\cellcolor{babyblue}O15	&	\cellcolor{babyblue}O15	&	18	&	35	&	17	&	33	\\
62	&	18	&&	\cellcolor{babyblue}O15	&	\cellcolor{babyblue}O15	&	18	&	35	&	--	&	--	\\
        			\vdots &\vdots &&\vdots &\vdots &\vdots  &\vdots &\vdots &\vdots \\
        			\specialrule{0.10em}{0.2em}{.2em}
        		\end{tabular}
        		
                \begin{tablenotes}
                    \scriptsize
                    \item[] \sethlcolor{babyblue} \hl{------} = optimized support location
                \end{tablenotes}
        		\end{threeparttable}	
        			
        	\label{table:optimization2}
        	\end{table}

            An optimization is performed for each brick placed into the arch (excluding the first two), where $\sum_{n=3}^{n_{max}} 4 \cdot 2n$, is the number of individual analysis models that are needed for the full fabrication (i.e., 2,576 models with $n_{max} = 25$ bricks). It is important to point out that this constitutes a simplification to a cascading optimization, where each cycle is linked to the next as their first and last steps overlap. For example, looking at \Cref{table:optimization2}, Step 11 falls on the last step in O2 and the first step in O3. This means that an exhaustive evaluation would require $(2n)(2(n-1))$ models to run for each of these overlapping steps, resulting in a total $\sum_{n=3}^{n_{max}} 3 \cdot 2n  + (2n)(2(n-1)$  individual analysis models that would need to be run for the full fabrication (i.e., 22,724 models with $n_{max} = 25$ bricks). Even with the simplified modeling approach this is considered an excessive number of analyses to perform. Therefore we treat this overlapping step as pertaining only to one of the optimizations, thereby reducing the order of the problem from $O(n^2)$ to $O(n)$.
            
\begin{figure}[H]
                \centering
        	    \includegraphics [angle=0,trim={0.5cm 0.5cm 0.5cm 0.5cm},clip,width=0.95\linewidth]{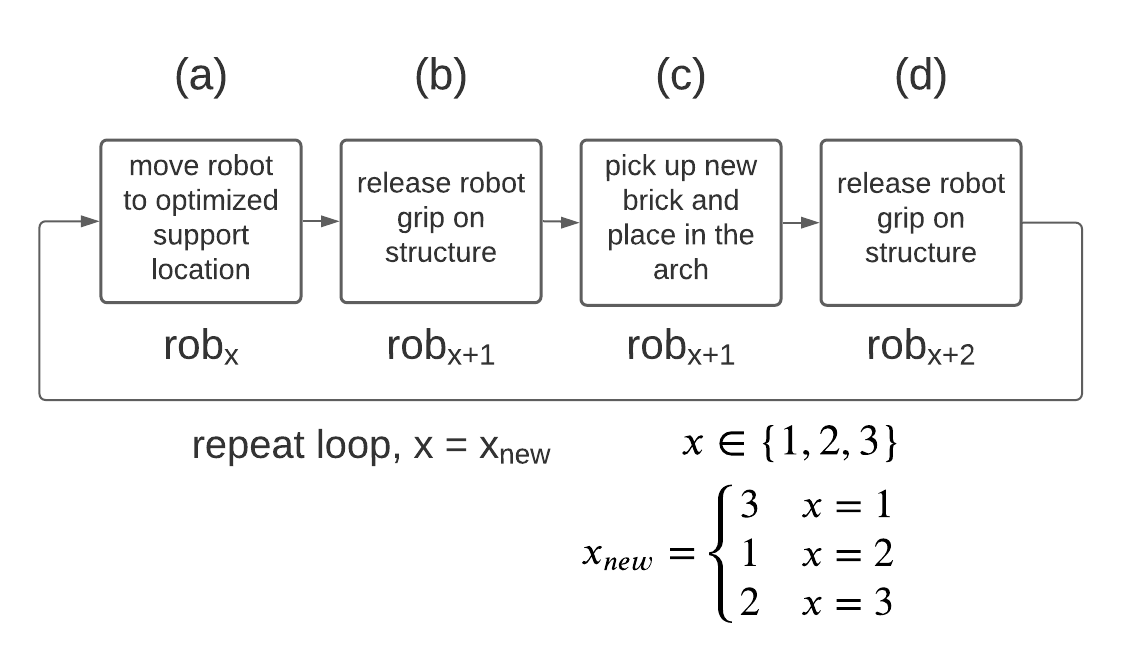}
                \caption{Four-step fabrication loop.}
                \label{fig:fab_loop}
            \end{figure}

            The general fabrication process can be described by the four-step loop (a to d) in \Cref{fig:fab_loop}. This loop is repeated over the construction of the whole arch, with the roles of the three robots varying depending on what stage of the construction is being completed. The placement of bricks is sequential, but each robot takes a turn being the optimized support every three bricks, creating the cascading pattern of optimizations as seen in \Cref{table:optimization2}.
            
\begin{figure}[H]
            \centering
            
            \begin{subfigure}[b]{0.49\linewidth}
        	    \centering
        		\includegraphics [angle=0,trim={5cm 165cm 100cm 2cm},clip,width=0.99\linewidth]{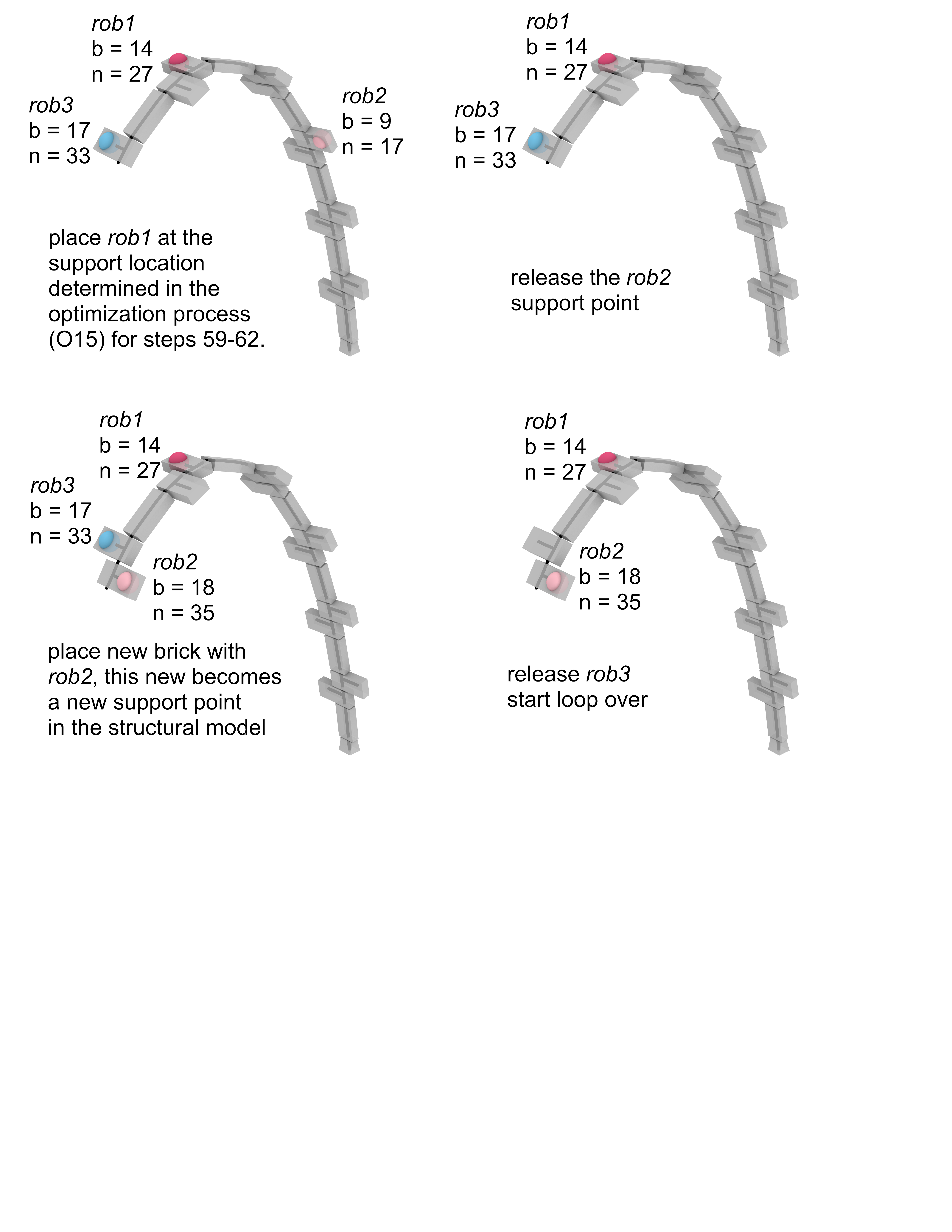}
        		\caption{Step 59}
        		\label{fig:3rob_58}
        	\end{subfigure}
        	\begin{subfigure}[b]{0.49\linewidth}
        	    \centering
        		\includegraphics [angle=0,trim={85cm 165cm 20cm 2cm},clip,width=0.99\linewidth]{3rob}
        		\caption{Step 60}
        		\label{fig:3rob_59}
        	\end{subfigure}
        	
        	\vspace{1em}
        	
            \begin{subfigure}[b]{0.49\linewidth}
        	    \centering
        		\includegraphics [angle=0,trim={5cm 89cm 100cm 78cm},clip,width=0.99\linewidth]{3rob}
        		\caption{Step 61}
        		\label{fig:3rob_60}
        	\end{subfigure}
        	\begin{subfigure}[b]{0.49\linewidth}
        	    \centering
        		\includegraphics [angle=0,trim={85cm 89cm 20cm 78cm},clip,width=0.99\linewidth]{3rob}
        		\caption{Step 62}
        		\label{fig:3rob_61}
        	\end{subfigure}

            \caption{An example four step loop in the 3-robot construction sequence.}
            \label{fig:3rob_sequence}
        \end{figure}

            \Cref{fig:3rob_sequence} shows a concrete implementation of this for just one loop starting at step \#59 (the full arch fabrication sequence can be found in Appendix B, \Cref{table:experimental_data3}), where the four-step loop (a to d) is described in detail at each step. In \Cref{fig:3rob_sequence} the optimized location (O15) in step \#59 is calculated for the structure based on the algorithm that is explained in \Cref{sec:opt_algo}.

    \subsection{Optimization Results}   \label{sec:optresults}  
        With the optimization criteria in place, a total of 23 independent optimizations were required in the fabrication simulation of the full arch (25 total bricks) using the fully \textit{optimized three robot method}. The full set of structural analysis results using these optimized locations are shown in Appendix B, \Cref{table:experimental_data3}. With this method the construction takes a total of 90 steps, which is almost double what is required in a 2-robot method. However, this is due to the re-positioning of the optimized support every four steps. This longer fabrication sequence is partially offset by the fact that waiting for mortar curing is not required as there is always a robot supporting the latest brick in the sequence and there is no cantilevering action.
        
        The \textit{optimized three robot method} is able to significantly improve the structural behavior over the full construction sequence when compared to both 2-robot methods discussed in \Cref{sec:sequential,,sec:cantilever}. \Cref{table:optimization_results} compares the maximum structural forces, moments, and displacements measured during each of the three fabrication sequences. The 3-robot sequence is able to re-distribute the forces in the robotic supports reducing the maximum force in \textit{rob1} and \textit{rob2} by 27\% and 37\% respectively. Twisting behavior is also mitigated, thereby reducing the maximum moment and displacement by 67\% and 93\% respectively when compared to the \textit{sequential method}. The maximum tension in an element is reduced by 23\% when compared to the 2-robot sequences. 
        
           	\begin{table}[ht]
        	\normalsize
        	\renewcommand{\arraystretch}{1.2}
        	\centering
        	\caption{Improvement to structural behavior with 3-robot optimized sequence}
        	\vspace{-2.5mm}
                
                \begin{threeparttable}
        		\begin{tabular}{c l ccc}
        			\specialrule{.10em}{0.2em}{.2em}
        			\phantom{\makecell{\vspace{0.5em}}}
        			&\phantom{a}
        			&\multicolumn{1}{c}{\normalsize  Seq.}
        			&\multicolumn{1}{c}{\normalsize Cant.}
        			&\multicolumn{1}{c}{\normalsize 3-rob}
        			\\	
        			\specialrule{0.06em}{0.2em}{.2em}
                    \normalsize $F_{rob1}$ &\normalsize $[N]$ &369 &362 &270	\\
                    \normalsize $F_{rob2}$ &\normalsize $[N]$ &406 &409 &257	\\
                    \normalsize $F_{rob3}$ &\normalsize $[N]$ &-- &-- &254	\\
                    \normalsize $M_{sup}$ &\normalsize $[N \cdot m]$ &66.0 &21.7 &21.5	\\
                    \normalsize $T_{max}$ &\normalsize $[N]$ &114.4 &111.5 &87.7	\\
                    \normalsize $\Delta_{max}$ &\normalsize $[10^{-3} \: mm]$ &306.5 &15.9 &21.9	\\
        			\specialrule{0.10em}{0.2em}{.2em}
        		\end{tabular}
        		\end{threeparttable}	
        			
        	\label{table:optimization_results}
        	\end{table}       

%% file: sections/6conclusions.tex

\section{Conclusion} \label{conclusion}
    In this paper we build on previous research \citep{han_concept_2020,parascho_robotic_2020, parascho_lightvault_2021} that has demonstrated how industrial robots sequenced in a cooperative manner are a viable method for building a discrete brick arch without temporary scaffolding. In the context of fabricating a masonry arch, a minimum of two robots are needed to successfully execute a scaffold-free cooperative placement and support sequence. A simplified geometric structural analysis framework, based on rigid elements and flexible joints, is presented, validated and used to evaluate three fabrication methods: sequential, cantilever, and optimized. While the \textit{sequential method}, where two robots take turns placing bricks, is simple to implement, the out-of-plane twisting observed is detrimental in larger-scale structures. The \textit{cantilever method}, where the robots take turns placing three bricks in a row, solves the issue of twisting. But this method raises questions about applicability to larger masonry structures due to a reliance on the strength of the mortar to support bricks throughout the fabrication.
    
    The addition of a third robot was analyzed to further improve upon the 2-robot scaffold-free fabrication methods. We show that including a third robot improves the structural behavior during fabrication when formulating the fabrication sequence as the result of an optimization problem. This approach creates an opportunity to design a fabrication sequence on the basis of a set of user-specified structural optimization targets (e.g., displacements, forces, moments) in addition to the goal of removing the need for temporary scaffolding. Generally the \textit{optimized three robot method} is an improvement as it combines many positive aspects of the 2-robot methods, while simultaneously mitigating the negative: twisting deformations, tensile forces, and large support moments.
    
    Geometrically complex discrete element systems have significant economic and environmental consequences due to their challenging construction process, which is still deeply rooted in a pre-robotic construction rationale. This work re-imagines the construction of discrete element structures in response to emerging innovations in robotic fabrication in the building industry and thus hopes to bring about transformative sustainable changes in construction practices while increasing onsite productivity. This study has generated new knowledge on how to design and analyze robotic sequences for spatial structural systems, and illustrates the potential of utilizing cooperative robotic fabrication for scaffold-free construction applications.

%% file: sections/7acknowledgements.tex

\section{Acknowledgements} \label{acknowledge}

The authors would like to acknowledge the Metropolis Project of Princeton University, and the Princeton Catalysis Initiative for financially supporting this project. We would also like to thank Itasca C.G.A for providing a 3DEC software license under the Education Partnership Program.

%% file: sections/99appendix.tex
\appendix
\clearpage

\setcounter{table}{0}
\renewcommand{\thetable}{A.\arabic{table}}


\begin{landscape}

\section{2-Robot Fabrication Sequence Analysis Results} \label{sec:appendixA}
    
	\begin{table}[h!]
	\scriptsize
	\renewcommand{\arraystretch}{1.0}
	\centering
	\caption{Analysis results for the sequential method}
        
        \begin{threeparttable}
		\begin{tabular}{c cccr ccccccr cccccr ccccr ccc}
			\specialrule{.10em}{0.2em}{.2em}
			\centering
			&\multirow{2}{*}{\normalsize  Step}
			&\multirow{2}{*}{\normalsize  Type}
			&\multirow{2}{*}{\normalsize \makecell{Total \\ Bricks}}
			&\phantom{\makecell{\vspace{0.5em}}}
			&\multicolumn{2}{c}{\normalsize $rob1$}
			&\multicolumn{2}{c}{\normalsize $rob2$}
			&\multicolumn{2}{c}{\normalsize $rob3$}
			&\phantom{\tiny a}
			&\multicolumn{1}{c}{\normalsize  $F_{rob1}$}
			&\multicolumn{1}{c}{\normalsize  $F_{rob2}$}
			&\multicolumn{1}{c}{\normalsize  $F_{rob3}$}
			&\multicolumn{1}{c}{\normalsize  $F_{sup}$}
			&\multicolumn{1}{c}{\normalsize  $M_{sup}$}
			&\phantom{\tiny a}
	        &\multicolumn{1}{c}{\normalsize  $F_{min}$}
			&\multicolumn{1}{c}{\normalsize  $F_{max}$}
			&\multicolumn{1}{c}{\normalsize  $F_{avg}$}
			&\multicolumn{1}{c}{\normalsize  $T_{\%}\tnote{1}$}
			&\phantom{\tiny a}
	        &\multicolumn{1}{c}{\normalsize  $\Delta_{max}$}
			&\multicolumn{1}{c}{\normalsize  $\Delta_{avg}$}
			&\multicolumn{1}{c}{\normalsize  $\Delta_{\sigma}\tnote{2}$}
			\\	
			[0.8ex]
			\cmidrule{6-11}
			\cmidrule{13-17}
			\cmidrule{19-22}
			\cmidrule{24-26}
			&\multicolumn{1}{c}{}
			&\phantom{a}
			&\multicolumn{1}{c}{}
			&\phantom{a}
			&\multicolumn{1}{c}{\small b}
			&\multicolumn{1}{c}{\small n}
			&\multicolumn{1}{c}{\small b}
			&\multicolumn{1}{c}{\small n}
			&\multicolumn{1}{c}{\small b}
			&\multicolumn{1}{c}{\small n}
			&\phantom{\tiny a}
			&\multicolumn{1}{c}{\small $N$}
			&\multicolumn{1}{c}{\small $N$}
			&\multicolumn{1}{c}{\small $N$}
			&\multicolumn{1}{c}{\small $N$}
			&\multicolumn{1}{c}{\small $N \cdot m$}
			&\phantom{\tiny a}
			&\multicolumn{1}{c}{\small $N$}
			&\multicolumn{1}{c}{\small $N$}
			&\multicolumn{1}{c}{\small $N$}
			&\phantom{a}
			&\phantom{\tiny a}
			&\multicolumn{1}{c}{\scriptsize  $[10^{-3}]$ mm}
			&\multicolumn{1}{c}{\scriptsize  $[10^{-3}]$ mm}
			&\multicolumn{1}{c}{\scriptsize  $[10^{-3}]$ mm}
			\\
			\specialrule{0.06em}{0.2em}{.2em}
\rowcolor{lightgray} &1 &place &1 &&1 &1 &-- &-- &-- &-- &&35.6 &-- &-- &1.8 &0.1 &&0.0 &17.6 &4.4 &50.0 &&0.0 &0.0 &0.0 \\
\rowcolor{lightgray} &2 &place &2 &&1 &1 &2 &2 &-- &-- &&36.6 &38.3 &-- &2.9 &0.5 &&0.0 &17.6 &2.5 &71.4 &&0.1 &0.0 &0.0 \\
&3 &release &2 &&-- &-- &2 &2 &-- &-- &&-- &54.3 &-- &19.0 &0.8 &&-17.5 &17.6 &0.1 &71.4 &&0.1 &0.0 &0.0 \\
&4 &place &3 &&3 &4 &2 &2 &-- &-- &&36.1 &54.6 &-- &18.1 &0.4 &&-17.6 &17.6 &0.0 &50.0 &&0.1 &0.0 &0.0 \\
&5 &release &3 &&3 &4 &-- &-- &-- &-- &&71.5 &-- &-- &36.2 &1.1 &&-35.4 &35.1 &-1.9 &30.0 &&0.1 &0.1 &0.1 \\
&6 &place &4 &&3 &4 &4 &7 &-- &-- &&73.4 &35.2 &-- &35.9 &0.8 &&-35.4 &35.1 &-0.1 &46.2 &&0.2 &0.1 &0.1 \\
\rowcolor{lightgray} &7 &release &4 &&-- &-- &4 &7 &-- &-- &&-- &87.3 &-- &56.9 &3.6 &&-54.6 &68.3 &1.5 &53.8 &&0.4 &0.3 &0.2 \\
\rowcolor{lightgray} &8 &place &5 &&5 &8 &4 &7 &-- &-- &&25.7 &101.4 &-- &55.9 &3.0 &&-54.2 &68.6 &0.7 &43.8 &&0.4 &0.2 &0.1 \\
\rowcolor{yellow} &9 &release &5 &&5 &8 &-- &-- &-- &-- &&75.0 &-- &-- &106.1 &6.8 &&-103.9 &36.9 &-24.2 &43.8 &&1.3 &0.7 &0.4 \\
&10 &place &6 &&5 &8 &6 &11 &-- &-- &&76.9 &54.5 &-- &85.2 &5.4 &&-82.9 &57.7 &-6.6 &47.4 &&0.9 &0.4 &0.3 \\
\rowcolor{yellow} &11 &release &6 &&-- &-- &6 &11 &-- &-- &&-- &85.2 &-- &132.2 &8.3 &&-129.9 &45.5 &-31.8 &36.8 &&2.2 &1.1 &0.6 \\
&12 &place &7 &&7 &13 &6 &11 &-- &-- &&78.0 &68.5 &-- &107.9 &8.8 &&-104.8 &70.9 &-6.1 &59.1 &&2.0 &0.9 &0.6 \\
\rowcolor{lightgray} &13 &release &7 &&7 &13 &-- &-- &-- &-- &&127.7 &-- &-- &124.4 &11.5 &&-120.3 &105.4 &-8.7 &50.0 &&2.4 &1.5 &0.8 \\
\rowcolor{lightgray} &14 &place &8 &&7 &13 &8 &15 &-- &-- &&168.8 &56.5 &-- &114.7 &7.5 &&-112.3 &111.5 &-2.6 &48.0 &&1.4 &0.7 &0.5 \\
&15 &release &8 &&-- &-- &8 &15 &-- &-- &&-- &138.4 &-- &149.5 &13.9 &&-145.3 &97.7 &-16.6 &44.0 &&3.6 &2.0 &1.2 \\
&16 &place &9 &&9 &16 &8 &15 &-- &-- &&58.6 &186.2 &-- &138.1 &10.0 &&-135.4 &105.6 &-8.6 &50.0 &&2.3 &1.2 &0.9 \\
&17 &release &9 &&9 &16 &-- &-- &-- &-- &&144.5 &-- &-- &178.9 &16.7 &&-174.8 &101.1 &-27.4 &46.4 &&5.1 &2.8 &1.8 \\
&18 &place &10 &&9 &16 &10 &19 &-- &-- &&187.7 &44.5 &-- &161.7 &12.1 &&-158.8 &114.4 &-13.7 &48.4 &&3.2 &1.6 &1.2 \\
\rowcolor{lightgray} &19 &release &10 &&-- &-- &10 &19 &-- &-- &&-- &110.8 &-- &248.2 &21.7 &&-244.8 &72.2 &-65.0 &32.3 &&9.2 &5.3 &3.0 \\
\rowcolor{lightgray} &20 &place &11 &&11 &20 &10 &19 &-- &-- &&-46.0 &180.3 &-- &232.0 &19.8 &&-228.7 &85.2 &-48.1 &35.3 &&8.1 &4.3 &2.7 \\
\rowcolor{yellow} &21 &release &11 &&11 &20 &-- &-- &-- &-- &&86.6 &-- &-- &311.7 &21.5 &&-309.8 &23.7 &-103.8 &26.5 &&12.8 &6.9 &3.9 \\
&22 &place &12 &&11 &20 &12 &23 &-- &-- &&79.0 &46.4 &-- &307.3 &21.5 &&-305.2 &28.7 &-92.1 &24.3 &&11.5 &6.1 &4.0 \\
\rowcolor{yellow} &23 &release &12 &&-- &-- &12 &23 &-- &-- &&-- &89.7 &-- &347.9 &22.0 &&-346.5 &11.4 &-120.2 &18.9 &&16.9 &8.3 &4.6 \\
&24 &place &13 &&13 &25 &12 &23 &-- &-- &&34.6 &119.1 &-- &352.6 &17.9 &&-351.5 &34.7 &-117.6 &17.5 &&12.8 &6.4 &3.9 \\
\rowcolor{lightgray} &25 &release &13 &&13 &25 &-- &-- &-- &-- &&77.2 &-- &-- &416.5 &14.4 &&-416.2 &0.8 &-163.0 &10.0 &&12.9 &7.6 &4.2 \\
\rowcolor{lightgray} &26 &place &14 &&13 &25 &14 &26 &-- &-- &&362.2 &-203.2 &-- &346.4 &16.8 &&-345.4 &79.9 &-102.3 &16.3 &&10.7 &5.6 &3.8 \\
&27 &release &14 &&-- &-- &14 &26 &-- &-- &&-- &96.7 &-- &436.5 &10.7 &&-436.3 &0.8 &-169.4 &11.6 &&12.2 &7.0 &4.0 \\
&28 &place &15 &&15 &28 &14 &26 &-- &-- &&-146.6 &280.0 &-- &393.3 &13.6 &&-392.8 &38.2 &-129.4 &15.2 &&11.2 &5.9 &4.0 \\
&29 &release &15 &&15 &28 &-- &-- &-- &-- &&119.9 &-- &-- &448.9 &7.7 &&-448.9 &0.8 &-171.1 &10.9 &&11.3 &6.5 &3.8 \\
&30 &place &16 &&15 &28 &16 &31 &-- &-- &&197.6 &-52.0 &-- &418.3 &10.9 &&-418.1 &20.2 &-140.8 &16.3 &&11.1 &5.8 &4.0 \\
\rowcolor{lightgray} &31 &release &16 &&-- &-- &16 &31 &-- &-- &&-- &143.9 &-- &463.0 &3.4 &&-463.0 &0.8 &-177.6 &12.2 &&10.3 &6.4 &3.2 \\
\rowcolor{lightgray} &32 &place &17 &&17 &33 &16 &31 &-- &-- &&-31.1 &171.4 &-- &454.7 &4.8 &&-454.7 &14.4 &-162.1 &15.4 &&10.5 &6.0 &3.5 \\
\rowcolor{yellow} &33 &release &17 &&17 &33 &-- &-- &-- &-- &&170.8 &-- &-- &470.0 &21.3 &&-470.0 &0.8 &-177.0 &9.6 &&70.8 &33.1 &23.1 \\
&34 &place &18 &&17 &33 &18 &35 &-- &-- &&123.1 &93.1 &-- &465.0 &2.8 &&-465.0 &33.6 &-165.7 &14.5 &&10.2 &6.0 &3.4 \\
\rowcolor{yellow} &35 &release &18 &&-- &-- &18 &35 &-- &-- &&-- &203.8 &-- &468.9 &29.6 &&-468.9 &0.8 &-173.2 &9.1 &&97.5 &47.3 &33.2 \\
&36 &place &19 &&19 &37 &18 &35 &-- &-- &&152.2 &109.2 &-- &463.7 &1.7 &&-463.7 &44.9 &-167.3 &12.1 &&10.2 &5.9 &3.3 \\
\rowcolor{lightgray} &37 &release &19 &&19 &37 &-- &-- &-- &-- &&236.3 &-- &-- &471.3 &1.1 &&-471.3 &0.8 &-176.8 &12.1 &&10.9 &6.6 &3.3 \\
\rowcolor{lightgray} &38 &place &20 &&19 &37 &20 &39 &-- &-- &&256.1 &68.0 &-- &462.7 &2.2 &&-462.7 &3.5 &-163.8 &11.5 &&10.1 &5.6 &3.5 \\
&39 &release &20 &&-- &-- &20 &39 &-- &-- &&-- &269.9 &-- &471.5 &2.8 &&-471.5 &0.8 &-176.0 &9.8 &&12.5 &7.4 &3.9 \\
&40 &place &21 &&21 &40 &20 &39 &-- &-- &&54.5 &281.1 &-- &465.9 &1.8 &&-465.9 &2.2 &-165.1 &12.5 &&10.2 &5.7 &3.5 \\
&41 &release &21 &&21 &40 &-- &-- &-- &-- &&303.8 &-- &-- &472.0 &1.7 &&-471.9 &0.8 &-176.5 &10.9 &&10.8 &6.5 &3.4 \\
&42 &place &22 &&21 &40 &22 &43 &-- &-- &&308.8 &36.0 &-- &467.8 &1.5 &&-467.7 &1.0 &-166.9 &13.4 &&10.4 &5.7 &3.6 \\
\rowcolor{lightgray} &43 &release &22 &&-- &-- &22 &43 &-- &-- &&-- &338.1 &-- &472.6 &2.0 &&-472.6 &0.8 &-182.8 &10.4 &&11.3 &6.8 &3.4 \\
\rowcolor{lightgray} &44 &place &23 &&23 &44 &22 &43 &-- &-- &&26.0 &349.5 &-- &471.3 &1.3 &&-471.3 &10.5 &-174.2 &12.9 &&10.6 &6.2 &3.5 \\
\rowcolor{yellow} &45 &release &23 &&23 &44 &-- &-- &-- &-- &&369.3 &-- &-- &477.1 &56.4 &&-476.6 &0.8 &-185.6 &10.0 &&264.4 &129.4 &89.9 \\
&46 &place &24 &&23 &44 &24 &47 &-- &-- &&250.8 &179.6 &-- &472.7 &1.6 &&-472.6 &68.3 &-178.7 &12.3 &&12.3 &7.1 &3.9 \\
\rowcolor{yellow} &47 &release &24 &&-- &-- &24 &47 &-- &-- &&-- &405.5 &-- &474.7 &66.0 &&-474.4 &0.8 &-186.9 &9.6 &&306.5 &149.6 &103.4 \\
&48 &place &25 &&25 &49 &24 &47 &-- &-- &&277.4 &206.6 &-- &471.3 &1.7 &&-471.3 &91.3 &-184.7 &11.7 &&13.0 &7.2 &4.2 \\
\rowcolor{lightgray} &49 &release &25 &&-- &-- &-- &-- &-- &-- &&-- &-- &-- &472.3 &0.7 &&-471.9 &0.8 &-204.7 &10.4 &&11.9 &7.4 &3.6 \\
			\specialrule{0.10em}{0.2em}{.2em}
		\end{tabular}
		
        \begin{tablenotes}
            \scriptsize
            \sethlcolor{yellow} \item[] \hl{------} = twisting step
            \sethlcolor{lightgray} \item[]  \hl{------} = same in cantilever and sequential
            \item[1] \% of total edges in the analysis model that are in tension
            \item[2] standard deviation of all nodal displacements
        \end{tablenotes}
		\end{threeparttable}	
			
	\label{table:experimental_data1}
	\end{table}
    
    \end{landscape}
    
\begin{landscape}

	\begin{table}[h!]
	\scriptsize
	\renewcommand{\arraystretch}{1.0}
	\centering
	\caption{Analysis results for the cantilever method}
        
        \begin{threeparttable}
		\begin{tabular}{c cccr ccccccr cccccr ccccr ccc}
			\specialrule{.10em}{0.2em}{.2em}
			\centering
			&\multirow{2}{*}{\normalsize  Step}
			&\multirow{2}{*}{\normalsize  Type}
			&\multirow{2}{*}{\normalsize \makecell{Total \\ Bricks}}
			&\phantom{\makecell{\vspace{0.5em}}}
			&\multicolumn{2}{c}{\normalsize $rob1$}
			&\multicolumn{2}{c}{\normalsize $rob2$}
			&\multicolumn{2}{c}{\normalsize $rob3$}
			&\phantom{\tiny a}
			&\multicolumn{1}{c}{\normalsize  $F_{rob1}$}
			&\multicolumn{1}{c}{\normalsize  $F_{rob2}$}
			&\multicolumn{1}{c}{\normalsize  $F_{rob3}$}
			&\multicolumn{1}{c}{\normalsize  $F_{sup}$}
			&\multicolumn{1}{c}{\normalsize  $M_{sup}$}
			&\phantom{\tiny a}
	        &\multicolumn{1}{c}{\normalsize  $F_{min}$}
			&\multicolumn{1}{c}{\normalsize  $F_{max}$}
			&\multicolumn{1}{c}{\normalsize  $F_{avg}$}
			&\multicolumn{1}{c}{\normalsize  $T_{\%}\tnote{1}$}
			&\phantom{\tiny a}
	        &\multicolumn{1}{c}{\normalsize  $\Delta_{max}$}
			&\multicolumn{1}{c}{\normalsize  $\Delta_{avg}$}
			&\multicolumn{1}{c}{\normalsize  $\Delta_{\sigma}\tnote{2}$}
			\\	
			[0.8ex]
			\cmidrule{6-11}
			\cmidrule{13-17}
			\cmidrule{19-22}
			\cmidrule{24-26}
			&\multicolumn{1}{c}{}
			&\phantom{a}
			&\multicolumn{1}{c}{}
			&\phantom{a}
			&\multicolumn{1}{c}{\small b}
			&\multicolumn{1}{c}{\small n}
			&\multicolumn{1}{c}{\small b}
			&\multicolumn{1}{c}{\small n}
			&\multicolumn{1}{c}{\small b}
			&\multicolumn{1}{c}{\small n}
			&\phantom{\tiny a}
			&\multicolumn{1}{c}{\small $N$}
			&\multicolumn{1}{c}{\small $N$}
			&\multicolumn{1}{c}{\small $N$}
			&\multicolumn{1}{c}{\small $N$}
			&\multicolumn{1}{c}{\small $N \cdot m$}
			&\phantom{\tiny a}
			&\multicolumn{1}{c}{\small $N$}
			&\multicolumn{1}{c}{\small $N$}
			&\multicolumn{1}{c}{\small $N$}
			&\phantom{\tiny a}
			&\phantom{a}
			&\multicolumn{1}{c}{\scriptsize  $[10^{-3}]$ mm}
			&\multicolumn{1}{c}{\scriptsize  $[10^{-3}]$ mm}
			&\multicolumn{1}{c}{\scriptsize  $[10^{-3}]$ mm}
			\\
			\specialrule{0.06em}{0.2em}{.2em}
\rowcolor{lightgray} &1 &place &1 &&1 &1 &-- &-- &-- &-- &&35.6 &-- &-- &1.8 &0.1 &&0.0 &17.6 &4.4 &50.0 &&0.0 &0.0 &0.0 \\
\rowcolor{lightgray} &2 &place &2 &&1 &1 &2 &2 &-- &-- &&36.6 &38.3 &-- &2.9 &0.5 &&0.0 &17.6 &2.5 &71.4 &&0.1 &0.0 &0.0 \\
&3 &release &2 &&1 &1 &-- &-- &-- &-- &&72.9 &-- &-- &-10.0 &0.7 &&-35.1 &17.6 &-7.5 &57.1 &&0.3 &0.1 &0.1 \\
&4 &place &3 &&1 &1 &3 &5 &-- &-- &&74.2 &40.3 &-- &-3.3 &0.8 &&-35.2 &17.9 &-3.3 &80.0 &&0.2 &0.1 &0.1 \\
&5 &release &3 &&1 &1 &-- &-- &-- &-- &&110.0 &-- &-- &-10.2 &0.5 &&-70.3 &17.5 &-19.3 &40.0 &&0.5 &0.2 &0.2 \\
&6 &place &4 &&1 &1 &4 &7 &-- &-- &&75.3 &70.3 &-- &-5.7 &0.3 &&-35.6 &51.9 &5.2 &61.5 &&0.2 &0.1 &0.1 \\
\rowcolor{lightgray} &7 &release &4 &&-- &-- &4 &7 &-- &-- &&-- &87.3 &-- &56.9 &3.6 &&-54.6 &68.3 &1.5 &53.8 &&0.4 &0.3 &0.2 \\
\rowcolor{lightgray} &8 &place &5 &&5 &8 &4 &7 &-- &-- &&25.7 &101.4 &-- &55.9 &3.0 &&-54.2 &68.6 &0.7 &43.8 &&0.4 &0.2 &0.1 \\
&9 &release & &&-- &-- &4 &7 &-- &-- &&-- &124.0 &-- &55.8 &2.9 &&-54.1 &68.6 &-2.9 &43.8 &&0.4 &0.2 &0.1 \\
&10 &place &6 &&6 &10 &4 &7 &-- &-- &&52.2 &109.3 &-- &55.0 &2.4 &&-53.8 &68.8 &1.4 &47.4 &&0.3 &0.1 &0.1 \\
&11 &release &6 &&-- &-- &4 &7 &-- &-- &&-- &161.6 &-- &53.9 &1.6 &&-69.2 &69.1 &-9.3 &42.1 &&0.6 &0.2 &0.2 \\
&12 &place &7 &&7 &13 &4 &7 &-- &-- &&68.6 &129.4 &-- &54.6 &2.0 &&-53.6 &68.9 &3.3 &50.0 &&0.3 &0.1 &0.1 \\
\rowcolor{lightgray} &13 &release &7 &&7 &13 &-- &-- &-- &-- &&127.7 &-- &-- &124.4 &11.5 &&-120.3 &105.4 &-8.7 &50.0 &&2.4 &1.5 &0.8 \\
\rowcolor{lightgray} &14 &place &8 &&7 &13 &8 &15 &-- &-- &&168.8 &56.5 &-- &114.7 &7.5 &&-112.3 &111.5 &-2.6 &48.0 &&1.4 &0.7 &0.5 \\
&15 &release &8 &&7 &13 &-- &-- &-- &-- &&165.9 &-- &-- &121.9 &10.4 &&-118.2 &107.0 &-8.9 &48.0 &&2.1 &1.2 &0.7 \\
&16 &place &9 &&7 &13 &9 &17 &-- &-- &&178.3 &36.1 &-- &116.3 &8.1 &&-113.6 &110.4 &-4.8 &50.0 &&1.5 &0.8 &0.5 \\
&17 &release &9 &&7 &13 &-- &-- &-- &-- &&206.3 &-- &-- &117.0 &8.4 &&-114.2 &110.0 &-10.4 &42.9 &&1.6 &0.9 &0.5 \\
&18 &place &10 &&7 &13 &10 &19 &-- &-- &&184.1 &64.8 &-- &114.9 &7.5 &&-112.5 &111.3 &-0.7 &48.4 &&1.4 &0.6 &0.5 \\
\rowcolor{lightgray} &19 &release &10 &&-- &-- &10 &19 &-- &-- &&-- &110.8 &-- &248.2 &21.7 &&-244.8 &72.2 &-65.0 &32.3 &&9.2 &5.3 &3.0 \\
\rowcolor{lightgray} &20 &place &11 &&11 &20 &10 &19 &-- &-- &&-46.0 &180.3 &-- &232.0 &19.8 &&-228.7 &85.2 &-48.1 &35.3 &&8.1 &4.3 &2.7 \\
&21 &release &11 &&-- &-- &10 &19 &-- &-- &&-- &154.4 &-- &240.6 &20.7 &&-237.2 &78.4 &-55.6 &32.4 &&8.5 &4.7 &2.8 \\
&22 &place &12 &&12 &22 &10 &19 &-- &-- &&60.5 &233.7 &-- &202.4 &15.6 &&-199.5 &108.8 &-25.4 &45.9 &&5.5 &2.6 &2.0 \\
&23 &release &12 &&-- &-- &10 &19 &-- &-- &&-- &205.9 &-- &224.9 &18.6 &&-221.8 &90.8 &-44.2 &32.4 &&7.3 &3.9 &2.3 \\
&24 &place &13 &&13 &25 &10 &19 &-- &-- &&58.9 &230.5 &-- &204.1 &15.9 &&-201.2 &107.4 &-22.4 &47.5 &&5.6 &2.5 &2.0 \\
\rowcolor{lightgray} &25 &release &13 &&13 &25 &-- &-- &-- &-- &&77.2 &-- &-- &416.5 &14.4 &&-416.2 &0.8 &-163.0 &10.0 &&12.9 &7.6 &4.2 \\
\rowcolor{lightgray} &26 &place &14 &&13 &25 &14 &26 &-- &-- &&362.2 &-203.2 &-- &346.4 &16.8 &&-345.4 &79.9 &-102.3 &16.3 &&10.7 &5.6 &3.8 \\
&27 &release &14 &&13 &25 &-- &-- &-- &-- &&112.9 &-- &-- &405.9 &14.8 &&-405.5 &12.8 &-144.2 &11.6 &&12.9 &7.1 &4.1 \\
&28 &place &15 &&13 &25 &15 &29 &-- &-- &&218.1 &-40.4 &-- &365.3 &16.6 &&-364.6 &40.4 &-106.8 &19.6 &&11.5 &5.9 &4.0 \\
&29 &release &15 &&13 &25 &-- &-- &-- &-- &&161.7 &-- &-- &386.4 &15.5 &&-385.9 &25.7 &-121.4 &17.4 &&11.8 &6.4 &3.9 \\
&30 &place &16 &&13 &25 &16 &31 &-- &-- &&239.0 &-21.4 &-- &354.6 &16.6 &&-353.7 &61.2 &-89.9 &28.6 &&10.9 &5.2 &4.0 \\
\rowcolor{lightgray} &31 &release &16 &&-- &-- &16 &31 &-- &-- &&-- &143.9 &-- &463.0 &3.4 &&-463.0 &0.8 &-177.6 &12.2 &&10.3 &6.4 &3.2 \\
\rowcolor{lightgray} &32 &place &17 &&17 &33 &16 &31 &-- &-- &&-31.1 &171.4 &-- &454.7 &4.8 &&-454.7 &14.4 &-162.1 &15.4 &&10.5 &6.0 &3.5 \\
&33 &release &17 &&-- &-- &16 &31 &-- &-- &&-- &177.2 &-- &458.9 &4.6 &&-458.9 &29.6 &-163.8 &13.5 &&11.2 &6.4 &3.5 \\
&34 &place &18 &&18 &34 &16 &31 &-- &-- &&48.4 &202.7 &-- &440.5 &7.4 &&-440.5 &24.2 &-145.1 &14.5 &&10.5 &5.4 &3.8 \\
&35 &release &18 &&-- &-- &16 &31 &-- &-- &&-- &214.8 &-- &451.8 &5.3 &&-451.8 &59.3 &-148.8 &20.0 &&10.4 &5.7 &3.4 \\
&36 &place &19 &&19 &37 &16 &31 &-- &-- &&48.0 &208.9 &-- &442.5 &7.0 &&-442.5 &41.2 &-138.5 &17.2 &&10.5 &5.2 &3.8 \\
\rowcolor{lightgray} &37 &release &19 &&19 &37 &-- &-- &-- &-- &&236.3 &-- &-- &471.3 &1.1 &&-471.3 &0.8 &-176.8 &12.1 &&10.9 &6.6 &3.3 \\
\rowcolor{lightgray} &38 &place &20 &&19 &37 &20 &39 &-- &-- &&256.1 &68.0 &-- &462.7 &2.2 &&-462.7 &3.5 &-163.8 &11.5 &&10.1 &5.6 &3.5 \\
&39 &release &20 &&19 &37 &-- &-- &-- &-- &&271.0 &-- &-- &469.6 &3.0 &&-469.6 &33.6 &-166.2 &13.1 &&13.2 &7.4 &4.1 \\
&40 &place &21 &&19 &37 &21 &41 &-- &-- &&277.3 &35.6 &-- &464.9 &2.4 &&-464.9 &35.1 &-156.3 &15.6 &&10.2 &5.5 &3.6 \\
&41 &release &21 &&19 &37 &-- &-- &-- &-- &&307.2 &-- &-- &466.7 &1.0 &&-466.7 &67.3 &-154.9 &17.2 &&10.5 &5.8 &3.4 \\
&42 &place &22 &&19 &37 &22 &43 &-- &-- &&279.3 &64.9 &-- &463.5 &2.3 &&-463.4 &36.2 &-150.0 &17.9 &&10.1 &5.2 &3.7 \\
\rowcolor{lightgray} &43 &release &22 &&-- &-- &22 &43 &-- &-- &&-- &338.1 &-- &472.6 &2.0 &&-472.6 &0.8 &-182.8 &10.4 &&11.3 &6.8 &3.4 \\
\rowcolor{lightgray} &44 &place &23 &&23 &44 &22 &43 &-- &-- &&26.0 &349.5 &-- &471.3 &1.3 &&-471.3 &10.5 &-174.2 &12.9 &&10.6 &6.2 &3.5 \\
&45 &release &23 &&-- &-- &22 &43 &-- &-- &&-- &373.4 &-- &471.7 &3.7 &&-471.7 &34.8 &-173.6 &12.9 &&15.9 &8.5 &4.8 \\
&46 &place &24 &&24 &46 &22 &43 &-- &-- &&52.1 &358.3 &-- &469.5 &1.4 &&-469.4 &17.8 &-166.5 &13.7 &&10.6 &5.7 &3.7 \\
&47 &release &24 &&-- &-- &22 &43 &-- &-- &&-- &409.3 &-- &470.3 &0.9 &&-470.2 &69.6 &-164.0 &16.4 &&11.1 &6.0 &3.6 \\
&48 &place &25 &&25 &49 &22 &43 &-- &-- &&69.4 &375.5 &-- &469.5 &0.5 &&-469.5 &35.2 &-158.5 &15.6 &&10.6 &5.4 &3.8 \\
\rowcolor{lightgray} &49 &release &25 &&-- &-- &-- &-- &-- &-- && & &-- & &0.7 &&-471.9 &0.8 &-204.7 &10.4 &&11.9 &7.4 &3.6 \\
			\specialrule{0.10em}{0.2em}{.2em}
		\end{tabular}
		
        \begin{tablenotes}
            \scriptsize
            \sethlcolor{lightgray} \item[]  \hl{------} = same in cantilever and sequential
            \item[1] \% of total edges in the analysis model that are in tension
            \item[2] standard deviation of all nodal displacements
        \end{tablenotes}
		\end{threeparttable}	
			
	\label{table:experimental_data2}
	\end{table}
    
\end{landscape}


\renewcommand{\thetable}{B.1}

\clearpage
\onecolumn
\begin{landscape}

\section{Optimized Three Robot Method Analysis Results} \label{sec:appendixB}
    
    \begin{scriptsize}
    \begin{longtable}{c cccr ccccccr cccccr ccccr ccc}
    %
    \caption{Analysis results for the three-robot sequence}\\
    \specialrule{.10em}{0.2em}{.2em}
	&\multirow{2}{*}{\normalsize  Step}
	&\multirow{2}{*}{\normalsize  Type}
	&\multirow{2}{*}{\normalsize \makecell{Total \\ Bricks}}
	&\phantom{\makecell{\vspace{0.7em}}}
	&\multicolumn{2}{c}{\normalsize $rob1$}
	&\multicolumn{2}{c}{\normalsize $rob2$}
	&\multicolumn{2}{c}{\normalsize $rob3$}
	&\phantom{a}
	&\multicolumn{1}{c}{\normalsize  $F_{rob1}$}
	&\multicolumn{1}{c}{\normalsize  $F_{rob2}$}
	&\multicolumn{1}{c}{\normalsize  $F_{rob3}$}
	&\multicolumn{1}{c}{\normalsize  $F_{sup}$}
	&\multicolumn{1}{c}{\normalsize  $M_{sup}$}
	&\phantom{a}
    &\multicolumn{1}{c}{\normalsize  $F_{min}$}
	&\multicolumn{1}{c}{\normalsize  $F_{max}$}
	&\multicolumn{1}{c}{\normalsize  $F_{avg}$}
	&\multicolumn{1}{c}{\normalsize  $T_{\%}^1$}
	&\phantom{a}
    &\multicolumn{1}{c}{\normalsize  $\Delta_{max}$}
	&\multicolumn{1}{c}{\normalsize  $\Delta_{avg}$}
	&\multicolumn{1}{c}{\normalsize  $\Delta_{\sigma}^2$}
	\\	
	[0.8ex]
	\cmidrule{6-11}
	\cmidrule{13-17}
	\cmidrule{19-22}
	\cmidrule{24-26}
	&\multicolumn{1}{c}{}
	&\phantom{a}
	&\multicolumn{1}{c}{}
	&\phantom{a}
	&\multicolumn{1}{c}{\small b}
	&\multicolumn{1}{c}{\small n}
	&\multicolumn{1}{c}{\small b}
	&\multicolumn{1}{c}{\small n}
	&\multicolumn{1}{c}{\small b}
	&\multicolumn{1}{c}{\small n}
	&\phantom{a}
	&\multicolumn{1}{c}{\small $N$}
	&\multicolumn{1}{c}{\small $N$}
	&\multicolumn{1}{c}{\small $N$}
	&\multicolumn{1}{c}{\small $N$}
	&\multicolumn{1}{c}{\small $N \cdot m$}
	&\phantom{a}
	&\multicolumn{1}{c}{\small $N$}
	&\multicolumn{1}{c}{\small $N$}
	&\multicolumn{1}{c}{\small $N$}
	&\phantom{a}
	&\phantom{a}
	&\multicolumn{1}{c}{\scriptsize  $[10^{-3}]$ mm}
	&\multicolumn{1}{c}{\scriptsize  $[10^{-3}]$ mm}
	&\multicolumn{1}{c}{\scriptsize  $[10^{-3}]$ mm}
	\\	
    \specialrule{.10em}{0.2em}{.2em}
    \endhead
    \specialrule{.10em}{0.2em}{.2em}
    &\multicolumn{10}{l}{\sethlcolor{babyblue} \hl{------} = optimized support location}\\
    &\multicolumn{10}{l}{$1$ \% of total edges in the analysis model that are in tension}\\
    &\multicolumn{10}{l}{$2$ standard deviation of all nodal displacements}
    \endfoot
    %
&	1&place&1	&&	1&1	&	--&--	&	--&--	&&	35.6&--&--&1.8&0.1	&&	0&17.6&4.4&50	&&	0&0&0	\\
&	2&place&2	&&	1&1	&	2&2	&	--&--	&&	36.6&38.3&--&2.9&0.5	&&	0&17.6&2.5&71.4	&&	0.1&0&0	\\
&	3&place&3	&&	1&1	&	2&2	&	\cellcolor{babyblue}3&\cellcolor{babyblue}4	&&	35.8&37.4&36.4&1.7&0.1	&&	-0.2&17.6&1.7&40	&&	0.1&0&0	\\
&	4&release&3	&&	--&--	&	2&2	&	\cellcolor{babyblue}3&\cellcolor{babyblue}4	&&	--&54.6&36.1&18.1&0.4	&&	-17.6&17.6&0&50	&&	0.1&0&0	\\
&	5&place&4	&&	4&7	&	2&2	&	\cellcolor{babyblue}3&\cellcolor{babyblue}4	&&	35.5&54.1&36.7&18.1&0.4	&&	-17.6&17.6&1.3&53.8	&&	0.1&0&0	\\
&	6&release&4	&&	4&7	&	--&--	&	\cellcolor{babyblue}3&\cellcolor{babyblue}4	&&	35.2&--&73.4&35.9&0.8	&&	-35.4&35.1&-0.1&46.2	&&	0.2&0.1&0.1	\\
&	7&reposition&4	&&	4&7	&	\cellcolor{babyblue}2&\cellcolor{babyblue}2	&	\cellcolor{babyblue}3&\cellcolor{babyblue}4	&&	35.5&54.1&36.7&18.1&0.4	&&	-17.6&17.6&1.3&53.8	&&	0.1&0&0	\\
&	8&release&4	&&	4&7	&	\cellcolor{babyblue}2&\cellcolor{babyblue}2	&	--&--	&&	52.9&73.7&--&17.9&0.4	&&	-17.9&34.6&2.6&61.5	&&	0.1&0&0	\\
&	9&place&5	&&	4&7	&	\cellcolor{babyblue}2&\cellcolor{babyblue}2	&	5&8	&&	64.3&72.9&25.7&18&0.4	&&	-17.8&34.6&1.3&56.3	&&	0.1&0&0	\\
&	10&release&5	&&	--&--	&	\cellcolor{babyblue}2&\cellcolor{babyblue}2	&	5&8	&&	--&112.2&52.6&17.9&1.3	&&	-55.3&17.6&-6.9&43.8	&&	0.4&0.2&0.2	\\
&	11&reposition&5	&&	\cellcolor{babyblue}4&\cellcolor{babyblue}6	&	\cellcolor{babyblue}2&\cellcolor{babyblue}2	&	5&8	&&	65&71.2&26.4&18.1&0.5	&&	-27&17.6&-2.6&50	&&	0.1&0&0	\\
&	12&release&5	&&	\cellcolor{babyblue}4&\cellcolor{babyblue}6	&	--&--	&	5&8	&&	100.8&--&26.6&53.6&1.3	&&	-53.1&52.2&-4.8&37.5	&&	0.3&0.1&0.1	\\
&	13&place&6	&&	\cellcolor{babyblue}4&\cellcolor{babyblue}6	&	6&11	&	5&8	&&	98.1&30.6&34.7&53.8&1.4	&&	-53.1&52.2&-4&36.8	&&	0.3&0.1&0.1	\\
&	14&release&6	&&	\cellcolor{babyblue}4&\cellcolor{babyblue}6	&	6&11	&	--&--	&&	128.9&35.4&--&53&1.5	&&	-54.3&52.4&-8.3&36.8	&&	0.5&0.2&0.1	\\
&	15&reposition&6	&&	\cellcolor{babyblue}4&\cellcolor{babyblue}6	&	6&11	&	\cellcolor{babyblue}5&\cellcolor{babyblue}8	&&	98.1&30.6&34.7&53.8&1.4	&&	-53.1&52.2&-4&36.8	&&	0.3&0.1&0.1	\\
&	16&release&6	&&	--&--	&	6&11	&	\cellcolor{babyblue}5&\cellcolor{babyblue}8	&&	--&54.5&76.9&85.2&5.4	&&	-82.9&57.7&-6.6&47.4	&&	0.9&0.4&0.3	\\
&	17&place&7	&&	7&13	&	6&11	&	\cellcolor{babyblue}5&\cellcolor{babyblue}8	&&	47.7&57&70.7&82.9&4.4	&&	-81.2&59.2&-2&54.5	&&	0.7&0.3&0.3	\\
&	18&release&7	&&	7&13	&	--&--	&	\cellcolor{babyblue}5&\cellcolor{babyblue}8	&&	83.3&--&76.6&94.4&8	&&	-91.6&62.8&-3.5&54.5	&&	1.7&0.8&0.5	\\
&	19&reposition&7	&&	7&13	&	\cellcolor{babyblue}4&\cellcolor{babyblue}6	&	\cellcolor{babyblue}5&\cellcolor{babyblue}8	&&	58.2&110.4&31.9&53.1&1	&&	-52.8&52.4&-1.9&45.5	&&	0.5&0.2&0.1	\\
&	20&release&7	&&	7&13	&	\cellcolor{babyblue}4&\cellcolor{babyblue}6	&	--&--	&&	67.8&131.3&--&53&0.7	&&	-54.7&52.4&-2.6&45.5	&&	0.4&0.2&0.1	\\
&	21&place&8	&&	7&13	&	\cellcolor{babyblue}4&\cellcolor{babyblue}6	&	8&15	&&	76.8&128.4&34.1&53.3&1.2	&&	-53.4&52.3&-2&44	&&	0.3&0.1&0.1	\\
&	22&release&8	&&	--&--	&	\cellcolor{babyblue}4&\cellcolor{babyblue}6	&	8&15	&&	--&151&84.5&52.6&0.7	&&	-73.6&52.6&-4.5&48	&&	0.5&0.2&0.2	\\
&	23&reposition&8	&&	\cellcolor{babyblue}3&\cellcolor{babyblue}4	&	\cellcolor{babyblue}4&\cellcolor{babyblue}6	&	8&15	&&	70.1&104.5&84.6&35.7&0.9	&&	-73.4&46.4&-3.8&44	&&	0.5&0.2&0.2	\\
&	24&release&8	&&	\cellcolor{babyblue}3&\cellcolor{babyblue}4	&	--&--	&	8&15	&&	155.5&--&98.7&34.8&0.9	&&	-77.7&59.6&-4.5&48	&&	1.1&0.4&0.4	\\
&	25&place&9	&&	\cellcolor{babyblue}3&\cellcolor{babyblue}4	&	9&17	&	8&15	&&	152.5&25.5&118&34.8&0.4	&&	-75.8&61&-3.6&50	&&	1&0.3&0.3	\\
&	26&release&9	&&	\cellcolor{babyblue}3&\cellcolor{babyblue}4	&	9&17	&	--&--	&&	212.7&79.6&--&35.1&4.6	&&	-135&35.9&-25.4&46.4	&&	4.1&1.3&1.1	\\
&	27&reposition&9	&&	\cellcolor{babyblue}3&\cellcolor{babyblue}4	&	9&17	&	\cellcolor{babyblue}4&\cellcolor{babyblue}6	&&	75.6&70.9&162.2&35.5&3.7	&&	-124.9&35.2&-19.2&35.7	&&	3.3&0.8&0.8	\\
&	28&release&9	&&	--&--	&	9&17	&	\cellcolor{babyblue}4&\cellcolor{babyblue}6	&&	--&69.6&204.1&51.9&4	&&	-126.5&53&-20.4&35.7	&&	3.4&0.9&0.8	\\
&	29&place&10	&&	10&19	&	9&17	&	\cellcolor{babyblue}4&\cellcolor{babyblue}6	&&	52.9&57.8&199&51.7&3	&&	-118.7&53.1&-13.7&48.4	&&	2.7&0.8&0.7	\\
&	30&release&10	&&	10&19	&	--&--	&	\cellcolor{babyblue}4&\cellcolor{babyblue}6	&&	85.9&--&224.9&51.7&2.3	&&	-142.5&53.7&-19.9&38.7	&&	2.7&1.2&1	\\
&	31&reposition&10	&&	10&19	&	\cellcolor{babyblue}6&\cellcolor{babyblue}11	&	\cellcolor{babyblue}4&\cellcolor{babyblue}6	&&	76.4&75.1&160.2&52.2&3.2	&&	-82.8&52.8&-12.6&41.9	&&	2.7&0.7&0.7	\\
&	32&release&10	&&	10&19	&	\cellcolor{babyblue}6&\cellcolor{babyblue}11	&	--&--	&&	93&131.5&--&137.2&11	&&	-134.8&64.8&-19.3&41.9	&&	4.6&1.7&1.1	\\
&	33&place&11	&&	10&19	&	\cellcolor{babyblue}6&\cellcolor{babyblue}11	&	11&20	&&	129.3&141.8&53.4&126.6&8.6	&&	-124.7&64.8&-13.5&38.2	&&	3.1&1.2&0.8	\\
&	34&release&11	&&	--&--	&	\cellcolor{babyblue}6&\cellcolor{babyblue}11	&	11&20	&&	--&176&74.7&147.4&8.3	&&	-145.4&29.5&-37.1&26.5	&&	3.7&1.7&1	\\
&	35&reposition&11	&&	\cellcolor{babyblue}5&\cellcolor{babyblue}8	&	\cellcolor{babyblue}6&\cellcolor{babyblue}11	&	11&20	&&	116.9&129.5&68&85.7&3.4	&&	-109.3&55.3&-20.5&38.2	&&	2.6&0.9&0.7	\\
&	36&release&11	&&	\cellcolor{babyblue}5&\cellcolor{babyblue}8	&	--&--	&	11&20	&&	174.9&--&81.3&143.4&15.7	&&	-140.4&28.8&-42.8&32.4	&&	11.1&3.8&2.7	\\
&	37&place&12	&&	\cellcolor{babyblue}5&\cellcolor{babyblue}8	&	12&23	&	11&20	&&	184.8&52.5&80.9&127&9.6	&&	-124.9&38.6&-31.7&35.1	&&	5.1&2&1.5	\\
&	38&release&12	&&	\cellcolor{babyblue}5&\cellcolor{babyblue}8	&	12&23	&	--&--	&&	209.6&85.4&--&143.5&9.4	&&	-164.9&26.4&-50.2&21.6	&&	5&2.5&1.6	\\
&	39&reposition&12	&&	\cellcolor{babyblue}5&\cellcolor{babyblue}8	&	12&23	&	\cellcolor{babyblue}11&\cellcolor{babyblue}20	&&	184.8&52.5&80.9&127&9.6	&&	-124.9&38.6&-31.7&35.1	&&	5.1&2&1.5	\\
&	40&release&12	&&	--&--	&	12&23	&	\cellcolor{babyblue}11&\cellcolor{babyblue}20	&&	--&46.4&79&307.3&21.5	&&	-305.2&28.7&-92.1&24.3	&&	11.5&6.1&4	\\
&	41&place&13	&&	13&25	&	12&23	&	\cellcolor{babyblue}11&\cellcolor{babyblue}20	&&	-62.8&106.2&126.3&262&18.2	&&	-259.8&67.2&-53.3&40	&&	8.4&4&3.1	\\
&	42&release&13	&&	13&25	&	--&--	&	\cellcolor{babyblue}11&\cellcolor{babyblue}20	&&	28.8&--&125.8& &19.3	&&	-323.4&16.6&-101&15	&&	11.9&5.9&3.6	\\
&	43&reposition&13	&&	13&25	&	\cellcolor{babyblue}6&\cellcolor{babyblue}11	&	\cellcolor{babyblue}11&\cellcolor{babyblue}20	&&	43.9&188&96.7&142.3&6.3	&&	-140.8&33.5&-31.9&30	&&	4&1.6&1.1	\\
&	44&release&13	&&	13&25	&	\cellcolor{babyblue}6&\cellcolor{babyblue}11	&	--&--	&&	69.9&229.4&--&189.9&13.5	&&	-200.8&31.4&-79.2&12.5	&&	8.9&4&2.7	\\
&	45&place&14	&&	13&25	&	\cellcolor{babyblue}6&\cellcolor{babyblue}11	&	14&26	&&	267.9&207.4&-130.3&156.6&9.3	&&	-154.5&50.8&-42.7&25.6	&&	5.4&2.1&1.5	\\
&	46&release&14	&&	--&--	&	\cellcolor{babyblue}6&\cellcolor{babyblue}11	&	14&26	&&	--&239.9&91.3&203.9&14.8	&&	-226.6&34.3&-91.2&11.6	&&	11.3&4.8&3.2	\\
&	47&reposition&14	&&	\cellcolor{babyblue}8&\cellcolor{babyblue}14	&	\cellcolor{babyblue}6&\cellcolor{babyblue}11	&	14&26	&&	170.4&149.6&82&118.3&4.8	&&	-148.3&57.7&-38.6&25.6	&&	3.7&1.3&1	\\
&	48&release&14	&&	\cellcolor{babyblue}8&\cellcolor{babyblue}14	&	--&--	&	14&26	&&	209.8&--&81.1&226.3&16.3	&&	-224.1&18.2&-69.4&18.6	&&	8.7&4&2.1	\\
&	49&place&15	&&	\cellcolor{babyblue}8&\cellcolor{babyblue}14	&	15&29	&	14&26	&&	219.4&-68.6&146.8&199.5&13.4	&&	-197.5&63.9&-48.1&21.7	&&	5.2&2.5&1.5	\\
&	50&release&15	&&	\cellcolor{babyblue}8&\cellcolor{babyblue}14	&	15&29	&	--&--	&&	254.1&104.4&--&198.7&12.5	&&	-196.8&44.7&-62.8&21.7	&&	4.5&2.8&1.2	\\
&	51&reposition&15	&&	\cellcolor{babyblue}8&\cellcolor{babyblue}14	&	15&29	&	\cellcolor{babyblue}15&\cellcolor{babyblue}28	&&	241.9&87.9&54&207.4&13.7	&&	-205.5&36.4&-64.9&19.6	&&	5.2&2.8&1.5	\\
&	52&release&15	&&	--&--	&	15&29	&	\cellcolor{babyblue}15&\cellcolor{babyblue}28	&&	--&7&115.7&448.9&7.7	&&	-448.8&0.8&-171.1&8.7	&&	11.2&6.5&3.8	\\
&	53&place&16	&&	16&31	&	15&29	&	\cellcolor{babyblue}15&\cellcolor{babyblue}28	&&	-52&11&186.6&418.3&10.9	&&	-418.1&20.1&-140.8&14.3	&&	11.1&5.8&4	\\
&	54&release&16	&&	16&31	&	--&--	&	\cellcolor{babyblue}15&\cellcolor{babyblue}28	&&	-52&--&197.6&418.3&10.9	&&	-418.1&20.2&-140.8&16.3	&&	11.1&5.8&4	\\
&	55&reposition&16	&&	16&31	&	\cellcolor{babyblue}15&\cellcolor{babyblue}29	&	\cellcolor{babyblue}15&\cellcolor{babyblue}28	&&	-52&11&186.6&418.3&10.9	&&	-418.1&20.1&-140.8&14.3	&&	11.1&5.8&4	\\
&	56&release&16	&&	16&31	&	\cellcolor{babyblue}15&\cellcolor{babyblue}29	&	--&--	&&	93&94.9&--&440&7.8	&&	-439.9&31.2&-159.1&12.2	&&	10.8&6.1&3.7	\\
&	57&place&17	&&	16&31	&	\cellcolor{babyblue}15&\cellcolor{babyblue}29	&	17&33	&&	113.7&115.6&61.2&435&8.3	&&	-434.9&20.1&-146.9&15.4	&&	10.7&5.7&3.8	\\
&	58&release&17	&&	--&--	&	\cellcolor{babyblue}15&\cellcolor{babyblue}29	&	17&33	&&	--&126.3&69&438.6&7.9	&&	-438.6&17.9&-147.7&11.5	&&	10.7&5.8&3.7	\\
&	59&reposition&17	&&	\cellcolor{babyblue}16&\cellcolor{babyblue}31	&	\cellcolor{babyblue}15&\cellcolor{babyblue}29	&	17&33	&&	113.7&115.6&61.2&435&8.3	&&	-434.9&20.1&-146.9&15.4	&&	10.7&5.7&3.8	\\
&	60&release&17	&&	\cellcolor{babyblue}16&\cellcolor{babyblue}31	&	--&--	&	17&33	&&	171.4&--&-31.1&454.7&4.8	&&	-454.7&14.4&-162.1&15.4	&&	10.5&6&3.5	\\
&	61&place&18	&&	\cellcolor{babyblue}16&\cellcolor{babyblue}31	&	18&35	&	17&33	&&	206.7&30.5&-36&440.4&7.3	&&	-440.3&19.5&-145.1&16.4	&&	10.5&5.4&3.8	\\
&	62&release&18	&&	\cellcolor{babyblue}16&\cellcolor{babyblue}31	&	18&35	&	--&--	&&	200.9&31.2&--&445.9&6.8	&&	-445.9&34.8&-147.5&18.2	&&	10.6&5.6&3.7	\\
&	63&reposition&18	&&	\cellcolor{babyblue}16&\cellcolor{babyblue}31	&	18&35	&	\cellcolor{babyblue}17&\cellcolor{babyblue}32	&&	221.5&22.8&-85.1&439.2&7.5	&&	-439.2&15.8&-144.5&14.5	&&	10.5&5.3&3.8	\\
&	64&release&18	&&	--&--	&	18&35	&	\cellcolor{babyblue}17&\cellcolor{babyblue}32	&&	--&27.8&194.4&459.8&4.3	&&	-459.8&12.1&-161.7&14.5	&&	10.4&5.9&3.5	\\
&	65&place&19	&&	19&37	&	18&35	&	\cellcolor{babyblue}17&\cellcolor{babyblue}32	&&	31.4&20.3&203.1&451.9&5.1	&&	-451.9&13&-149.3&15.5	&&	10.2&5.3&3.7	\\
&	66&release&19	&&	19&37	&	--&--	&	\cellcolor{babyblue}17&\cellcolor{babyblue}32	&&	40.7&--&212.3&451.9&5.1	&&	-451.9&22.7&-149.3&13.8	&&	10.2&5.3&3.7	\\
&	67&reposition&19	&&	19&37	&	\cellcolor{babyblue}15&\cellcolor{babyblue}29	&	\cellcolor{babyblue}17&\cellcolor{babyblue}32	&&	45.9&107.7&113.9&433&8.5	&&	-432.9&21&-132.2&13.8	&&	10.7&5.1&3.9	\\
&	68&release&19	&&	19&37	&	\cellcolor{babyblue}15&\cellcolor{babyblue}29	&	--&--	&&	139.6&128.7&--&431.6&8.6	&&	-431.5&16.9&-138.3&12.1	&&	10.7&5.2&3.8	\\
&	69&place&20	&&	19&37	&	\cellcolor{babyblue}15&\cellcolor{babyblue}29	&	20&39	&&	139.8&129.7&35.4&432.8&8.5	&&	-432.7&18.4&-131.3&11.5	&&	10.7&5&3.8	\\
&	70&release&20	&&	--&--	&	\cellcolor{babyblue}15&\cellcolor{babyblue}29	&	20&39	&&	--&139.6&162.9&431.4&8.6	&&	-431.3&18.2&-133.8&9.8	&&	10.7&5.1&3.7	\\
&	71&reposition&20	&&	\cellcolor{babyblue}17&\cellcolor{babyblue}33	&	\cellcolor{babyblue}15&\cellcolor{babyblue}29	&	20&39	&&	89.9&132.3&83.1&429.5&9.2	&&	-429.4&24.1&-123.3&16.4	&&	10.9&4.9&4	\\
&	72&release&20	&&	\cellcolor{babyblue}17&\cellcolor{babyblue}33	&	--&--	&	20&39	&&	129.1&--&155.9&457.9&4	&&	-457.9&31.3&-155.2&11.5	&&	10.2&5.5&3.5	\\
&	73&place&21	&&	\cellcolor{babyblue}17&\cellcolor{babyblue}33	&	21&41	&	20&39	&&	136.5&43.3&178.2&456.5&4	&&	-456.5&39.3&-146.7&14.1	&&	10.4&5.2&3.7	\\
&	74&release&21	&&	\cellcolor{babyblue}17&\cellcolor{babyblue}33	&	21&41	&	--&--	&&	201.3&114.7&--&456.5&4.2	&&	-456.5&20.5&-141.4&15.6	&&	12.8&6.3&3.9	\\
&	75&reposition&21	&&	\cellcolor{babyblue}17&\cellcolor{babyblue}33	&	21&41	&	\cellcolor{babyblue}15&\cellcolor{babyblue}29	&&	124.4&72.3&144&428.1&9.4	&&	-428&61.8&-112&18.8	&&	11.1&4.8&3.9	\\
&	76&release&21	&&	--&--	&	21&41	&	\cellcolor{babyblue}15&\cellcolor{babyblue}29	&&	--&154.1&194.4&428.4&10	&&	-428.2&40.7&-121&17.2	&&	12.1&6.7&3.4	\\
&	77&place&22	&&	22&43	&	21&41	&	\cellcolor{babyblue}15&\cellcolor{babyblue}29	&&	126.4&107.2&149.7&427&9	&&	-426.9&20.2&-128&10.4	&&	10.7&5&3.5	\\
&	78&release&22	&&	22&43	&	--&--	&	\cellcolor{babyblue}15&\cellcolor{babyblue}29	&&	227&--&148.2&426.8&9.1	&&	-426.6&10.9&-132.8&10.4	&&	10.9&5.2&3.4	\\
&	79&reposition&22	&&	22&43	&	\cellcolor{babyblue}18&\cellcolor{babyblue}35	&	\cellcolor{babyblue}15&\cellcolor{babyblue}29	&&	153.4&96.2&129.6&430.8&8.8	&&	-430.7&31.1&-125.5&9	&&	11&5.1&3.6	\\
&	80&release&22	&&	22&43	&	\cellcolor{babyblue}18&\cellcolor{babyblue}35	&	--&--	&&	188.2&161.5&--&460.6&2.9	&&	-460.6&22.5&-154.2&10.4	&&	11.4&5.7&3.7	\\
&	81&place&23	&&	22&43	&	\cellcolor{babyblue}18&\cellcolor{babyblue}35	&	23&44	&&	215.6&174.5&57.3&459.7&3.3	&&	-459.7&33.4&-145.7&12.9	&&	11.3&5.3&3.8	\\
&	82&release&23	&&	--&--	&	\cellcolor{babyblue}18&\cellcolor{babyblue}35	&	23&44	&&	--&243&141.8&457.8&6.2	&&	-457.8&27.1&-137.7&15.7	&&	17&8.2&5	\\
&	83&reposition&23	&&	\cellcolor{babyblue}14&\cellcolor{babyblue}27	&	\cellcolor{babyblue}18&\cellcolor{babyblue}35	&	23&44	&&	164.4&154.5&108.9&409&12.1	&&	-408.7&62.2&-94.1&24.3	&&	12.1&4.9&3.7	\\
&	84&release&23	&&	\cellcolor{babyblue}14&\cellcolor{babyblue}27	&	--&--	&	23&44	&&	213.5&--&225.3&401.6&15.6	&&	-401.2&52&-108.8&22.9	&&	19.6&9.7&4.7	\\
&	85&place&24	&&	\cellcolor{babyblue}14&\cellcolor{babyblue}27	&	24&47	&	23&44	&&	165.5&113.9&196.7&404.2&12.4	&&	-403.9&36.9&-118.6&13.7	&&	11.4&5.3&3.3	\\
&	86&release&24	&&	\cellcolor{babyblue}14&\cellcolor{babyblue}27	&	24&47	&	--&--	&&	224.6&257.2&--&385.6&16.2	&&	-385.1&33&-103.2&20.5	&&	27.9&12.9&7.4	\\
&	87&reposition&24	&&	\cellcolor{babyblue}14&\cellcolor{babyblue}27	&	24&47	&	\cellcolor{babyblue}17&\cellcolor{babyblue}33	&&	151.8&150.5&175.6&416.8&12.4	&&	-416.6&87.7&-96&23.3	&&	12&6&3.4	\\
&	88&release&24	&&	--&--	&	24&47	&	\cellcolor{babyblue}17&\cellcolor{babyblue}33	&&	--&170&254.1&452.4&7.5	&&	-452.4&63.8&-125.6&21.9	&&	21.9&10.2&5.8	\\
&	89&place&25	&&	25&49	&	24&47	&	\cellcolor{babyblue}17&\cellcolor{babyblue}33	&&	191.8&115.5&199.1&455.7&4.3	&&	-455.7&82.4&-136.3&14.3	&&	13.4&5.8&4.1	\\
&	90&release&25	&&	25&49	&	--&--	&	\cellcolor{babyblue}17&\cellcolor{babyblue}33	&&	270.1&--&200.4&455.4&4.3	&&	-455.4&30&-140&16.9	&&	13.5&5.9&3.9	\\

    \label{table:experimental_data3}
    \end{longtable}
    \end{scriptsize}

\end{landscape}

%% file: RobotFab_Paper.bbl
\begin{thebibliography}{100}
\expandafter\ifx\csname url\endcsname\relax
  \def\url#1{\texttt{#1}}\fi
\expandafter\ifx\csname urlprefix\endcsname\relax\def\urlprefix{URL }\fi
\expandafter\ifx\csname href\endcsname\relax
  \def\href#1#2{#2} \def\path#1{#1}\fi

\bibitem{deuss_assembling_2014}
M.~Deuss, D.~Panozzo, E.~Whiting, Y.~Liu, P.~Block, O.~Sorkine-Hornung,
  M.~Pauly, Assembling self-supporting structures, ACM Transactions on Graphics
  33~(6) (2014) 1--10.
\newblock \href {https://doi.org/10.1145/2661229.2661266}
  {\path{doi:10.1145/2661229.2661266}}.

\bibitem{paris_statics_2020}
V.~Paris, A.~Pizzigoni, S.~Adriaenssens, Statics of {Self}-{Balancing}
  {Masonry} {Domes} {Constructed} with a {Cross}-{Herringbone} {Spiraling}
  {Pattern}, Engineering Structures 215 (Jul. 2020).
\newblock \href {https://doi.org/10.1016/j.engstruct.2020.110440}
  {\path{doi:10.1016/j.engstruct.2020.110440}}.

\bibitem{choisy_art_1883}
A.~Choisy, L'{Art} de bâtir chez les {Byzantins}, Société anonyme de
  publications périodiques, Paris, 1883.

\bibitem{el-naggar_les_1999}
S.~El-Naggar, Les voutes dans l'architecture de l'{Egypte} ancienne, Institut
  français d'archéologie orientale, Le Caire, 1999.

\bibitem{besenval_technologie_1984}
R.~Besenval, Technologie de la {Voûte} dans l'{Orient} {Ancien}, Éditions
  Recherche sur les Civilisations, Paris, 1984.

\bibitem{heyman_stone_1997}
J.~Heyman, The {Stone} {Skeleton}: {Structural} {Engineering} of {Masonry}
  {Architecture}, reprint Edition, Cambridge University Press, 1997.

\bibitem{pizzigoni_brunelleschis_2015}
A.~Pizzigoni, Brunelleschi's {Bricks}, Journal of the International Association
  for Shell and Spatial Structures 56~(2) (2015) 1--12.

\bibitem{pizzigoni_herringbone_2018}
A.~Pizzigoni, V.~Paris, M.~Pasta, M.~Morandi, A.~Parsani, Herringbone {Naked}
  {Structure}, in: Proceedings of {IASS} {Annual} {Symposia} 2018, Boston,
  2018, pp. 1--6.

\bibitem{ochsendorf_guastavino_2010}
J.~A. Ochsendorf, Guastavino vaulting: the art of structural tile, Princeton
  Architectural Press, New York, 2010.

\bibitem{vouga_design_2012}
E.~Vouga, M.~Höbinger, J.~Wallner, H.~Pottmann, Design of {Self}-{Supporting}
  {Surfaces}, ACM Transactions on Graphics 31~(4) (2012) 1--11.

\bibitem{goessens_feasibility_2018}
S.~Goessens, C.~Mueller, P.~Latteur, Feasibility study for drone-based masonry
  construction of real-scale structures, Automation in Construction 94 (2018)
  458--480.
\newblock \href {https://doi.org/10.1016/j.autcon.2018.06.015}
  {\path{doi:10.1016/j.autcon.2018.06.015}}.

\bibitem{som_bricks_2019}
{SOM}, {University of Alcalá},
  \href{https://som.medium.com/bricks-mortar-and-robots-solutions-for-sustainable-construction-f404b90fe9ab}{Bricks,
  {Mortar}, and {Robots}: {Solutions} for {Sustainable} {Construction}} (Aug.
  2019).
\newline\urlprefix\url{https://som.medium.com/bricks-mortar-and-robots-solutions-for-sustainable-construction-f404b90fe9ab}

\bibitem{loing_free-form_2020}
V.~Loing, O.~Baverel, J.-F. Caron, R.~Mesnil, Free-form structures from
  topologically interlocking masonries, Automation in Construction 113 (May
  2020).
\newblock \href {https://doi.org/10.1016/j.autcon.2020.103117}
  {\path{doi:10.1016/j.autcon.2020.103117}}.

\bibitem{drew_lock-block_2021}
J.~Drew, \href{http://www.lockblock.com/}{Lock-{Block} {Ltd}.} (2021).
\newline\urlprefix\url{http://www.lockblock.com/}

\bibitem{wu_robotic_2018}
K.~Wu, A.~Kilian, Robotic {Equilibrium}: {Scaffold} {Free} {Arch} {Assemblies},
  in: Proceedings of the 38th {Annual} {Conference} of the {Association} for
  {Computer} {Aided} {Design} in {Architecture}, Mexico City, 2018, pp. 1--9.

\bibitem{bock_robot-oriented_2015}
T.~Bock, T.~Linner, Robot-{Oriented} {Design}: {Design} and {Management}
  {Tools} for the {Deployment} of {Automation} and {Robotics} in
  {Construction}, Vol.~1, Cambridge University Press, 2015.

\bibitem{bravo-palacios_one_2020}
G.~Bravo-Palacios, A.~D. Prete, P.~M. Wensing, One {Robot} for {Many} {Tasks}:
  {Versatile} {Co}-{Design} {Through} {Stochastic} {Programming}, IEEE Robotics
  and Automation Letters 5~(2) (2020) 1680--1687.
\newblock \href {https://doi.org/10.1109/LRA.2020.2969948}
  {\path{doi:10.1109/LRA.2020.2969948}}.

\bibitem{eversmann_robotic_2017}
P.~Eversmann, F.~Gramazio, M.~Kohler, Robotic prefabrication of timber
  structures: towards automated large-scale spatial assembly, Construction
  Robotics 1 (2017) 49--60.
\newblock \href {https://doi.org/10.1007/s41693-017-0006-2}
  {\path{doi:10.1007/s41693-017-0006-2}}.

\bibitem{ifr_world_2018}
{IFR},
  \href{https://ifr.org/ifr-press-releases/news/global-industrial-robot-sales-doubled-over-the-past-five-years}{World
  {Robotics} {Report}} (2018).
\newline\urlprefix\url{https://ifr.org/ifr-press-releases/news/global-industrial-robot-sales-doubled-over-the-past-five-years}

\bibitem{han_concept_2020}
I.~X. Han, E.~P. Bruun, S.~Marsh, S.~Adriaenssens, S.~Parascho, From {Concept}
  to {Construction}: {A} {Transferable} {Design} and {Robotic} {Fabrication}
  {Method} for a {Building}-{Scale} {Vault}, in: Proceedings of the 40th
  {Annual} {Conference} of the {Association} for {Computer} {Aided} {Design} in
  {Architecture}, 2020, pp. 1--12, (To Appear).

\bibitem{parascho_robotic_2020}
S.~Parascho, I.~X. Han, S.~Walker, A.~Beghini, E.~P. Bruun, S.~Adriaenssens,
  Robotic {Vault}: {A} {Cooperative} {Robotic} {Assembly} {Method} for {Brick}
  {Vault} {Construction}, Construction Robotics 4~(3) (Nov. 2020).
\newblock \href {https://doi.org/10.1007/s41693-020-00041-w}
  {\path{doi:10.1007/s41693-020-00041-w}}.

\bibitem{parascho_lightvault_2021}
S.~Parascho, I.~X. Han, A.~Beghini, M.~Miki, S.~Walker, E.~P. Bruun,
  S.~Adriaenssens, {LightVault}: {A} {Design} and {Robotic} {Fabrication}
  {Method} for {Complex} {Masonry} {Structures}, in: Advances in
  {Architectural} {Geometry} 2021, Paris, France, 2021, p.~25, (To Appear).

\bibitem{parascho_cooperative_2017}
S.~Parascho, A.~Gandia, A.~Mirjan, F.~Gramazio, M.~Kohler, Cooperative
  {Fabrication} of {Spatial} {Metal} {Structures}, in: A.~Menges, B.~Sheil,
  R.~Glynn, M.~Skavara (Eds.), Fabricate 2017, UCL Press, London, 2017, pp.
  24--29.
\newblock \href {https://doi.org/10.3929/ethz-b-000219566}
  {\path{doi:10.3929/ethz-b-000219566}}.

\bibitem{sharif_bim_2015}
S.~Sharif, R.~Gentry, {BIM} for {Masonry}: {Development} of {BIM} {Plugins} for
  the {Masonry} {Unit} {Database}, in: Proceedings of the 33rd {Education} and
  {Research} in {Computer} {Aided} {Architectural} {Design} in {Europe}
  {Conference}, Vienna, Austria, 2015, pp. 567--576.

\bibitem{bock_construction_2007}
T.~Bock, Construction robotics, Autonomous Robots 22~(3) (2007) 201--209.
\newblock \href {https://doi.org/10.1007/s10514-006-9008-5}
  {\path{doi:10.1007/s10514-006-9008-5}}.

\bibitem{bock_future_2015}
T.~Bock, The future of construction automation: {Technological} disruption and
  the upcoming ubiquity of robotics, Automation in Construction 59 (2015)
  113--121.
\newblock \href {https://doi.org/10.1016/j.autcon.2015.07.022}
  {\path{doi:10.1016/j.autcon.2015.07.022}}.

\bibitem{albus_trip_1986}
J.~S. Albus, Trip report: {Japanese} progress in robotics for construction,
  Robotics 2~(2) (1986) 103--112, publisher: North-Holland.
\newblock \href {https://doi.org/10.1016/0167-8493(86)90047-1}
  {\path{doi:10.1016/0167-8493(86)90047-1}}.

\bibitem{cousineau_construction_1998}
L.~Cousineau, N.~Miura, Construction {Robots}: {The} {Search} for {New}
  {Building} {Technology} in {Japan}, American Society of Civil Engineers,
  Reston, VA, 1998.

\bibitem{huang_factor_1990}
P.~Y. Huang, M.~Sakurai, Factor automation: the {Japanese} experience, IEEE
  Transactions on Engineering Management 37~(2) (1990) 102--108.
\newblock \href {https://doi.org/10.1109/17.53712}
  {\path{doi:10.1109/17.53712}}.

\bibitem{ueno_construction_1986}
T.~Ueno, J.~Maeda, T.~Yoshida, S.~Suzuki, Construction {Robots} for {Site}
  {Automation}, in: Proceedings of the 3rd {ISARC}, Marseille, France, 1986,
  pp. 627--639.
\newblock \href {https://doi.org/10.22260/ISARC1986/0049}
  {\path{doi:10.22260/ISARC1986/0049}}.

\bibitem{bock_site_2016}
T.~Bock, T.~Linner, Site {Automation}: {Automated}/{Robotic} {On}-{Site}
  {Factories}, Vol.~3, Cambridge University Press, 2016.

\bibitem{skibniewski_robotics_1989}
M.~J. Skibniewski, Robotics in {Civil} {Engineering}, Van Nostrand Reinhold,
  Southampton, 1989.

\bibitem{davila_delgado_robotics_2019}
J.~M. Davila~Delgado, L.~Oyedele, A.~Ajayi, L.~Akanbi, O.~Akinade, M.~Bilal,
  H.~Owolabi, Robotics and automated systems in construction: {Understanding}
  industry-specific challenges for adoption, Journal of Building Engineering 26
  (Nov. 2019).
\newblock \href {https://doi.org/10.1016/j.jobe.2019.100868}
  {\path{doi:10.1016/j.jobe.2019.100868}}.

\bibitem{castro-lacouture_construction_2009}
D.~Castro-Lacouture, Construction {Automation}, in: S.~Y. Nof (Ed.), Springer
  {Handbook} of {Automation}, Springer {Handbooks}, Springer, Berlin,
  Heidelberg, 2009, pp. 1063--1078.

\bibitem{garcia_de_soto_productivity_2018}
B.~García~de Soto, I.~Agustí-Juan, J.~Hunhevicz, S.~Joss, K.~Graser,
  G.~Habert, B.~T. Adey, Productivity of digital fabrication in construction:
  {Cost} and time analysis of a robotically built wall, Automation in
  Construction 92 (2018) 297--311.
\newblock \href {https://doi.org/10.1016/j.autcon.2018.04.004}
  {\path{doi:10.1016/j.autcon.2018.04.004}}.

\bibitem{kumar_robotics_2016}
V.~R.~P. Kumar, M.~B. a. S.~J. Raj, Robotics in {Construction} {Industry},
  Indian Journal of Science and Technology 9~(23) (Jun. 2016).
\newblock \href {https://doi.org/10.17485/ijst/2016/v9i23/95974}
  {\path{doi:10.17485/ijst/2016/v9i23/95974}}.

\bibitem{bock_robotic_2015}
T.~Bock, T.~Linner, Robotic {Industrialization}: {Automation} and {Robotic}
  {Technologies} for {Customized} {Component}, {Module}, and {Building}
  {Prefabrication}, Vol.~2, Cambridge University Press, 2015.

\bibitem{block_structural_2018}
P.~Block, T.~Van~Mele, A.~Liew, M.~DeJong, D.~Escobedo, J.~Ochsendorf,
  Structural design, fabrication and construction of the {Armadillo} {Vault},
  The Structural Engineer 96~(5) (2018) 10--20.

\bibitem{rippmann_computational_2016}
M.~Rippmann, T.~V. Mele, M.~Popescu, E.~Augustynowicz, T.~M. Echenagucia, C.~C.
  Barentin, U.~Frick, P.~Block, Computational {Design} and {Digital}
  {Fabrication} of a {Freeform} {Stone} {Shell}, in: Advances in
  {Architectural} {Geometry} 2016, Zurich, 2016, pp. 344--363.

\bibitem{davis_innovative_2012}
L.~Davis, M.~Rippmann, T.~Pawlofsky, P.~Block, Innovative {Funicular} {Tile}
  {Vaulting}: {A} prototype in {Switzerland}, The Structural Engineer 90~(11)
  (2012) 46--56.

\bibitem{borne_droneport_2016}
E.~Borne, E.~e.~a. Heathcote, The {Droneport} {Project}, AA: L’Architecture
  d’Aujourd’hui (Nov. 2016).

\bibitem{chilton_heinz_2010}
J.~Chilton, Heinz {Isler}'s {Infinite} {Spectrum}: {Form}-{Finding} in
  {Design}, Architectural Design 80~(4) (2010) 64--71.
\newblock \href {https://doi.org/10.1002/ad.1108} {\path{doi:10.1002/ad.1108}}.

\bibitem{chilton_rooted_2017}
J.~Chilton, C.-C. Chuang, Rooted in {Nature}: {Aesthetics}, {Geometry} and
  {Structure} in the {Shells} of {Heinz} {Isler}, Nexus Network Journal 19~(3)
  (2017) 763--785.
\newblock \href {https://doi.org/10.1007/s00004-017-0357-5}
  {\path{doi:10.1007/s00004-017-0357-5}}.

\bibitem{bock_construction_2016}
T.~Bock, T.~Linner, Construction {Robots}: {Elementary} {Technologies} and
  {Single}-{Task} {Construction} {Robots}, Vol.~4, Cambridge University Press,
  2016.

\bibitem{vink_physical_2002}
P.~Vink, M.~Miedema, E.~Koningsveld, H.~v.~d. Molen, Physical {Effects} of
  {New} {Devices} for {Bricklayers}, International Journal of Occupational
  Safety and Ergonomics 8~(1) (2002) 71--82.
\newblock \href {https://doi.org/10.1080/10803548.2002.11076515}
  {\path{doi:10.1080/10803548.2002.11076515}}.

\bibitem{hess_ergonomic_2010}
J.~Hess, M.~Weinstein, L.~Welch, Ergonomic {Best} {Practices} in {Masonry}:
  {Regional} {Differences}, {Benefits}, {Barriers}, and {Recommendations} for
  {Dissemination}, Journal of Occupational and Environmental Hygiene 7~(8)
  (2010) 446--455.
\newblock \href {https://doi.org/10.1080/15459624.2010.484795}
  {\path{doi:10.1080/15459624.2010.484795}}.

\bibitem{dakhli_robotic_2017}
Z.~Dakhli, Z.~Lafhaj, Robotic mechanical design for brick-laying automation,
  Cogent Engineering 4 (Aug. 2017).
\newblock \href {https://doi.org/10.1080/23311916.2017.1361600}
  {\path{doi:10.1080/23311916.2017.1361600}}.

\bibitem{thomson_brick-laying_1904}
J.~Thomson,
  \href{https://patents.google.com/patent/US772191A/en}{Brick-{Laying}
  {Machine}}, patent Number: US772191A (Oct. 1904).
\newline\urlprefix\url{https://patents.google.com/patent/US772191A/en}

\bibitem{british_pathe_mechanical_1967}
{British Pathé}, \href{https://www.youtube.com/watch?v=4MWald1Goqk}{Mechanical
  {Bricklayer}} (1967).
\newline\urlprefix\url{https://www.youtube.com/watch?v=4MWald1Goqk}

\bibitem{malinovsky_robotic_1990}
E.~Y. Malinovsky, A.~A. Borshchevsky, E.~A. Elder, V.~M. Pogodin, A {Robotic}
  {Complex} for {Brick}-{Laying} {Applications}, in: 7th {International}
  {Symposium} on {Automation} and {Robotics} in {Construction}, Bristol,
  England, 1990, pp. 32--38.

\bibitem{chamberlain_progress_1991}
D.~Chamberlain, P.~Speare, S.~Ala, Progress in a {Masonry} {Tasking} {Robot},
  in: 8th {Int}. {Symposium} on {Automation} and {Robotics} in {Construction},
  Stuttgart, Germany, 1991, pp. 909--918.

\bibitem{altobelli_prototype_1993}
F.~Altobelli, H.~F. Taylor, L.~E. Bernold, Prototype {Robotic} {Masonry}
  {System}, Journal of Aerospace Engineering 6~(1) (1993) 19--33.
\newblock \href {https://doi.org/10.1061/(ASCE)0893-1321(1993)6:1(19)}
  {\path{doi:10.1061/(ASCE)0893-1321(1993)6:1(19)}}.

\bibitem{bock_early_2009}
T.~Bock, T.~Linner, From {Early} {Trials} to {Advanced} {Computer} {Integrated}
  {Prefabrication} of {Brickwork}, in: G.~Grimscheid, F.~Scheublin (Eds.), New
  {Perspective} in {Industrialization} in {Construction} – {A}
  {State}-of-the-{Art} {Report}, CIB International Council for Research and
  Innovation in Building and Research, Zurich, 2009, pp. 161--181.

\bibitem{kodama_robotized_1988}
Y.~Kodama, Y.~Yamazaki, H.~Kato, Y.~Iguchi, H.~Naoi, A {Robotized} {Wall}
  {Erection} {System} with {Solid} {Components}, in: Proceedings of the 5th
  {International} {Symposium} on {Automation} and {Robotics} in {Construction}
  ({ISARC}), Tokyo, Japan, 1988, pp. 441--448.
\newblock \href {https://doi.org/10.22260/ISARC1988/0052}
  {\path{doi:10.22260/ISARC1988/0052}}.

\bibitem{andres_first_1994}
J.~Andres, T.~Bock, F.~Gebhart, W.~Steck, First {Results} of the {Development}
  of the {Masonry} {Robot} {System} {ROCCO}: a {Fault} {Tolerant} {Assembly}
  {Tool}, Automation and Robotics in Construction XI (1994) 87--93\href
  {https://doi.org/10.1016/B978-0-444-82044-0.50016-3}
  {\path{doi:10.1016/B978-0-444-82044-0.50016-3}}.

\bibitem{slocum_blockbot_1988}
A.~H. Slocum, B.~Schena, Blockbot: {A} robot to automate construction of cement
  block walls, Robotics and Autonomous Systems 4~(2) (1988) 111--129.
\newblock \href {https://doi.org/10.1016/0921-8890(88)90020-6}
  {\path{doi:10.1016/0921-8890(88)90020-6}}.

\bibitem{lehtinen_outlines_1989}
H.~Lehtinen, E.~Salo, H.~Aalto, Outlines of {Two} {Masonry} {Robot} {Systems},
  in: Proceedings of the 6th {International} {Symposium} on {Automation} and
  {Robotics} in {Construction}., San Francisco, California, 1989, p.~8.

\bibitem{podkaminer_sam100_2021}
N.~Podkaminer, L.~Peters,
  \href{https://www.construction-robotics.com/sam100/}{{SAM100} –
  {Construction} {Robotics}} (2021).
\newline\urlprefix\url{https://www.construction-robotics.com/sam100/}

\bibitem{pivac_hadrian_2021}
M.~Pivac, M.~Pivac, M.~Sheridan, \href{https://www.fbr.com.au/}{Hadrian {X} -
  {Fastbrick} {Robotics}} (2021).
\newline\urlprefix\url{https://www.fbr.com.au/}

\bibitem{pritschow_mobile_1994}
G.~Pritschow, M.~Dalacker, J.~Kurz, J.~Zeiher, A mobile robot for on-site
  construction of masonry, Proceedings of IEEE/RSJ International Conference on
  Intelligent Robots and Systems (IROS'94) 3 (1994) 1701--1707.
\newblock \href {https://doi.org/10.1109/IROS.1994.407628}
  {\path{doi:10.1109/IROS.1994.407628}}.

\bibitem{pritschow_technological_1996}
G.~Pritschow, M.~Dalacker, J.~Kurz, M.~Gaenssle, Technological aspects in the
  development of a mobile bricklaying robot, Automation in Construction 5~(1)
  (1996) 3--13.
\newblock \href {https://doi.org/10.1016/0926-5805(95)00015-1}
  {\path{doi:10.1016/0926-5805(95)00015-1}}.

\bibitem{sklar_robots_2015}
J.~Sklar,
  \href{https://www.technologyreview.com/2015/09/02/10587/robots-lay-three-times-as-many-bricks-as-construction-workers/}{Robots
  {Lay} {Three} {Times} as {Many} {Bricks} as {Construction} {Workers}} (2015).
\newline\urlprefix\url{https://www.technologyreview.com/2015/09/02/10587/robots-lay-three-times-as-many-bricks-as-construction-workers/}

\bibitem{gramazio_digital_2008}
F.~Gramazio, M.~Kohler, Digital {Materiality} in {Architecture}, 1st Edition,
  Lars Muller, Baden, 2008.

\bibitem{gramazio_made_2014}
F.~Gramazio, M.~Kohler (Eds.), Made by {Robots}: {Challenging} {Architecture}
  at a {Larger} {Scale}, 1st Edition, Academy Press, London, 2014.

\bibitem{bonwetsch_informed_2006}
T.~Bonwetsch, D.~Kobel, F.~Gramazio, M.~Kohler, The {Informed} {Wall}:
  {Applying} {Additive} {Digital} {Fabrication} {Techniques} on {Architecture},
  in: Proceedings of the 25th {Annual} {Conference} of the {Association} for
  {Computer}-{Aided} {Design} in {Architecture}, 2006, pp. 489--495.

\bibitem{bonwetsch_digitally_2007}
T.~Bonwetsch, F.~Gramazio, M.~Kohler, Digitally {Fabricating}
  {Non}-{Standardised} {Brick} {Walls}, in: {ManuBuild}, 2007, pp. 191--196.

\bibitem{kohler_gantenbein_2014}
M.~Kohler, F.~Gramazio, J.~Willmann, Gantenbein {Vineyard} {Façade}, in: The
  {Robotic} {Touch}: {How} {Robots} {Change} {Architecture}, The University of
  Chicago Press, 2014, pp. 66--75.

\bibitem{empa_dfab_2021}
EMPA, \href{https://www.empa.ch/web/nest/digital-fabrication}{{DFAB} {HOUSE}
  – {Digital} {Fabrication} and {Living}.} (2021).
\newline\urlprefix\url{https://www.empa.ch/web/nest/digital-fabrication}

\bibitem{willmann_robotic_2016}
J.~Willmann, M.~Knauss, T.~Bonwetsch, A.~A. Apolinarska, F.~Gramazio,
  M.~Kohler, Robotic timber construction — {Expanding} additive fabrication
  to new dimensions, Automation in Construction 61 (2016) 16--23.
\newblock \href {https://doi.org/10.1016/j.autcon.2015.09.011}
  {\path{doi:10.1016/j.autcon.2015.09.011}}.

\bibitem{hack_mesh_2017}
N.~Hack, T.~Wangler, J.~Mata-Falcón, K.~Dörfler, N.~Kumar, N.~Walzer,
  K.~Graser, L.~Reiter, H.~Richner, J.~Buchli, W.~Kaufmann, R.~J. Flatt,
  F.~Gramazio, M.~Kohler, Mesh {Mould}: {An} {On} {Site}, {Robotically}
  {Fabricated}, {Functional} {Formwork}, in: Second {Concrete} {Innovation}
  {Conference}, Vol.~19, Tromsø, Norway, 2017, pp. 1--11.

\bibitem{hack_structural_2020}
N.~Hack, K.~Dörfler, A.~N. Walzer, T.~Wangler, J.~Mata-Falcón, N.~Kumar,
  J.~Buchli, W.~Kaufmann, R.~J. Flatt, F.~Gramazio, M.~Kohler, Structural
  stay-in-place formwork for robotic in situ fabrication of non-standard
  concrete structures: {A} real scale architectural demonstrator, Automation in
  Construction 115 (2020) 1--13.
\newblock \href {https://doi.org/10.1016/j.autcon.2020.103197}
  {\path{doi:10.1016/j.autcon.2020.103197}}.

\bibitem{bonwetsch_brickdesign_2012}
T.~Bonwetsch, R.~Bärtschi, M.~Helmreich, {BrickDesign}, in: Robotic
  {Fabrication} in {Architecture}, {Art} and {Design} 2012, Springer
  International Publishing, Vienna, 2012, pp. 102--109.

\bibitem{mele_compas_2017}
T.~V. Mele, A.~Liew, T.~M. Echenagucia, M.~Rippmann,
  \href{https://compas-dev.github.io/}{{COMPAS}: {A} framework for
  computational research in architecture and structures.} (2017).
\newline\urlprefix\url{https://compas-dev.github.io/}

\bibitem{jahn_holographic_2018}
G.~Jahn, C.~Newnham, N.~v.~d. Berg, M.~Beanland, Holographic design,
  fabrication, assembly and analysis of woven steel structures, in: Proceedings
  of the 38th {Annual} {Conference} of the {Association} for {Computer} {Aided}
  {Design} in {Architecture}, Mexico City, 2018, pp. 1--11.

\bibitem{jahn_holographic_2019}
G.~Jahn, C.~Newnham, N.~Berg, M.~Iraheta, J.~Wells, Holographic {Construction},
  in: Design {Modelling} {Symposium}, Berlin, 2019, pp. 1--13.

\bibitem{mitterberger_augmented_2020}
D.~Mitterberger, K.~Dörfler, T.~Sandy, F.~Salveridou, M.~Hutter, F.~Gramazio,
  M.~Kohler, Augmented bricklaying, Construction Robotics 4~(3) (2020)
  151--161.
\newblock \href {https://doi.org/10.1007/s41693-020-00035-8}
  {\path{doi:10.1007/s41693-020-00035-8}}.

\bibitem{piskorec_brick_2018}
L.~Piškorec, D.~Jenny, S.~Parascho, H.~Mayer, F.~Gramazio, M.~Kohler, The
  {Brick} {Labyrinth}, in: Robotic {Fabrication} in {Architecture}, {Art} and
  {Design} 2018, Springer International Publishing, Zurich, 2018, pp. 489--500.

\bibitem{dorfler_mobile_2016}
K.~Dörfler, T.~Sandy, M.~Giftthaler, F.~Gramazio, M.~Kohler, J.~Buchli, Mobile
  {Robotic} {Brickwork}. {Automation} of a {Discrete} {Robotic} {Fabrication}
  {Process} {Using} an {Autonomous} {Mobile} {Robot}, in: Robotic {Fabrication}
  in {Architecture}, {Art} and {Design} 2016, Springer International
  Publishing, Sydney, Australia, 2016, pp. 204--217.
\newblock \href {https://doi.org/10.1007/978-3-319-26378-6-15}
  {\path{doi:10.1007/978-3-319-26378-6-15}}.

\bibitem{giftthaler_mobile_2017}
M.~Giftthaler, T.~Sandy, K.~Dörfler, I.~Brooks, M.~Buckingham, G.~Rey,
  M.~Kohler, F.~Gramazio, J.~Buchli, Mobile {Robotic} {Fabrication} at 1:1
  {Scale}: the {In} situ {Fabricator}, Construction Robotics 1~(1) (2017)
  3--14.
\newblock \href {https://doi.org/10.1007/s41693-017-0003-5}
  {\path{doi:10.1007/s41693-017-0003-5}}.

\bibitem{bartschi_wiggled_2010}
R.~Bärtschi, M.~Knauss, T.~Bonwetsch, F.~Gramazio, M.~Kohler, Wiggled {Brick}
  {Bond}, in: Advances in {Architectural} {Geometry} 2010, Springer, Wien/New
  York, 2010, pp. 137--147.

\bibitem{kohler_programmed_2014}
M.~Kohler, F.~Gramazio, J.~Willmann, The {Programmed} {Column}, in: The
  {Robotic} {Touch}: {How} {Robots} {Change} {Architecture}, The University of
  Chicago Press, 2014, pp. 208--215.

\bibitem{carneau_exploration_2019}
P.~Carneau, R.~Mesnil, N.~Roussel, O.~Baverel, An exploration of 3d printing
  design space inspired by masonry, in: Proceedings of {IASS} {Annual}
  {Symposium} 2019, Zenodo, Barcelona, 2019, pp. 1--9.
\newblock \href {https://doi.org/10.5281/ZENODO.3563672}
  {\path{doi:10.5281/ZENODO.3563672}}.

\bibitem{carneau_additive_2020}
P.~Carneau, R.~Mesnil, N.~Roussel, O.~Baverel, Additive manufacturing of
  cantilever - {From} masonry to concrete {3D} printing, Automation in
  Construction 116 (2020) 1--16.
\newblock \href {https://doi.org/10.1016/j.autcon.2020.103184}
  {\path{doi:10.1016/j.autcon.2020.103184}}.

\bibitem{motamedi_supportless_2019}
M.~Motamedi, R.~Oval, P.~Carneau, O.~Baverel, Supportless {3D} {Printing} of
  {Shells}: {Adaptation} of {Ancient} {Vaulting} {Techniques} to {Digital}
  {Fabrication}, in: Proceedings of the {Design} {Modelling} {Symposium}
  {Berlin}, Berlin, 2019, pp. 714--726.
\newblock \href {https://doi.org/10.1007/978-3-030-29829-6-55}
  {\path{doi:10.1007/978-3-030-29829-6-55}}.

\bibitem{parascho_cooperative_2019}
S.~Parascho, Cooperative {Robotic} {Assembly}: {Computational} {Design} and
  {Robotic} {Fabrication} of {Spatial} {Metal} {Structures}, {PhD} {Thesis},
  ETH, Zurich (2019).

\bibitem{thoma_robotic_2018}
A.~Thoma, A.~Adel, M.~Helmreich, T.~Wehrle, F.~Gramazio, M.~Kohler, Robotic
  {Fabrication} of {Bespoke} {Timber} {Frame} {Modules}, in: Robotic
  {Fabrication} in {Architecture}, {Art} and {Design} 2018, Springer
  International Publishing, Zurich, 2018, pp. 447--458.

\bibitem{bruun_humanrobot_2020}
E.~P.~G. Bruun, I.~Ting, S.~Adriaenssens, S.~Parascho, Human–{Robot}
  {Collaboration}: {A} {Fabrication} {Framework} for the {Sequential} {Design}
  and {Construction} of {Unplanned} {Spatial} {Structures}, Digital Creativity
  31~(4) (2020) 320--336.
\newblock \href {https://doi.org/10.1080/14626268.2020.1845214}
  {\path{doi:10.1080/14626268.2020.1845214}}.

\bibitem{sarhosis_review_2016}
V.~Sarhosis, S.~D. Santis, G.~d. Felice, A review of experimental
  investigations and assessment methods for masonry arch bridges, Structure and
  Infrastructure Engineering 12~(11) (2016) 1439--1464.
\newblock \href {https://doi.org/10.1080/15732479.2015.1136655}
  {\path{doi:10.1080/15732479.2015.1136655}}.

\bibitem{daltri_modeling_2020}
A.~M. D'Altri, V.~Sarhosis, G.~Milani, J.~Rots, S.~Cattari, S.~Lagomarsino,
  E.~Sacco, A.~Tralli, G.~Castellazzi, S.~de~Miranda, Modeling {Strategies} for
  the {Computational} {Analysis} of {Unreinforced} {Masonry} {Structures}:
  {Review} and {Classification}, Archives of Computational Methods in
  Engineering 27~(4) (2020) 1153--1185.
\newblock \href {https://doi.org/10.1007/s11831-019-09351-x}
  {\path{doi:10.1007/s11831-019-09351-x}}.

\bibitem{fang_assessing_2019}
D.~L. Fang, R.~K. Napolitano, T.~L. Michiels, S.~M. Adriaenssens, Assessing the
  stability of unreinforced masonry arches and vaults: a comparison of
  analytical and numerical strategies, International Journal of Architectural
  Heritage 13~(5) (2019) 648--662.
\newblock \href {https://doi.org/10.1080/15583058.2018.1463413}
  {\path{doi:10.1080/15583058.2018.1463413}}.

\bibitem{kaminski_tests_2010}
T.~Kamiński, Tests to {Collapse} of {Masonry} {Arch} {Bridges} {Simulated} by
  {Means} of {FEM}, in: Proceedings of the {Fifth} {International} {Conference}
  on {Bridge} {Maintenance}, {Safety} and {Management}, CRC Press,
  Philadelphia, USA, 2010, pp. 1431--1438.
\newblock \href {https://doi.org/10.13140/2.1.4381.2649}
  {\path{doi:10.13140/2.1.4381.2649}}.

\bibitem{thavalingam_computational_2001}
A.~Thavalingam, N.~Bicanic, J.~I. Robinson, D.~A. Ponniah, Computational
  framework for discontinuous modelling of masonry arch bridges, Computers \&
  Structures 79~(19) (2001) 1821--1830.
\newblock \href {https://doi.org/10.1016/S0045-7949(01)00102-X}
  {\path{doi:10.1016/S0045-7949(01)00102-X}}.

\bibitem{simon_discrete_2016}
J.~Simon, K.~Bagi, Discrete {Element} {Analysis} of the {Minimum} {Thickness}
  of {Oval} {Masonry} {Domes}, International Journal of Architectural Heritage
  10~(4) (2016) 457--475.
\newblock \href {https://doi.org/10.1080/15583058.2014.996921}
  {\path{doi:10.1080/15583058.2014.996921}}.

\bibitem{cundall_formulation_1988}
P.~A. Cundall, Formulation of a {Three}-{Dimensional} {Distinct} {Element}
  {Model}—{Part} {I}. {A} {Scheme} to {Detect} and {Represent} {Contacts} in
  a {System} {Composed} of {Many} {Polyhedral} {Blocks}, in: International
  {Journal} of {Rock} {Mechanics} and {Mining} {Sciences} \& {Geomechanics}
  {Abstracts}, Vol.~25, 1988, pp. 107--116.

\bibitem{hart_formulation_1988}
R.~Hart, P.~A. Cundall, J.~Lemos, Formulation of a {Three}-{Dimensional}
  {Distinct} element model—{Part} {II}. {Mechanical} calculations for motion
  and interaction of a {System} {Composed} of {Many} {Polyhedral} {Blocks}, in:
  International journal of rock mechanics and mining sciences \& {Geomechanics}
  abstracts, Vol.~25, Elsevier, 1988, pp. 117--125.

\bibitem{roberti_distinct_1998}
G.~M. Roberti, F.~Calvetti, Distinct {Element} {Analysis} of {Stone} {Arches},
  Arch Bridges (1998) 181--186.

\bibitem{kassotakis_discrete_2017}
N.~Kassotakis, V.~Sarhosis, T.~Forgàcs, K.~Bagi, Discrete {Element}
  {Modelling} of {Multi}-{Ring} {Brickwork} {Masonry} {Arches}, in: 13th
  {Canadian} {Masonry} {Symposium}, Newcastle University, 2017, pp. 1--11.

\bibitem{meriggi_distinct_2019}
P.~Meriggi, G.~de~Felice, S.~De~Santis, F.~Gobbin, A.~Mordanova, B.~Pantò,
  Distinct {Element} {Modelling} of {Masonry} {Walls} {Under}
  {Out}-{Of}-{Plane} {Seismic} {Loading}, International Journal of
  Architectural Heritage 13~(7) (2019).

\bibitem{abaqus_inc_abaqus_2020}
{ABAQUS Inc.},
  \href{https://www.3ds.com/products-services/simulia/products/abaqus/}{Abaqus
  {FEA}} (2020).
\newline\urlprefix\url{https://www.3ds.com/products-services/simulia/products/abaqus/}

\bibitem{itasca_consulting_group_inc_3dec_2016}
{Itasca Consulting Group Inc.},
  \href{https://www.itascacg.com/software/3dec}{{3DEC}} (2016).
\newline\urlprefix\url{https://www.itascacg.com/software/3dec}

\bibitem{heyman_stone_1966}
J.~Heyman, The {Stone} {Skeleton}, International Journal of Solids and
  Structures 2~(2) (1966) 249--279.
\newblock \href {https://doi.org/10.1016/0020-7683(66)90018-7}
  {\path{doi:10.1016/0020-7683(66)90018-7}}.

\bibitem{block_as_2006}
P.~Block, M.~DeJong, J.~Ochsendorf, As {Hangs} the {Flexible} {Line}:
  {Equilibrium} of {Masonry} {Arches}, Nexus Network Journal 8~(2) (2006)
  13--24.
\newblock \href {https://doi.org/10.1007/s00004-006-0015-9}
  {\path{doi:10.1007/s00004-006-0015-9}}.

\bibitem{lau_equilibrium_2006}
W.~W. Lau, \href{https://dspace.mit.edu/handle/1721.1/34984}{Equilibrium
  analysis of masonry domes}, {PhD} {Thesis}, Massachusetts Institute of
  Technology (2006).
\newline\urlprefix\url{https://dspace.mit.edu/handle/1721.1/34984}

\bibitem{michiels_form-finding_2018}
T.~Michiels, S.~Adriaenssens, Form-finding algorithm for masonry arches
  subjected to in-plane earthquake loading, Computers \& Structures 195 (2018)
  85--98.
\newblock \href {https://doi.org/10.1016/j.compstruc.2017.10.001}
  {\path{doi:10.1016/j.compstruc.2017.10.001}}.

\bibitem{odwyer_funicular_1999}
D.~O’Dwyer, Funicular analysis of masonry vaults, Computers \& Structures
  73~(1) (1999) 187--197.
\newblock \href {https://doi.org/10.1016/S0045-7949(98)00279-X}
  {\path{doi:10.1016/S0045-7949(98)00279-X}}.

\bibitem{block_thrust_2007}
P.~Block, J.~Ochsendorf, Thrust {Network} {Analysis}: {A} new methodology for
  three-dimensional equilibrium, Journal of the International Association for
  Shell and Spatial Structures 48~(3) (2007) 167--173.

\bibitem{abb_product_2020}
{ABB}, Product specification - {IRB} 4600, Tech. Rep. 3HAC032885-001, ABB
  Group, workspace 20D, version a7 (Dec. 2020).

\bibitem{kaveh_role_2005}
A.~Kaveh, The {Role} of {Algebraic} {Graph} {Theory} in {Structural}
  {Mechanics}, in: B.~Topping, C.~Mota~Soares (Eds.), Progress in
  {Computational} {Structures} {Technology}, 1st Edition, Saxe-Coburg
  Publications, 2005, pp. 77--110.

\bibitem{kaveh_efficient_2010}
A.~Kaveh, H.~Rahami, An efficient analysis of repetitive structures generated
  by graph products, International Journal for Numerical Methods in Engineering
  84~(1) (2010) 108--126.
\newblock \href {https://doi.org/10.1002/nme.2893}
  {\path{doi:10.1002/nme.2893}}.

\bibitem{rutten_grasshopper_2007}
D.~Rutten, \href{https://www.grasshopper3d.com/}{Grasshopper} (2007).
\newline\urlprefix\url{https://www.grasshopper3d.com/}

\bibitem{preisinger_karambatoolkit_2014}
C.~Preisinger, M.~Heimrath, Karamba—{A} {Toolkit} for {Parametric}
  {Structural} {Design}, Structural Engineering International 24~(2) (2014)
  217--221.
\newblock \href {https://doi.org/10.2749/101686614X13830790993483}
  {\path{doi:10.2749/101686614X13830790993483}}.

\bibitem{royles_model_1991}
R.~Royles, A.~W. Hendry, Model {Tests} on {Masonry} {Arches}, Proceedings of
  the Institution of Civil Engineers 91~(2) (1991) 299--321.
\newblock \href {https://doi.org/10.1680/iicep.1991.14997}
  {\path{doi:10.1680/iicep.1991.14997}}.

\bibitem{melbourne_load_1988}
C.~Melbourne, P.~Walker, Load {Tests} to {Collapse} of {Model} {Brickwork}
  {Masonry} {Arches}, in: Proceedings of the 8th {International}
  {Brick}/{Block} {Masonry} {Conference}, Dublin, Ireland, 1988, pp. 991--1002.

\bibitem{melbourne_collapse_1997}
C.~Melbourne, M.~Gilbert, M.~Wagstaff, The {Collapse} {Behaviour} of
  {Multispan} {Brickwork} {Arch} {Bridges}, The Structural Engineer 75~(17)
  (1997) 297--305.

\bibitem{melbourne_behaviour_1995}
C.~Melbourne, M.~Gilbert, The {Behaviour} of {Multi}-{Ring} {Brickwork}
  {Bridges}, The Structural Engineer 73 (1995) 39--47.

\bibitem{gilbert_small_2007}
M.~Gilbert, C.~C. Smith, J.~Wang, Small and {Large}-{Scale} {Experimental}
  {Studies} of {Soil}-{Arch} {Interaction} in {Masonry} {Bridges}, in: 5th
  {International} {Conference} on {Arch} {Bridges}, Funchal, Madeira, 2007, pp.
  1--8.

\bibitem{towler_limit_1982}
K.~Towler, F.~Sawko, Limit {State} {Behaviour} of {Brickwork} {Arches}, in: 6th
  {International} {Conference} on {Brick} {Masonry}, Rome, Italy, 1982, pp.
  422--429.

\bibitem{swift_physical_2013}
G.~Swift, L.~Augusthus-Nelson, C.~Melbourne, M.~Gilbert, Physical {Modelling}
  of {Masonry} {Arch} {Bridges}, in: 7th {International} {Conference} on {Arch}
  {Bridges}, Split, Croatia, 2013, pp. 621--628.

\bibitem{turner_stiffness_1956}
M.~J. Turner, R.~W. Clough, H.~C. Martin, L.~J. Topp, Stiffness and
  {Deflection} {Analysis} of {Complex} {Structures}, Journal of the
  Aeronautical Sciences 23~(9) (1956) 805--823.
\newblock \href {https://doi.org/10.2514/8.3664} {\path{doi:10.2514/8.3664}}.

\bibitem{oatey_oatey_2020}
Oatey,
  \href{https://www.oatey.com/products/oatey-fixit-stick-epoxy-putty-1829259701}{Oatey®
  {Fix}-{It}™ {Stick} {Epoxy} {Putty}} (2020).
\newline\urlprefix\url{https://www.oatey.com/products/oatey-fixit-stick-epoxy-putty-1829259701}

\bibitem{forgacs_minimum_2017}
T.~Forgács, V.~Sarhosis, K.~Bagi, Minimum {Thickness} of {Semi}-{Circular}
  {Skewed} {Masonry} {Arches}, Engineering Structures 140 (2017) 317 -- 336.
\newblock \href {https://doi.org/10.1016/j.engstruct.2017.02.036}
  {\path{doi:10.1016/j.engstruct.2017.02.036}}.

\end{thebibliography}
